\documentclass[preprint,review,authoryear,3p]{elsarticle}
\usepackage{amssymb}
\usepackage{natbib}
\usepackage{longtable}

\usepackage{caption2}
\usepackage{graphicx}
\usepackage{url}
\usepackage[colorlinks,citecolor=blue,linkcolor=blue,urlcolor=black,hyperindex]{hyperref}
\usepackage{booktabs}
\usepackage{enumerate}
\usepackage{setspace}

\usepackage{bm}
\usepackage{threeparttable}
\usepackage{float}
\usepackage{xfrac}
\usepackage{array}

\usepackage{geometry}
\usepackage{multirow}
\usepackage[OT2,T1]{fontenc}

\usepackage{amsmath}
\usepackage{amsthm}
\renewcommand{\thefigure}{\arabic{figure}}
\usepackage[switch]{lineno}

\newtheorem{theorem}{Theorem}

\def\ack{\section*{Acknowledgements}%
  \addtocontents{toc}{\protect\vspace{6pt}}%
  \addcontentsline{toc}{section}{Acknowledgements}%
}

\theoremstyle{definition}
\newtheorem{definition}{Definition}

\newtheorem{algorithm}{Algorithm}
\renewcommand{\thealgorithm}{\Roman{algorithm}}

\newtheorem{remark}{Remark}

\allowdisplaybreaks

\journal{Elsevier}

\newcommand{\bea}{\begin{eqnarray}}
\newcommand{\eea}{\end{eqnarray}}
\newcommand{\ben}{\begin{equation}}
\newcommand{\een}{\end{equation}}

\usepackage{algorithm}
\usepackage{algorithmicx}
\usepackage{algpseudocode}
\floatname{algorithm}{Algorithm}

\newcommand{\RNum}[1]{\uppercase\expandafter{\romannumeral #1\relax}}

\usepackage{graphicx}

\begin{document}
\begin{frontmatter}
\title{An incremental preference elicitation-based approach to learning potentially non-monotonic preferences in multi-criteria sorting}

\author[dut1]{Zhuolin Li}
\ead{lizhuolin@mail.dlut.edu.cn}

\author[dut1]{Zhen Zhang\corref{cor1}}
\ead{zhen.zhang@dlut.edu.cn}

\author[UA1,UA2,UA3]{Witold Pedrycz}
\ead{wpedrycz@ualberta.ca}

\cortext[cor1]{Corresponding author.}

\address[dut1]{School of Economics and Management, Dalian University of Technology, Dalian 116024, P. R. China}

\address[UA1]{Department of Electrical and Computer Engineering, University of Alberta, Edmonton T6G 2R3, Canada}
\address[UA2]{Systems Research Institute, Polish Academy of Sciences, Warsaw 00-901, Poland}
\address[UA3]{Department of Computer Engineering, Faculty of Engineering and Natural Sciences, Istinye University, Sariyer/Istanbul Turkiye}

\begin{abstract}
This paper introduces a novel incremental preference elicitation-based approach to learning potentially non-monotonic preferences in MCS problems, enabling decision makers to progressively provide assignment example preference information. Specifically, we first construct a max-margin optimization-based model to model potentially non-monotonic preferences and inconsistent assignment example preference information in each iteration of the incremental preference elicitation process. Using the optimal objective function value of the max-margin optimization-based model, we devise information amount measurement methods and question selection strategies to pinpoint the most informative alternative in each iteration within the framework of uncertainty sampling in active learning. Once the termination criterion is satisfied, the sorting result for non-reference alternatives can be determined through the use of two optimization models, i.e., the max-margin optimization-based model and the complexity controlling optimization model. Subsequently, two incremental preference elicitation-based algorithms are developed to learn potentially non-monotonic preferences, considering different termination criteria. Ultimately, we apply the proposed approach to a credit rating problem to elucidate the detailed implementation steps, and perform computational experiments on both artificial and real-world data sets to compare the proposed question selection strategies with several benchmark strategies.
\end{abstract}
\begin{keyword}
Multi-criteria sorting, preference learning, preference elicitation, non-monotonic preferences, active learning
\end{keyword}
\end{frontmatter}

\section{Introduction}\label{sec:1}
Multi-criteria sorting (MCS) arises as a rapidly developing branch within the realm of multi-criteria decision making. Its primary objective is to assign a given set of alternatives to several predefined ordered categories, taking into account multiple criteria \citep{Zopounidis02ejor,Belahcene234or,Li24jors}. It finds extensive applications in various real-life scenarios, including but not limited to credit rating \citep{Doumpos19omega}, inventory management \citep{Liu16omega}, supplier evaluation \citep{Pelissari22eswa}, green building rating \citep{Zhang23aor} and policy assessments \citep{Dias18omega}. In the literature, numerous MCS approaches have been proposed, which can be broadly categorized into the following four distinct groups: utility function-based MCS approaches \citep{Greco10ejor,Liu20joc,Wu23omega,Ru23ejor}, outranking-based MCS approaches \citep{Doumpos09ejor,Fernandez19asoc,Almeida10ejor,Kadzinski15omega}, distance-based MCS approaches \citep{de20caie,de20eswa} and decision rule-based MCS approaches \citep{Kadzinski16ins}.

Among various approaches, the threshold-based MCS model, falling under the group of utility function-based MCS approaches, has garnered considerable attention due to its high interpretability and intuitive understandability for decision makers \citep{Liu19ejor,Li24tcss}. This model quantifies each alternative utilizing an additive utility function, introduces category thresholds to delimit each category, and then determines the assignment of each alternative by comparing its comprehensive utility value with category thresholds \citep{Kadzinski20ejor,Liu23joc}. Prior to adopting this model, it is essential to determine the shape of marginal utility functions and category thresholds.
To reduce the cognitive effort required by decision makers \citep{Doumpos11ejor,Kadzinski21kbs,Hullermeier24-4or1,Hullermeier24-4or2}, the indirect elicitation-based MCS approach, which infers the shape of marginal utility functions and category thresholds based on preference information such as assignment examples, stand out \citep{Liu20ejor,Jacquet01ejor,Kadzinski17caor,Doumpos19springer}.
{\color{black} There are usually two types of indirect elicitation approaches: batch elicitation and incremental elicitation \citep{Khannoussi224or,Khannoussi24or}. In batch elicitation, preference information is provided as a ``batch'' by the decision maker, and numerous batch elicitation-based MCS models have been developed in the literature.} For instance, the classic MCS method, the UTilit\'{e}s Additives DIScriminantes (UTADIS), transforms the batch assignment example preference information into some constraints by using mathematical programming techniques \citep{Devaud80ewg}. Followed by this, various variants of UTADIS have been proposed, incorporating distinct objective functions or considering diverse extensions \citep{Esmaelian16asoc,Wojcik23kbs}. To consider all compatible information with the assignment example preference information provided by the decision maker, the Robust Ordinal Regression (ROR) and Stochastic Ordinal Regression (SOR) methods were proposed \citep{Kadzinski13dss,Doumpos14ejor}. {\color{black}In incremental elicitation, preference information is supplied sequentially, allowing the preference model to improve iteratively. Incremental elicitation-based MCS models typically start with an initial set of reference alternatives, where the decision maker is asked to provide category assignment for these reference alternatives. This information is then incrementally integrated into the preference model. This process continues until a predetermined termination criterion is met. Compared to batch elicitation-based MCS models, incremental elicitation-based MCS models reduce the number of holistic judgments required from the decision maker, thereby lowering their cognitive effort and enhancing the efficiency of the decision making process.} For instance, \cite{Benabbou17ai} proposed an incremental elicitation method of Choquet capacities for multi-criteria decision making by using minimax regret strategy. \cite{Nefla19adt} presented an interactive elicitation approach for the majority rule sorting model. \cite{Ozpeynirci18AOR} developed an interactive algorithm for MCS problems with category size restrictions. \cite{Kadzinski21ejor} designed some active learning strategies to interactively elicit assignment examples for the threshold-based MCS model based on the outcomes of ROR and SOR. \cite{Gehrlein23omega} designed an active learning approach by interactively eliciting pairwise preferences and applied it for validator selection problems in blockchain environments.

Although these incremental elicitation-based MCS models are effective, they may face challenges.
{\color{black}
In some MCS scenarios, decision makers may provide answers that are inconsistent with previously given preference information during the incremental preference elicitation process \citep{Teso16ijcai}, and may exhibit non-monotonic preferences \citep{Ghaderi17ejor,Guo19eswa}. The oversight of inconsistencies and potential non-monotonic preferences may lead to conflicts and limit the applicability  of these methods. Therefore, it is crucial to take into account both inconsistencies and potentially non-monotonic preferences in incremental elicitation-based MCS models.}
Despite \cite{Guo19eswa} introduced a progressive approach for MCS problems with non-monotonic preferences and inconsistent preference information, this method overlooks the potential discrepancy in information amount across different alternatives, which may escalate the cognitive load on the decision maker by necessitating increased number of assignment example preference information. As a result, it is imperative to introduce a novel incremental preference elicitation-based approach to learning potentially non-monotonic preferences in MCS problems, taking into account inconsistent assignment example preference information. Consequently, the primary contributions of this paper are summarized as follows:

First, we develop a max-margin optimization-based model to effectively handle potentially non-monotonic preferences and inconsistent assignment example preference information encountered in each iteration of the incremental preference elicitation process. Specifically, this model introduces some auxiliary variables to tolerate such inconsistencies, aiming to simultaneously maximize discriminative power while minimizing inconsistencies.

Second, leveraging the optimal objective function values of the developed max-margin optimization-based model, we design some information amount measurement methods and question selection strategies to identify the most informative alternative in each iteration of the incremental preference elicitation process. In particular, all proposed question selection strategies adhere to the uncertainty sampling framework in active learning \citep{Aggarwal14dc}.

Third, taking into account different termination criteria, two incremental preference elicitation-based algorithms are introduced to learn potentially non-monotonic preferences for MCS problems. Subsequently, we conduct extensive computational experiments, comparing the proposed question selection strategies with several benchmark strategies. These experiments are performed on both artificial and real-world data sets to demonstrate the effectiveness of the proposed approach.

The rest of the paper is organized in the following manner. Section \ref{sec:2} presents the underlying sorting method utilized in this paper. Section \ref{sec:3} describes the research problem and then provides a resolution framework. Section \ref{sec:4} provides a comprehensive exposition of the proposed approach. Section \ref{sec:5} performs an illustrative example and some detailed computation experiments to elaborate and justify the proposed approach. Section \ref{sec:6} offers a comprehensive summary of this paper and identifies future directions.

\section{The threshold-based MCS model}\label{sec:2}
Let us consider a finite set of alternatives $A=\{a_1,a_2,\ldots,a_n\}$ which are evaluated in light of a family of criteria $G=\{g_1,g_2,\ldots,g_m\}$, where $a_i$ is the $i$-th alternative, $i\in N=\{1,2,\ldots,n\}$, $g_j$ denotes the $j$-th criterion, $j\in M=\{1,2,\ldots,m\}$. The performance level of the alternative $a_i$ over the criterion $g_j$ is denoted by $x_{ij}$, and the performance levels of all alternatives in $A$ over all criteria in $G$ form a decision matrix $X=(x_{ij})_{n\times m}$. The aim of the MCS problem is to assign each alternative in $A$ to a predefined category in $C=\{C_1,\ldots,C_q\}$ , where $C_h$ is the $h$-th category, $h\in Q=\{1,\ldots,q\}$ and $C_{h}\succ C_{h-1}$, $h=2,\ldots,q$. To accomplish the assignment of each alternative, the threshold-based MCS model adopts an additive utility function to aggregate the marginal utility of each alternative $a_i$ over all criteria $g_j$, $j\in M$ into a comprehensive utility \citep{Greco10ejor}, i.e.,
\begin{equation}\label{eq:global_v}
U(a_i)=\sum\limits_{j=1}^mu_j(x_{ij}), i\in N,
\end{equation}
where $0\le U(a_i)\le 1$ is the comprehensive utility of the alternative $a_i$,
$u_j(x_{ij})$ is the marginal utility of $x_{ij}$, and $u_j(\cdot)$ is the marginal utility function over the criterion $g_j$, $i\in N$, $j\in M$.

In the basic setting of the threshold-based MCS model, all criteria are assumed to be maximized, i.e., the greater $x_{ij}$, the better the alternative $a_i$ on the criterion $g_j$, $i\in N$, $j\in M$. As a result, the marginal utility function $u_j(\cdot)$ which is assumed to be non-decreasing and piecewise linear is employed to depict the decision maker's preference over the criterion $g_j$, $j\in M$ \citep{Liu20joc}.

In particular, the marginal utility function is defined with $s_j+1$ characteristic points, $j\in M$. Let $\beta_j^-$ and $\beta_j^+$ be the minimum and maximum performance levels observed for all alternatives in $A$ over the criterion $g_j$ such that $\beta_j^- = \min\limits_{i\in N} x_{ij}$ and $\beta_j^+ = \max\limits_{i\in N} x_{ij}$, $j\in M$,  then the interval $[\beta_j^-,\beta_j^+]$ constitutes the performance range of the alternatives in $A$ over the criterion $g_j$, $j\in M$. The characteristic points $\beta_j^l$, $l=1,\ldots,s_j+1$ divides the interval $[\beta_j^-,\beta_j^+]$ into $s_j$ subintervals of equal length, i.e., $[\beta_j^1,\beta_j^2]$,\ldots,$[\beta_j^l,\beta_j^{l+1}]$, \ldots, $[\beta_j^{s_j},\beta_j^{s_j+1}]$, where $\beta_j^1=g_j^-$, $\beta_j^{s_j+1}=g_j^+$, and $\beta_j^l=\beta_j^1+\frac{l-1}{s_j}(\beta_j^{s_j+1}-\beta_j^1)$, $l=2,\ldots,s_j$.

Let $u_j(\beta_j^l)$ be the marginal utility of the characteristic point $\beta_j^l$ on the criterion $g_j$, where $u_j(\beta_j^1)=0$ and $\sum\nolimits_{j=1}^m u_j(\beta_j^{s_j+1})=1$. Through linear interpolation, the marginal utility of $x_{ij}$ can be computed as

\begin{equation}\label{eq:marginal_v0}
u_j(x_{ij})=u_j(\beta^l_j)+\frac{x_{ij}-\beta^l_j}{\beta^{l+1}_j-\beta^l_j}(u_j(\beta^{l+1}_j)-u_j(\beta^l_j)), \text{if}\ x_{ij}\in[\beta^l_j,\beta^{l+1}_j].
\end{equation}

Furthermore, in the threshold-based MCS model, each category is limited by some thresholds. To do so, a category vector $b=(b_0,b_1,\ldots,b_q)^{\rm T}$ is required, where $b_{h-1}$ and $b_{h}$ are the lower and upper limits of the category $C_h$, $h\in Q$, respectively, and $b_0=0$, $b_q=1+\varepsilon$, $b_h<b_{h+1}$, $h=2,\ldots,q-1$. On this basis, the alternative $a_i$ is assigned to the category $C_h$ if $b_{h-1}\le U(a_i)< b_h$.

\section{Problem description and resolution framework}\label{sec:3}

We consider the MCS problem, which aims to assign a set of alternatives $A$ to several predefined ordered categories $C$ based on the set of criteria $G$. In particular, it is assumed that the decision maker progressively provides assignment example preference information, and he/she may exhibit potentially non-monotonic preferences over some criteria. For the sake of clarity, we denote the assignment example preference information in the $t$-th iteration as $S^t=\{a_i\rightarrow C_{B_i}|a_i\in A^{R,t}\}$, where $A^{R,t} \subset A$ is the set of reference alternatives.

\begin{figure}[htbp]
\centering
\includegraphics[scale=0.75]{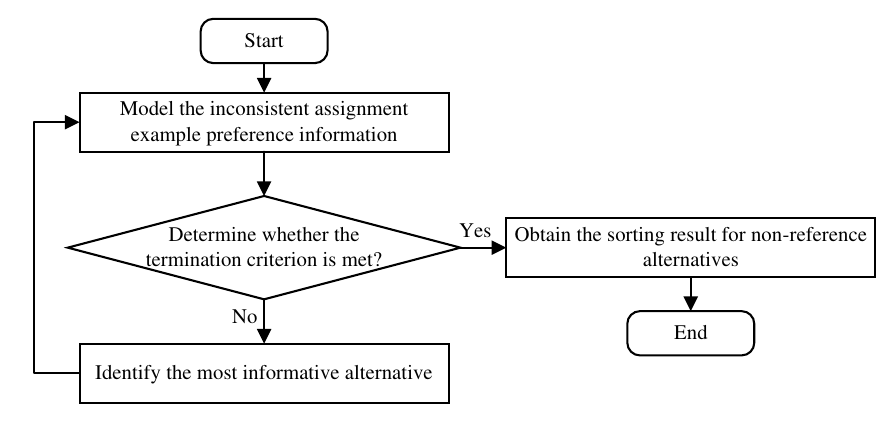}
\caption{A simple resolution framework of the proposed approach}
\label{fig:flowchart}
\end{figure}

The objective of this study is to develop an incremental preference elicitation-based approach to learn potentially non-monotonic preferences of the decision maker in MCS problems, capable of handling inconsistent assignment example preference information. To accomplish this, we present a simple resolution framework depicted in Fig. \ref{fig:flowchart}, and then provide a detailed explanation of the resolution framework as follows.

(1) Model the inconsistent assignment example preference information

We employ a max-margin optimization-based model to effectively handle the challenge of potentially non-monotonic preferences and inconsistent assignment example preference information in each iteration of the incremental preference elicitation process. This model guarantees a comprehensive representation of the decision maker's preferences by concurrently maximizing discriminative power and minimizing inconsistencies.

(2) Determine whether the termination criterion is met

Following the application of our developed max-margin optimization-based model,  we need to ascertain if the termination criterion is satisfied. Upon confirming the fulfillment of this criterion, the incremental preference elicitation process terminates, prompting the transition to yield the marginal utility functions and category thresholds, and then obtain the sorting result for non-reference alternatives. In case that the termination criterion is not met, progression to the next stage becomes necessary.

(3) Identify the most informative alternative

In this stage, we devise some question selection strategies aimed at identifying the most informative alternative within the framework of uncertainty sampling in active learning. Once the most informative alternative is pinpointed, the decision maker is prompted to provide an category assignment for it, thereby updating the set of assignment example preference information.

(4) Obtain the sorting result for non-reference alternatives

Upon meeting the termination criterion, the sorting result for non-reference alternatives can be determined by employing the max-margin optimization-based model and the complexity controlling optimization model. These models take into account all assignment example preference information provided by the decision maker throughout the incremental preference elicitation process to yield the marginal utility functions and category thresholds. On this basis, the sorting result for non-reference alternatives can be derived by the threshold-based MCS model.

\section{The proposed approach}\label{sec:4}
In this section, we elaborate on the proposed approach, providing comprehensive details. First, we present the inconsistent assignment example preference information modeling method in the incremental preference elicitation process. Following that, we introduce two termination criteria that are incorporated into our proposed approach and outline the design of various question selection strategies to identify the most informative alternative. Additionally, we demonstrate how to obtain the sorting result for non-reference alternatives. Lastly, we summarize the incremental preference elicitation-based approach.

\subsection{Inconsistent assignment example preference information modeling method}\label{sec:4.1}
{\color{black}
In each iteration of the incremental preference elicitation process, the decision maker may provide assignment example preference information that contradicts previously given information and may exhibit
non-monotonic preferences over some criteria simultaneously. Although existing studies have proposed various methods to address inconsistent assignment example preference information in MCS problems \citep{Kadzinski21kbs}, most of these methods are utilized within batch elicitation and are rarely applied in the context of incremental elicitation. While considering potentially non-monotonic preferences may reduce the possibility of inconsistencies in the assignment example preference information, it cannot entirely eliminate them, particularly for complex MCS problems (see the analysis in the supplemental file). Consequently, it becomes essential to take into account both inconsistency and potentially non-monotonic preferences during the incremental preference elicitation process.
}

To overcome the above challenges and motivated by the study of \cite{Teso16ijcai}, we propose a max-margin optimization-based model in this subsection, which aims to optimize two objectives. One is to maximize the discriminative power of the model, while the other is to be tolerant of inconsistent assignment example preference information. In what follows, we present the detailed modeling process.

Recall that $S^t=\{a_i\rightarrow C_{B_i}|a_i\in A^{R,t}\}$ is the assignment example preference information provided by the decision maker in the $t$-th iteration. As the decision maker may exhibit non-monotonic preferences in the MCS problem, {\color{black}inspired by \cite{Ghaderi17ejor},} we employ the following constraint set, i.e., $E^{NM}$, to model potentially non-monotonic preferences:
\begin{equation}
\resizebox{0.9\hsize}{!}{$
E^{NM}\begin{cases}
\begin{aligned}
& b_{B_{i}-1}\le U(a_{i})\le  b_{B_{i}} - \varepsilon, \text{ if } B_{i}\in\{2,\ldots,q-1\}, \forall a_{i}\in A^{R,t}\\
& U(a_{i}) \le b_1 - \varepsilon, \text{ if } B_{i}=1, \forall a_{i}\in A^{R,t}\\
& U(a_{i}) \ge b_{q-1}, \text{ if } B_{i}=q, \forall a_{i}\in A^{R,t}\\
&U(a_{i})=\sum\limits_{j=1}^mu_j(x_{ij}),\forall a_{i}\in A^{R,t} \\
&u_j(x_{ij})=u_j(\beta^l_j)+\frac{x_{ij}-\beta^l_j}{\beta^{l+1}_j-\beta^l_j}(u_j(\beta^{l+1}_j)-u_j(\beta^l_j)), \text{if}\ x_{ij}\in[\beta^l_j,\beta^{l+1}_j], \forall a_{i}\in A^{R,t}, j\in M \\
&b_{h} - b_{h-1} \ge \varepsilon, h=2,\ldots,q-1\\
& u_j(\beta_j^l)\ge 0, l=1,\ldots,s_j+1, j\in M\\
& u_j(\beta_j^l)\le 1, l=1,\ldots,s_j+1, j\in M,
\end{aligned}
\end{cases}$}
\end{equation}
where $\varepsilon$ is a small positive number.
\begin{remark}
The constraint set $E^{NM}$ can effectively capture potentially non-monotonic preferences, which can be explained from the following two aspects. On the one hand, the constraint set $E^{NM}$ does not set any limits on the difference between marginal utilities of two adjacent characteristic points. On the other hand, since the performance levels associated with the maximum and minimum marginal utilities taken for each criterion are not known beforehand, the normalization constraints required in the threshold-based MCS model (i.e., $u_j(\beta_j^1)=0$ and $\sum\nolimits_{j=1}^m u_j(\beta_j^{s_j+1})=1$) cannot be explicitly included in the constraint set $E^{NM}$. However, they can be guaranteed by utilizing the following two transformation functions for the marginal utilities and category thresholds, respectively:
\begin{equation}\label{eq:trandormation_function}
\begin{aligned}
&f_u(x_{ij}) = \frac{u_j(x_{ij})-u_j(g_j^-)}{\sum\limits_{j=1}^m(u_j(g_j^+) - u_j(g_j^-))}, i\in N, j\in M,\\
&f_b(b_h) = \frac{b_h - \sum\limits_{j=1}^mu_j(g_j^-)}{\sum\limits_{j=1}^m(u_j(g_j^+) - u_j(g_j^-))}, h\in Q,
\end{aligned}
\end{equation}
where $g_j^-$ and $g_j^+$ are the performance levels corresponding to the minimum and maximum marginal values of the criterion $g_j$, $j\in M$, respectively.
{\color{black}
In terms of Eq. \eqref{eq:trandormation_function}, for the marginal utility $u_j(x_{ij})$, if $x_{ij}=g_j^{-}$, we have that $f_u(x_{ij})=0$, and if $x_{ij}=g_j^+$, we have that $\sum\nolimits_{j=1}^m f_u(g_j^+) = \sum\nolimits_{j=1}^m \frac{u_j(g_j^+)-u_j(g_j^-)}{\sum\nolimits_{j=1}^m(u_j(g_j^+)-u_j(g_j^-))}=1$. Consequently, $f_u(x_{ij})$ can ensure that the normalization of the marginal utility.
For category thresholds, when $b_0=\sum\nolimits_{j=1}^m u_j(g_j^-)$, we obtain that $f_b(b_0)=0$, and when $b_q=\sum\nolimits_{j=1}^mu_j(g_j^+) + \varepsilon$, we obtain that $f_b(b_q)=1+\frac{\varepsilon}{\sum\nolimits_{j=1}^m(u_j(g_j^+)-u_j(g_j^-))}$. Let $\varepsilon^S=\frac{\varepsilon}{\sum\nolimits_{j=1}^m(u_j(g_j^+)-u_j(g_j^-))}$, we can derive that $f_b(b_q)=1+\varepsilon^S$. As a result, $f_b(b_h)$ can achieve the normalization of category thresholds.
}

\end{remark}

As the discriminative power of the sorting model can be measured by the value of the parameter $\varepsilon$ in the constraint set $E^{NM}$, the first objective of the max-margin optimization-based model is formulated as
\begin{equation}
\max \varepsilon.
\end{equation}

Furthermore, to tolerate the inconsistent assignment example preference information provided by the decision maker, we introduce some auxiliary variables $\delta^+_i$ and $\delta^-_i$, $\forall a_i\in A^{R,t}$ that are usually utilized in the UTA approach \citep{Ghaderi21omega}. Following this, we use $ICI^t$ to quantify the extent of inconsistency in the assignment example preference information provided by the decision maker in the $t$-th iteration, i.e.,
\begin{equation}
ICI^t = \sum\limits_{a_i\in A^{R,t}} \delta^+_i + \delta^-_i,
\end{equation}
where $\delta^+_i, \delta^-_i\ge 0$.

On this basis, the second objective of the max-margin optimization-based model is formulated as
\begin{equation}
\min \sum\limits_{a_i\in A^{R,t}} \delta^+_i + \delta^-_i.
\end{equation}

Taking into account both objectives, the final objective function of the max-margin optimization-based model can be formulated as
{\color{black}
\begin{equation}
\max \ \alpha \varepsilon - (1-\alpha) \frac{\sum\nolimits_{a_i\in A^{R,t}}\delta^+_i + \delta^-_i}{|A^{R,t}|},
\end{equation}
where $\alpha\in(0,1)$ is utilized to make a trade-off between the two objectives, and $|A^{R,t}|$ denotes the number of reference alternatives in $A^{R,t}$.

\begin{remark}
In practice, there are two alternative options for determining the value of $\alpha$. On the one hand, it can be directly assigned by the decision analyst during the incremental preference elicitation process, based on his/her knowledge and experience. On the other hand, it can be determined based on the model's performance, for instance, by selecting the $\alpha$ value that yields the best performance on test sets.
\end{remark}

}

Building upon the aforementioned analysis, the max-margin optimization-based model can be formulated as
{\color{black}
\begin{small}
\begin{equation}\tag{M-1}\label{m:max_margin}
\begin{aligned}
&\max \ \alpha \varepsilon - (1 - \alpha) \frac{\sum\nolimits_{a_i\in A^{R,t}}\delta^+_i + \delta^-_i}{|A^{R,t}|} \\
&\begin{aligned}
\rm{s.t.} \ & b_{B_{i}-1} - \delta^+_i \le U(a_{i})\le  b_{B_{i}} - \varepsilon + \delta^-_i, \text{ if } B_{i}\in\{2,\ldots,q-1\}, \forall a_{i}\in A^{R,t}\\
& U(a_{i}) \le b_1 - \varepsilon + \delta^-_i, \text{ if } B_{i}=1, \forall a_{i}\in A^{R,t}\\
& U(a_{i}) \ge b_{q-1} - \delta^+_i, \text{ if } B_{i}=q, \forall a_{i}\in A^{R,t}\\
&U(a_{i})=\sum\nolimits_{j=1}^mu_j(x_{ij}),\forall a_{i}\in A^{R,t} \\
&u_j(x_{ij})=u_j(\beta^l_j)+\frac{x_{ij}-\beta^l_j}{\beta^{l+1}_j-\beta^l_j}(u_j(\beta^{l+1}_j)-u_j(\beta^l_j)), \text{if}\ x_{ij}\in[\beta^l_j,\beta^{l+1}_j], \forall a_{i}\in A^{R,t}, j\in M \\
&b_{h} - b_{h-1} \ge \varepsilon, h=2,\ldots,q-1\\
& u_j(\beta_j^l)\ge 0, l=1,\ldots,s_j+1, j\in M\\
& u_j(\beta_j^l)\le 1, l=1,\ldots,s_j+1, j\in M \\
& \delta^+_i, \delta^-_i\ge 0, \forall a_i \in A^{R,t}\\
& 0< \varepsilon \le \frac{m}{q-1}.
\end{aligned}
\end{aligned}
\end{equation}
\end{small}
}

\begin{remark}
In particular, the last constraint in the model \eqref{m:max_margin} is set to prevent the model from being unbounded. The reason why we set the range of $\varepsilon$ to $(0,\frac{m}{q-1}]$ is illustrated as follows. Let $b_h^S$, $h\in Q$ be category thresholds in a UTA-like standard form, they should satisfy some conditions, i.e., $b_0=0$, $b_q=1+\varepsilon^S$, and $b_{h}^S - b_{h-1}^S \ge \varepsilon^S$, $h\in Q$, where $\varepsilon^S$ is an arbitrarily small positive number. By summing both sides of these $q$ inequalities, we can obtain that $\sum\nolimits_{h=1}^q (b_{h}^S - b_{h-1}^S)=b_q^S-b_0^S=1+\varepsilon^S\ge q\cdot \varepsilon^S$, thus we have that $\varepsilon^S\le \frac{1}{q-1}$. When considering non-monotonic preferences in the MCS problem, we have that $\varepsilon^S=\frac{\varepsilon}{\sum\nolimits_{j=1}^m(u_j(g_j^+)-u_j(g_j^-))}$. Since $\varepsilon^S\le \frac{1}{q-1}$, we have that $\frac{\varepsilon}{\sum\nolimits_{j=1}^m(u_j(g_j^+)-u_j(g_j^-))}\le \frac{1}{q-1}\iff \varepsilon\le \frac{1}{q-1}\sum\nolimits_{j=1}^m(u_j(g_j^+)-u_j(g_j^-))$. Furthermore, as $0 \le u_j(\beta_j^l)\le 1$, the maximum value of $u_j(g_j^+)-u_j(g_j^-)$ is 1. As a result, we can conclude that $\varepsilon\le \frac{m}{q-1}$. In addition, $\varepsilon^S>0$ ensures that $\varepsilon>0$. To sum up, we set the range of $\varepsilon$ to $(0,\frac{m}{q-1}]$.
\end{remark}

If the optimal solutions to the model \eqref{m:max_margin}, i.e., $\delta^{+,*}_i$, $\delta^{-,*}_i$, $\forall a_i\in A^{R,t}$, are all equal to 0, the assignment example preference information provided by the decision maker is consistent. Otherwise, it is inconsistent. The model maximizes the discriminative power while tolerating inconsistent assignment example preference information through the introduction of some auxiliary variables, effectively modeling the decision maker's inconsistent assignment example preference information in each iteration of the incremental preference elicitation process.

\subsection{Termination criteria}\label{sec:4.2}
The incremental preference elicitation process will come to an end once a termination criterion is satisfied. In this regard, we introduce two different termination criteria that can be utilized throughout the incremental preference elicitation process.

\textbf{Termination criterion \RNum{1}}

Termination criterion \RNum{1} assumes that the decision maker has a predetermined limit on the number of questions that he/she can answer during the incremental preference elicitation process. For convenience, let the maximum number of questions that can be answered by the decision maker be denoted as $T$, and the number of questions already answered by the decision maker be $t$. If $t=T$, the incremental preference elicitation process will be terminated. Otherwise, the incremental preference elicitation process is continued until the number of questions already answered by the decision maker is equal to $T$.

\textbf{Termination criterion \RNum{2}}

Termination criterion \RNum{2} makes an assumption that the incremental preference elicitation process should be terminated only after certain performance requirements are fulfilled. In this scenario, we employ the accuracy metric to evaluate the performance of the proposed approach.
\begin{definition}
Let $f_i\in C$ and $\overline{f_i}\in C$ be the real and inferred category assignment of the $i$-th non-reference alternative in the $t$-th iteration, $a_i\in A \textbackslash A^{R,t}$, then the accuracy metric, denoted as $Acc^t$, is computed by
\begin{equation}\label{eq:accuracy}
Acc^t = \frac{\sum\nolimits_{i=1}^{nr^t} y_i}{nr^t},
\end{equation}
where $nr^t$ is the number of non-reference alternatives in the set $A \textbackslash A^{R,t}$. In addition, $y_i$ is a binary variable, if $f_i=\overline{f_i}$, then $y_i=1$; otherwise, $y_i=0$.
\end{definition}

On this basis, let $Acc_{target}$ represent the target accuracy that the decision analyst aims to achieve throughout the incremental preference elicitation process. If $Acc^t\ge Acc_{target}$, the incremental preference elicitation process can be terminated. Otherwise, the incremental preference elicitation process needs to continue.

\subsection{Question selection strategies}\label{sec:4.3}
In each iteration of the incremental preference elicitation process of the MCS problem, the decision analyst will pose a question to the decision maker. This question can be formulated as ``which category can the alternative $a_i$ be assigned to?'' For the decision analyst, he/she should decide which alternative to ask about. To accomplish this objective, we initially design some metrics to measure the information amount associated with each alternative. Building upon this foundation, we provide some question selection strategies to identify the most informative alternative.

\subsubsection{Information amount measurement methods}\label{sec:4.3.1}

In this section, we present the metrics utilized to measure the information amount associated with each alternative in the following.

For each alternative $a_i\in A \textbackslash A^{R,t}$, we add $a_i \rightarrow C_h$, $h\in Q$ to $S^t$ separately, and solve the model \eqref{m:max_margin} $q$ times. Let $m_{ih}^*$ be the optimal objective function value when the assignment example preference information is $S^t\cup\{a_i\rightarrow C_h\}$, and $\bm{v_i}=(m_{i1}^*,\ldots,m_{iq}^*)^{\textrm T}$ be a vector composed of $m_{ih}^*$, $h\in Q$, then we can define the information amount of each alternative $a_i\in A \textbackslash A^{R,t}$ in terms of $\bm{v_i}$.

\textbf{Sum margin-based measure.} To aggregate the vector $\bm{v_i}$ into a comparable single value, sum margin-based measure is adopted \citep{Nefla19adt}. For convenience, let $IA_i^{SM}$ be the information amount of the alternative $a_i$ under the sum margin-based measure. This value can be conveniently calculated by summing the elements in the vector $\bm{v_i}$, i,e,
\begin{equation}\label{eq:IA_sum}
IA_i^{SM} = \sum\nolimits_{h=1}^q m_{ih}^*, \forall a_i \in A \textbackslash A^{R,t}.
\end{equation}

{\color{black}
Furthermore, given that the elements in $\bm{v_i}$ can reflect the probability that the alternative $a_i$ is assigned to each category to some extent, we incorporate the idea of uncertainty sampling in active learning into our approach \citep{Aggarwal14dc}. Uncertainty sampling in active learning refers to that the alternative with the highest uncertainty in sorting result are likely to be the most challenging and can provide the most information. This uncertainty can be quantified by the probability of the alternative being assigned to each category.
Building upon these points, we can use $\bm{v_i}$ to analyze the probability of the alternative being assigned into each category. This allows us to measure the uncertainty inherent in sorting result for each alternative and subsequently calculate the information amount associated with each alternative. To this end, we first provide the probability-based measure and then present several metrics utilized in the framework of uncertainty sampling in active learning.
}

\textbf{Probability-based measure.}
{\color{black}The larger the value of $m_{ih}^*$, the larger the value of $\varepsilon$, the smaller the extent of inconsistencies, and the more likely the alternative $a_i$ is assigned to the category $C_h$. Similarly, the smaller the value of $m_{ih}^*$, the less likely the alternative $a_i$ is assigned to the category $C_h$. To do so, we introduce several ways to transform the elements in $\bm{v_i}$ into the probability that the alternative $a_i$ is assigned to each category. }
In general, such a transformation should satisfy the following two properties:

(1) The probability of assigning each alternative $a_i$ to each category $C_h$ should be greater than zero, and the sum of the probabilities assigned to all categories should be equal to one.

(2) The probability of assigning each alternative $a_i$ to each category $C_h$ should be in line with the value of $m_{ih}^*$, i.e., the larger the value of $m_{ih}^*$, the higher the probability of assigning the alternative $a_i$ to the category $C_h$.

Without loss of generality, we devise two types of transformations to quantifying the probability of assigning each alternative $a_i$ to each category $C_h$ as follows.

$\bullet$ \textbf{Relu function-based transformation.} Let $p_{ih}$ be the transformed probability of the alternative $a_i$ is assigned to the category $C_h$, then we have
\begin{equation}\label{eq:relu}
p_{ih} = \frac{\text{Relu}(m_{ih}^*)}{\sum\nolimits_{h=1}^q \text{Relu}(m_{ih}^*)}, \forall a_i \in A \textbackslash A^{R,t}, h\in Q,
\end{equation}
where Relu$(\cdot)$ is the Rectified Linear Unit function. It is a non-linear function that sets negative values to zero while keeping positive values unchanged, namely, $\text{Relu}(x)=\max(x,0)$.

$\bullet$ \textbf{Softmax function-based transformation.} Let $p_{ih}$ be the transformed probability of the alternative $a_i$ is assigned to the category $C_h$. Utilizing the softmax function-based transformation, $p_{ih}$ is calculated by
\begin{equation}\label{eq:softmax}
p_{ih}=\frac{e^{m_{ih}^*}}{\sum\nolimits_{h=1}^q e^{m_{ih}^*}}, \forall a_i \in A \textbackslash A^{R,t}, h\in Q.
\end{equation}

{\color{black}
\begin{remark}
The Relu function has the capability to convert negative values to zero. In our proposed approach, when the value of $m_{ih}^*$ is negative, it indicates that the extent of inconsistencies is significantly greater than the discriminative power of the model. In such cases, the probability of the alternative $a_i$ being assigned to the category $C_h$ is very low. By using the Relu function to set negative values to zero, we can eliminate the influence of these values, thereby allowing us to focus more on the probability of the alternative being assigned to other categories.
Furthermore, the Softmax function provides a natural way to convert the elements of a vector into probability values, ensuring that the sum of all probabilities is equal to 1. It also reduces the impact of extreme values on the normalization results by using the exponential function, offering numerical stability.
In fact, numerous transformation methods can fulfill the above two properties in the probability-based measure. Using  different transformation methods will not alter the essence of the proposed approach.
\end{remark}
}

According to the transformed probability, we define three metrics for measuring the information amount of each alternative adhering to the framework of uncertainty sampling in active learning \citep{Aggarwal14dc}.

$\bullet$ \textbf{Entropy-based metric.} The information amount $IA_i$ associated with the alternative $a_i$ can be defined in light of entropy, which is commonly utilized to measure uncertainty. The higher the uncertainty of the alternative $a_i$, the greater the information amount. As a result, we can calculate $IA_i^{E}$ utilizing the following equation:
\begin{equation}\label{eq:IA_entropy}
IA_i^{E} = - \sum\limits_{h\in Q,p_{ih}>0} p_{ih}\log p_{ih}, \forall a_i \in A \textbackslash A^{R,t}.
\end{equation}

$\bullet$ \textbf{Least confidence-based metric.}
{\color{black}The least confidence-based metric focuses on the alternative with the highest model prediction probability assigned to different categories but lower confidence, and it believes that such alternative is more difficult to distinguish.} To do so, let $p_i^{max}$ be the highest probability assigned to different categories for the alternative $a_i$, then we have that
\begin{equation}
p_i^{max} = \max (p_{i1}^*,\ldots,p_{iq}^*),\forall a_i \in A \textbackslash A^{R,t}.
\end{equation}

According to the idea of the least confidence-based metric, the alternative with a smaller $p_i^{max}$ value has a greater uncertainty and provides more information. Therefore, the information amount associated with the alternative $a_i$ when considering the least confidence-metric can be defined as follows.
\begin{equation}
IA_i^L = 1 - p_i^{max},\forall a_i \in A \textbackslash A^{R,t}.
\end{equation}


$\bullet$ \textbf{Margin of confidence-based metric.}
{\color{black}In light of the margin of confidence-based metric, the alternative with similar probabilities of being assigned into two categories has greater uncertainty. To be specific, }the uncertainty for each alternative $a_i$ is reflected in the calculation of the difference between the highest and the second highest probabilities assigned by the alternative $a_i$ to different categories when utilizing margin of confidence-based metric. The smaller the difference, the greater the uncertainty, and the higher the information amount associated with the alternative $a_i$. For convenience, let $p_i^{max1}$ and $p_i^{max2}$ be the highest and the second highest probability assigned to different categories for the alternative $a_i$, respectively, then the information amount associated with the alternative $a_i$ can be defined as
\begin{equation}\label{eq:IA_M}
IA_i^M = p_i^{max2} - p_i^{max1}, \forall a_i \in A \textbackslash A^{R,t}.
\end{equation}

{\color{black}
\begin{remark}
The three metrics mentioned above are derived from the uncertainty sampling framework in active learning, with the main difference among them being the method of measuring uncertainty \citep{Aggarwal14dc}. The entropy-based metric considers all probabilities of an alternative being assigned to different categories, the least confident metric only considers the highest probability of an alternative being assigned to different categories, while the margin of confidence-based metric takes into account both the highest and the second highest probability of an alternative being assigned to different categories.
\end{remark}

}

\subsubsection{Identifying the most informative alternative}\label{sec:4.3.2}
Based on the information amount metrics defined in Section \ref{sec:4.3.1}, some specific question selection strategies are proposed to identify the most informative alternative in the incremental preference elicitation process for MCS problems, which are based on the principle that the higher the uncertainty associated with an alternative, the more the information amount of this alternative.

First, let $a_{SM}$ be the most informative alternative in the $t$-th iteration identified by the sum margin-based strategy (denoted as $SM$), then it can be derived by
\begin{equation}\label{eq:SM}
a_{SM} = \arg \max_{a_i\in A \textbackslash A^{R,t}}IA_i^{SM}.
\end{equation}

Following this, in the context of the probability-based measure, we present two types of transformation ways, namely, the Relu function-based transformation and the Softmax function-based transformation. Building upon these transformation ways, three distinct information amount metrics are utilized, namely, entropy-based metric, least confidence-based metric and margin of confidence-based metric. By combining these two probability transformations with three different information amount metrics, six question selection strategies are obtained to identify the alternative with the most information amount.

{\color{black}
For convenience, let $X\in \{E,L,M\}$ represent the three metrics quantifying the information amount within the framework of uncertainty sampling in active learning, and $Y\in\{R,S\}$ signify the two transformation methods employed in the probability-based measure. Subsequently, we denote $IA_i^{XY}$ as the information amount associated with the alternative $a_i$ when using the $XY$ strategy. Furthermore, let $a_{XY}$ be the most informative alternative identified by the $XY$ strategy during the $t$-th iteration. On this basis, $a_{XY}$ can be consistently determined for each question selection strategy by
\begin{equation}\label{eq:XY}
a_{XY} = \arg \max_{a_i\in A \textbackslash A^{R,t}}IA_i^{XY}, X\in \{E,L,M\}, Y\in\{R,S\}.
\end{equation}

To enhance comprehension the above equation, we elucidate the six question selection strategies as follows.

$\bullet$ $ER$ strategy: the Relu function-based transformation and the entropy-based metric are combined.

$\bullet$ $ES$ strategy: the Softmax function-based transformation and the entropy-based metric are combined.

$\bullet$ $LR$ strategy: the Relu function-based transformation and the least confidence-based metric are combined.

$\bullet$ $LS$ strategy: the Softmax function-based transformation and the least confidence-based metric are combined.

$\bullet$ $MR$ strategy: the Relu function-based transformation and the margin of confidence-based metric are combined.

$\bullet$ $MS$ strategy: the Softmax function-based transformation and the margin of confidence-based metric are combined.

}

\begin{remark}
In practice, the decision analyst has the flexibility to select an appropriate question selection strategy based on his/her knowledge and experience. Additionally, the decision analyst can also choose the most suitable question selection strategy in terms of the performance of different strategies in a specific MCS problem.
\end{remark}

{\color{black}
\subsection{Obtaining the sorting result for non-reference alternatives}
It is important to note that the model \eqref{m:max_margin} does not take into account the model complexity, which may deteriorate interpretability of the sorting results and increase the risk of overfitting. To do so, to determine the final sorting result for alternatives, we develop an additional model to control the model complexity on the basis of the model \eqref{m:max_margin}.

The model complexity can be controlled by managing the slope change of the piecewise linear marginal utility functions. The slope change between two consecutive subintervals $[\beta_j^{l-1},\beta_j^l)$ and $[\beta_j^l, \beta_j^{l+1})$ for the criterion $g_j$'s marginal utility function can be represented as $|\sfrac{(u_j(\beta_j^{l+1})-u_j(\beta_j^l))}{(\beta_j^{l+1} - \beta_j^l)} - \sfrac{(u_j(\beta_j^l) - u_j(\beta_j^{l-1}))}{(\beta_j^l - \beta_j^{l-1})} |$, $l=2,\ldots,s_j,j\in M$. Consequently, the objective of the complexity controlling optimization model is formulated as
\begin{equation}
\min \ \sum\limits_{j=1}^m \sum\limits_{l=2}^{s_j}\left|\frac{u_j(\beta_j^{l+1})-u_j(\beta_j^l)}{\beta_j^{l+1} - \beta_j^l} - \frac{u_j(\beta_j^l) - u_j(\beta_j^{l-1})}{\beta_j^l - \beta_j^{l-1}} \right|.
\end{equation}

Furthermore, both the optimal objective function value and constraints of the model \eqref{m:max_margin} should be guaranteed. Let $J^*$ be the optimal objective function value of the model \eqref{m:max_margin}, and $E^{MM}$ be the constraint set of the model \eqref{m:max_margin}, then the complexity controlling optimization model can be constructed as
\begin{equation}\tag{M-2}\label{m:min_slope0}
\begin{aligned}
&\min \ \sum\limits_{j=1}^m \sum\limits_{l=2}^{s_j}\left|\frac{u_j(\beta_j^{l+1})-u_j(\beta_j^l)}{\beta_j^{l+1} - \beta_j^l} - \frac{u_j(\beta_j^l) - u_j(\beta_j^{l-1})}{\beta_j^l - \beta_j^{l-1}} \right| \\
&\begin{aligned}
\rm{s.t.} \ &  J^* = \alpha \varepsilon - (1 - \alpha) \frac{\sum\nolimits_{a_i\in A^{R,t}}\delta^+_i + \delta^-_i}{|A^{R,t}|}\\
& E^{MM}.
\end{aligned}
\end{aligned}
\end{equation}

\begin{theorem}
Let $\gamma_{jl}=\left|\frac{u_j(\beta_j^{l+1})-u_j(\beta_j^l)}{\beta_j^{l+1} - \beta_j^l} - \frac{u_j(\beta_j^l) - u_j(\beta_j^{l-1})}{\beta_j^l - \beta_j^{l-1}} \right|$, the model \eqref{m:min_slope0} can be converted into the following linear programming model,
\begin{equation}\tag{M-3}\label{m:min_slope}
\begin{aligned}
&\min \ \sum\limits_{j=1}^m \sum\limits_{l=2}^{s_j} \gamma_{jl} \\
&\begin{aligned}
\rm{s.t.} \ & \gamma_{jl} \ge \frac{u_j(\beta_j^{l+1})-u_j(\beta_j^l)}{\beta_j^{l+1} - \beta_j^l} - \frac{u_j(\beta_j^l) - u_j(\beta_j^{l-1})}{\beta_j^l - \beta_j^{l-1}}, l=2,\ldots,s_j, j\in M\\
& \gamma_{jl} \ge -\frac{u_j(\beta_j^{l+1})-u_j(\beta_j^l)}{\beta_j^{l+1} - \beta_j^l} + \frac{u_j(\beta_j^l) - u_j(\beta_j^{l-1})}{\beta_j^l - \beta_j^{l-1}}, l=2,\ldots,s_j, j\in M\\
& J^* = \alpha \varepsilon - (1 - \alpha) \frac{\sum\nolimits_{a_i\in A^{R,t}}\delta^+_i + \delta^-_i}{|A^{R,t}|}\\
& E^{MM}.
\end{aligned}
\end{aligned}
\end{equation}
\end{theorem}

Following the above analysis, to determine the sorting result for non-reference alternatives, it is necessary to sequentially employ the two optimization models (i.e., the models \eqref{m:max_margin} and \eqref{m:min_slope}), in which all assignment example preference information provided by the decision maker throughout the incremental preference elicitation process should be considered. Given marginal utility functions and category thresholds derived from the models \eqref{m:max_margin} and \eqref{m:min_slope}, the sorting result for non-reference alternatives can be obtained by the threshold-based MCS model.

}

\subsection{Incremental preference elicitation-based algorithms}\label{sec:4.4}
\begin{figure}[H]
\centering
\includegraphics[scale=0.7]{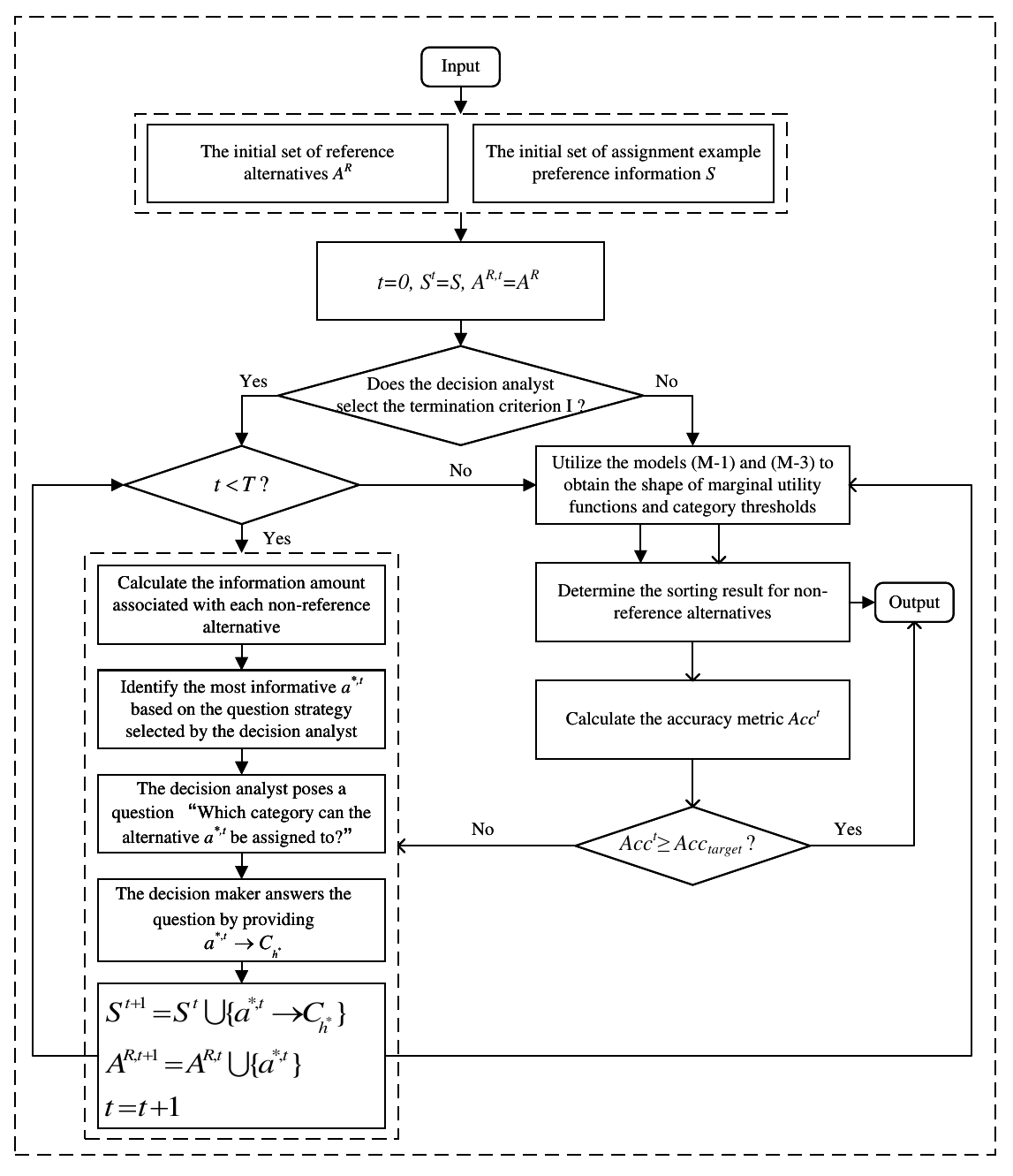}
\caption{The flowchart of the proposed incremental preference elicitation-based approach}
\label{fig:df}
\end{figure}

To provide a comprehensive summary of the proposed approach and enhance understanding of the overall incremental preference elicitation-based approach, this subsection presents two algorithms considering two different termination criteria (see Algorithms \ref{alg:1} and \ref{alg:2}) and a detailed flowchart of the proposed approach (see Fig. \ref{fig:df}).
\begin{algorithm}
\begin{small}
\caption{The incremental preference elicitation-based algorithm (termination criterion \RNum{1})}\label{alg:1}
\begin{algorithmic}[1]
     \Require  The decision matrix $X=(x_{ij})_{n\times m}$, the number of subinterval for each criterion $s_j$, $j\in M$, the initial assignment example preference information $S=\{a_i\rightarrow C_{B_i}|a_i\in A^R\}$, and the maximum number of questions the decision maker can answer $T$.
     \Ensure The marginal utility functions of each criteria, the category thresholds and the sorting result for non-reference alternatives.
        \end{algorithmic}
\begin{enumerate}[\bf Step 1:]
  \item Let $t=0$, $S^t=S$, and $A^{R,t}=A^R$.
  \item If $t<T$, proceed to the next step. Otherwise, proceed directly to Step 9.
  \item For each alternative $a_i\in A\textbackslash A^{R,t}$ and category $C_h\in C$, let $S^{'}=S^t\cup \{a_i\rightarrow C_h\}$, and then solve the model \eqref{m:max_margin} in terms of $S^{'}$. Denote the optimal objective function value as $m_{ih}^*$, $\forall a_i\in A\textbackslash A^{R,t}$, $h\in Q$ and let $\bm{v_i}=(m_{i1}^*,\ldots,m_{iq}^*)^{\rm T}$.
  \item The decision analyst chooses one strategy from the proposed seven question selection strategies, i.e., $SM$, $ER$, $ES$, $LR$, $LS$, $MR$ and $MS$.
  \item In light of the selected strategy and $\bm{v_i}$, identify the most informative alternative (denoted as $a^{*,t}$) by Eqs. \eqref{eq:SM}~-~\eqref{eq:XY}.
  \item The decision analyst is required to pose a question to the decision maker, i.e., ``which category can the alternative $a^{*,t}$ be assigned to?''
  \item The decision maker answers the question posed by the decision analyst, i.e., the alternative $a^{*,t}$ should be assigned to the category $C_{h^*}$.
  \item Update $S^{t+1}$ and $A^{R,{t+1}}$ as $S^t\cup \{a^{*,t}\rightarrow  C_{h^*}\}$ and $A^{R,t}\cup \{a^{*,t}\}$, respectively. Let $t=t+1$, return to Step 2.
  \item Based on all assignment example preference information provided by the decision maker, i.e., $S^T$, solve the models \eqref{m:max_margin} and \eqref{m:min_slope} to obtain marginal utility function $u_j(\beta_j^l)$, $l=1,\ldots,s_j$, $j\in M$, and category thresholds $b_h$, $h\in Q$. Subsequently, determine the sorting result for non-reference alternatives ($a_i\in A\textbackslash A^{R,T}$) by utilizing the threshold-based MCS model described in Section \ref{sec:2}.
  \item Output the marginal utility functions of each criteria, the category thresholds and the sorting result for non-reference alternatives.
\end{enumerate}
\end{small}
\end{algorithm}

\begin{algorithm}
\begin{small}
\caption{The incremental preference elicitation-based algorithm (termination criterion \RNum{2})}\label{alg:2}
\begin{algorithmic}[1]
     \Require  The decision matrix $X=(x_{ij})_{n\times m}$, the number of subinterval for each criterion $s_j$, $j\in M$, the initial assignment example preference information $S=\{a_i\rightarrow C_{B_i}|a_i\in A^R\}$, and the target accuracy that the decision analyst aims to achieve $Acc_{target}$.
     \Ensure The sorting result for non-reference alternatives.
        \end{algorithmic}
\begin{enumerate}[\bf Step 1:]
  \item The same as Algorithm \ref{alg:1}.
  \item Based on $S^t$, solve the models \eqref{m:max_margin} and \eqref{m:min_slope} to obtain marginal utility functions $u_j(\beta_j^l)$, $l=1,\ldots,s_j$, $j\in M$, and category thresholds $b_h$, $h\in Q$.
  \item Determine the sorting result for non-reference alternatives ($a_i\in A\textbackslash A^{R^t}$) by utilizing the threshold-based MCS model described in Section \ref{sec:2}.
  \item In light of the real and inferred sorting results for non-reference alternatives, calculate the accuracy metric (denoted as $Acc^t$) by Eq. \eqref{eq:accuracy}.
  \item If $Acc^t\ge Acc_{target}$, go to Step 7. Otherwise, go to the next step.
  \item Steps 3~-~8 in Algorithm \ref{alg:1}.
  \item Output the marginal utility functions of each criteria, the category thresholds and the sorting result for non-reference alternatives.
\end{enumerate}
\end{small}
    \end{algorithm}

\section{Numerical results}\label{sec:5}
This section begins by an illustrative example using a credit rating example to elucidate the implementation process of our proposed approach. On this basis, we perform some further discussions.

\subsection{Illustrative example}\label{sec:5.1}
We illustrate the incremental preference elicitation-based approach by using the $ES$ strategy. To this end, we consider a real-world example utilizing the data from Section 4 of \cite{Guo19eswa}, specifically focusing on credit rating. It is noteworthy that the majority of the data predominantly stems from studies \cite{Despotis95ama} and \cite{Ghaderi17ejor} initially.
In this example, twenty firms (denoted as $A=\{a_1,a_2,\ldots,a_{20}\}$) need to be assigned to four ordered categories, i.e., firms in the worst financial state ($C_1$), firms in a lower-intermediate financial state ($C_2$), firms in an upper-intermediate financial state ($C_3$), and firms in the best financial state ($C_4$). The firms are evaluated in terms of the following three criteria: the cash to total assets ($g_1$), the long term debt and stockholder's equity to fixed assets ($g_2$), and the total liabilities to total assets ($g_3$). The performance levels of the twenty firms are presented as
\begin{footnotesize}
\[\setlength{\arraycolsep}{1.5pt}
X= \left(
  \begin{array}{cccccccccccccccccccc}
  3.8 & 5.84 & 0.04 & 4.89 & 0.57 & 16.7 & 3.16 & 25.42 & 17.99 & 3.98 & 0.76 & 24.16 & 2.53 & 35.06 & 0.72 & 24 & 8.86 & 10.58 & 16.35 & 1.7\\
  2.4 & 1.96 & 1.14 & 2.92 & 1.72 & 2.32 & 4.1 & 3.35 & 1.34 & 3.26 & 2.74 & 2.83 & 2.54 & 9.56 & 0.97 & 2.5 & 29.06 & 4.03 & 3.6 & 5.92\\
  60.7 & 63.7 & 64.26 & 55.04 & 64.7 & 53.29 & 23.9 & 59.03 & 73.84 & 84.95 & 84.44 & 70.51 & 81.05 & 61.08 & 99.67 & 99.92 & 47.4 & 89.64 & 56.55 & 85.83\\
  \end{array}\right)^{\rm T}.\]
\end{footnotesize}

To elucidate the incremental preference elicitation process, we simulate the roles of a decision analyst and a decision maker. {\color{black}The answers provided by the decision maker in the incremental preference elicitation process are predetermined using the following procedure: we begin by simulating marginal utility functions and category thresholds to obtain the initial sorting result for the twenty firms. Subsequently, by introducing some random noise into the initial sorting result, the final predetermined sorting result for the twenty firms is obtained as $F=(C_2, C_4, C_2, C_3, C_1, C_4, C_1, C_4, C_2, C_3, C_4, C_3, C_1, C_4, C_3, C_4, C_4, C_2, C_3, C_1)^{\rm T}$.}

In what follows, we utilize Algorithm \ref{alg:1} to illustrate the specific implementation process of the proposed approach. First, the decision analyst requires the decision maker to provide some assignment example preference information in light of his/her experience. Then the decision maker provide his/her initial assignment example preference information as $S=\{a_3\rightarrow C_2, a_{12}\rightarrow C_3, a_{16}\rightarrow C_4, a_{20}\rightarrow C_1\}$. Afterwards, the decision analyst chooses to utilize termination criterion \RNum{1} and the decision maker provides the maximum number of questions that he/she can answer as $T=8$. Additionally, the number of subintervals for all criteria $s_j$ is set to 4, $j=1,2,3$, and the value of $\alpha$ in the model \eqref{m:max_margin} is set to 0.1 in terms of the experience of the decision analyst.

Let $t=0$, $S^0=S$, $A^{R,0}=\{a_3,a_{12}, a_{16}, a_{20}\}$. As $t<T=8$, we need to identify the most informative alternative from the set $A\textbackslash A^{R,0}=\{a_1,a_2,a_4,a_5,a_6,a_7,a_8,a_9,a_{10},a_{11},a_{13},a_{14},a_{15},a_{17},a_{18},a_{19}\}$.
For each alternative $a_i\in A\textbackslash A^{R,0}$ and category $C_h\in \{C_1,C_2,C_3,C_4\}$, let $S^{'}=S^0\cup \{a_i\rightarrow C_h\}$. Solve the model \eqref{m:max_margin} yields the optimal objective function value $m_{ih}^*$, and we further obtain $\bm{v_i}=(m_{i1}^*,\ldots,m_{iq}^*)^{\rm T}$. The results are displayed in the second column of Table \ref{table:v_0}.
\begin{small}
\begin{table}[htbp]
\centering
\setlength{\abovecaptionskip}{0pt}
\setlength{\belowcaptionskip}{10pt}
\caption{The values of $\bm{v_i}$ and the information amount of alternatives in  $A\textbackslash A^{R,0}$}
\label{table:v_0}
\scalebox{0.95}{
\begin{tabular}{ccc}
\toprule
$a_i$ & $\bm{v_i}$ & $IA_{i}$    \\ \midrule
$a_1$ & $(0.0433,0.0671,0.0509,0.0299)^{\rm T}$ & 1.386204 \\
$a_2$ & $(0.0423,0.0675,0.0613,0.0354)^{\rm T}$ & 1.386208 \\
$a_4$ & $(0.0553, 0.0684,0.0596,0.0357)^{\rm T}$ & 1.386223 \\
$a_5$ & $(0.0124,0.0658,0.0089,0.0048)^{\rm T}$ & 1.385979 \\
$a_6$ & $(0.0482,0.0592,0.0684,0.0589)^{\rm T}$ & 1.386269 \\
$a_7$ & $(0.0684,0.0684,0.0614,0.0439)^{\rm T}$ & 1.386245 \\
$a_8$ & $(0.0254,0.0381,0.0684,0.0518)^{\rm T}$ & 1.386166 \\
$a_9$ & $(0.0455,0.0548,0.0672,0.0499)^{\rm T}$ & 1.386261 \\
$a_{10}$ & $(0.0616,0.0605,0.0316,0.0213)^{\rm T}$ & 1.386138 \\
$a_{11}$ & $(0.0611,0.0470,0.0270,0.0189)^{\rm T}$ & 1.386156 \\
$a_{13}$ & $(0.0663,0.0460,0.0276,0.0197)^{\rm T}$ & 1.386132 \\
$a_{14}$ & $(0.0684,0.0684,0.0669,0.0573)^{\rm T}$ & 1.386284 \\
$a_{15}$ & $(0.0190,0.0509,0.0675,0.0488)^{\rm T}$ & 1.386142 \\
$a_{17}$ & $(0.0684,0.0684,0.0682,0.0645)^{\rm T}$ & 1.386293 \\
$a_{18}$ & $(0.0500,0.0679,0.0626,0.0427)^{\rm T}$ & 1.386244 \\
$a_{19}$ & $(0.0515,0.0612,0.0684,0.0523)^{\rm T}$ & 1.386270 \\
\bottomrule
\end{tabular}}
\end{table}
\end{small}

In light of the values of $\bm{v_i}$ and the $ES$ strategy, the information amount of each alternative $a_i$, $a_i\in A\textbackslash A^{R,0}$ can be calculated by Eq. \eqref{eq:IA_entropy}, which are recorded in the third column of Table \ref{table:v_0}. On this basis, the alternative $a_{17}$ is identified as the most informative alternative by Eq. \eqref{eq:XY}. Afterwards, the decision analyst poses a question to the decision maker, i.e., ``which category can the alternative $a_{17}$ be assigned to?'' and the decision maker answers that the alternative $a_{17}$ should be assigned to the category $C_4$.

Let $t=1$, then update $S^0$ and $A^{R,0}$ to obtain $S^1$ and $A^{R,1}$, respectively. Specifically, we have $S^1=\{a_3\rightarrow C_2,  a_{12}\rightarrow C_3, a_{16}\rightarrow C_4, a_{20}\rightarrow C_1, a_{17}\rightarrow C_4 \}$ and $A^{R,1}=\{a_3,a_{12},a_{16},a_{17},a_{20}\}$. As $t=1<T$, the most informative alternative should be identified from the set $A\textbackslash A^{R,1}=\{a_1,a_2,a_4,a_5,a_6,a_7,a_8,a_9,a_{10},a_{11},a_{13},a_{14},\\a_{15},a_{18},a_{19}\}$. For each alternative $a_i\in A\textbackslash A^{R^1}$ and category $C_h\in\{C_1,C_2,C_3,C_4\}$, let $S^{'}=S^1\cup \{a_i\rightarrow C_h\}$, and then solve the model \eqref{m:max_margin} to obtain the optimal objective function value as $m_{ih}^*$ and $\bm{v_i}=(m_{i1}^*,\ldots,m_{iq}^*)^{\rm T}$. The results are displayed in the second column of Table \ref{table:v_1}.

\begin{table}[htbp]
\centering
\setlength{\abovecaptionskip}{0pt}
\setlength{\belowcaptionskip}{10pt}
\caption{The values of $\bm{v_i}$ and the information amount of alternatives in  $A\textbackslash A^{R,1}$}
\label{table:v_1}
\scalebox{0.95}{
\begin{tabular}{ccc}
\toprule
 $a_i$ & $\bm{v_i}$ & $IA_{i}$    \\ \midrule
$a_1$ & $(0.00310, 0.0561, 0.0509, 0.0299)^{\rm T}$ &  1.386226 \\
$a_2$ & $(0.0379, 0.0558, 0.0613, 0.0354)^{\rm T}$ &  1.386232 \\
$a_4$ & $(0.0262, 0.0526, 0.0596, 0.0357)^{\rm T}$ &  1.386207 \\
$a_5$ & $(0.0117, 0.0631, 0.0089, 0.0048)^{\rm T}$ &  1.386006 \\
$a_6$ & $(0.0398, 0.0496, 0.0579, 0.0589)^{\rm T}$ &  1.386265 \\
$a_7$ & $(0.0639, 0.0645, 0.0614, 0.0439)^{\rm T}$ &  1.386259 \\
$a_8$ & $(0.0222, 0.0313, 0.0616, 0.0518)^{\rm T}$ &  1.386171 \\
$a_9$ & $(0.0455, 0.0548, 0.0638, 0.0490)^{\rm T}$ &  1.386270 \\
$a_{10}$ & $(0.0555, 0.0605, 0.0316, 0.0213)^{\rm T}$ &  1.386162 \\
$a_{11}$ & $(0.0598, 0.0425, 0.0244, 0.0171)^{\rm T}$ &  1.386157 \\
$a_{13}$ & $(0.0619, 0.0460, 0.0276, 0.0197)^{\rm T}$ &  1.386159 \\
$a_{14}$ & $(0.0645, 0.0645, 0.0640, 0.0569)^{\rm T}$ &  1.386289 \\
$a_{15}$ & $(0.0189, 0.0508, 0.0637, 0.0470)^{\rm T}$ &  1.386161 \\
$a_{18}$ & $(0.0449, 0.0597, 0.0623, 0.0427)^{\rm T}$ &  1.386256 \\
$a_{19}$ & $(0.0391, 0.0518, 0.0613, 0.0523)^{\rm T}$ &  1.386263 \\
\bottomrule
\end{tabular}}
\end{table}

Considering the values of $\bm{v_i}$ and the $ES$ strategy, the information amount of each alternative $a_i$, $a_i\in A\textbackslash A^{R,1}$ can be computed using Eq. \eqref{eq:IA_entropy}, and the results are recorded in the third column of Table \ref{table:v_1}.
Table \ref{table:v_1} reveals that according to Eq. \eqref{eq:XY}, the alternative $a_{14}$ is identified as the most informative alternative. On this basis, the decision analyst poses a question to the decision maker, i.e., ``which category can the alternative $a_{14}$ be assigned to?'' and the decision maker answers that the alternative $a_{14}$ should be assigned to the category $C_4$.

The above process is repeated and terminated until $t=8$. We summarize the identified most informative alternatives and the decision maker's assignment for these alternatives during the incremental preference elicitation process in Table \ref{table:most_IA}. Afterwards, in terms of all assignment example preference information, i.e., $S^8=\{a_3\rightarrow C_2,  a_{12}\rightarrow C_3, a_{16}\rightarrow C_4, a_{20}\rightarrow C_1, a_{17}\rightarrow C_4, a_{14}\rightarrow C_4, a_9\rightarrow C_2, a_7\rightarrow C_1, a_{18}\rightarrow C_2, a_{19}\rightarrow C_3, a_2\rightarrow C_4, a_{15}\rightarrow C_3 \}$, we solve the models \eqref{m:max_margin} and \eqref{m:min_slope} to yield the marginal utilities,
{\color{black} then we have $u_1(\beta_1^1)=0$, $u_1(\beta_1^2)=0.8376$, $u_1(\beta_1^3)=0.3518$, $u_1(\beta_1^4)=0.8998$, $u_1(\beta_1^5)=1$,  $u_2(\beta_2^1)=0.8848$, $u_2(\beta_2^2)=0$, $u_2(\beta_2^3)=1$, $u_2(\beta_2^4)=1$, $u_2(\beta_2^5)=1$,   $u_3(\beta_3^1)=0.5428$, $u_3(\beta_3^2)=1$, $u_3(\beta_3^3)=0.9511$, $u_3(\beta_3^4)=0$, $u_3(\beta_3^5)=1$.
}

\begin{table}[htbp]
\centering
\setlength{\abovecaptionskip}{0pt}
\setlength{\belowcaptionskip}{10pt}
\caption{The final identified most informative alternatives and category assignments}
\label{table:most_IA}
\begin{tabular}{>{\centering\arraybackslash}p{2cm} >{\centering\arraybackslash}p{2cm} >{\centering\arraybackslash}p{2cm}}
\toprule
$t$ & $a^{*,t}$ & $C_{h^*}$\\ \midrule
0 & $a_{17}$ & $C_4$\\
1 & $a_{14}$ & $C_4$\\
2 & $a_9$ & $C_2$\\
3 & $a_7$ & $C_1$\\
4 & $a_{18}$ & $C_2$\\
5 & $a_{19}$ & $C_3$\\
6 & $a_2$ & $C_4$\\
7 & $a_{15}$ & $C_3$\\\bottomrule
\end{tabular}
\end{table}

\begin{figure}[!t]
\centering
\includegraphics[scale=0.6]{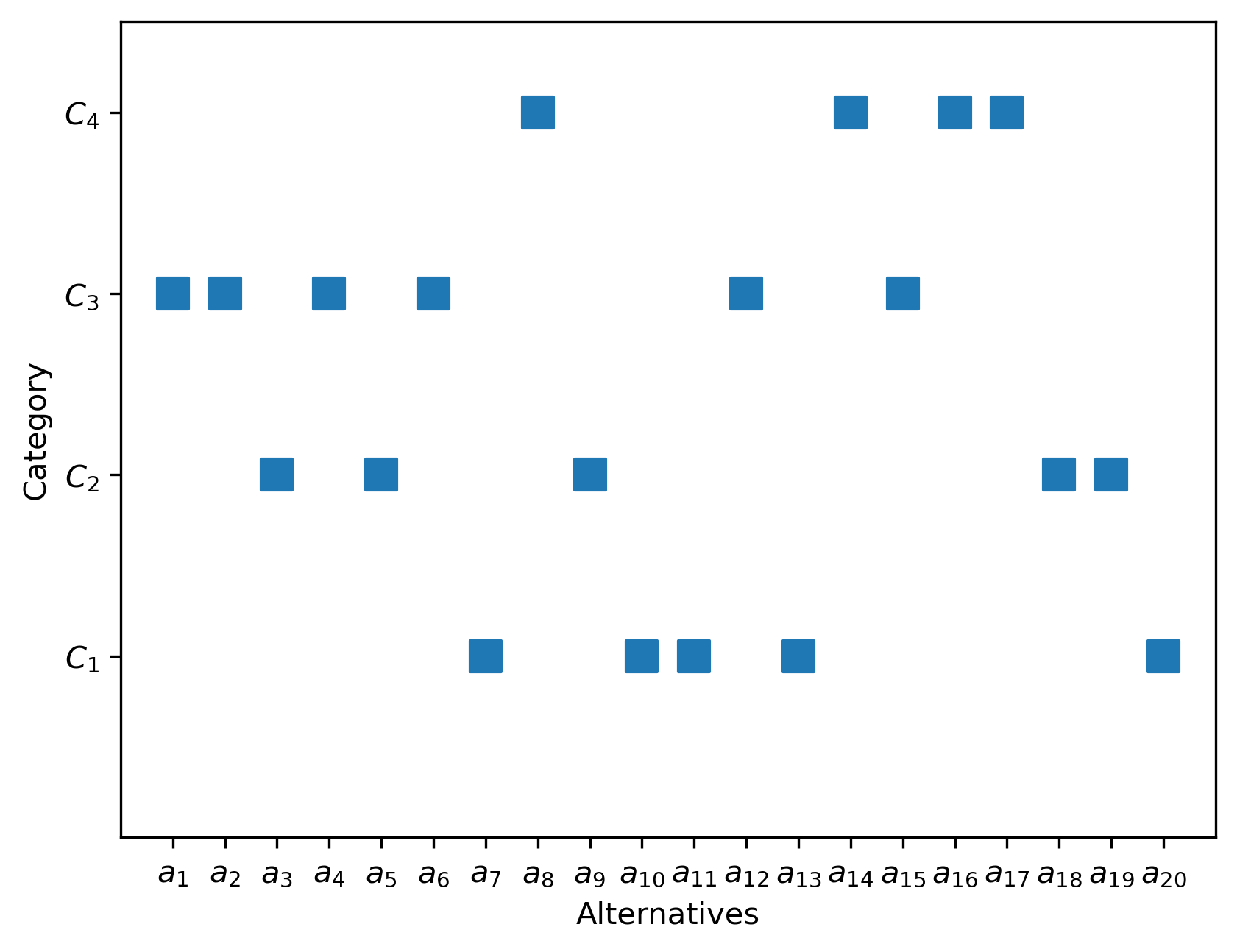}
\caption{The sorting result for all firms}
\label{fig:sorting_result}
\end{figure}

Moreover, the category threshold vector is obtained as $b=(0, 1.5715, 1.9367, 2.1765, 3.2398)^{\rm T}$. Followed by this, by employing the threshold-based MCS model, the sorting result for all firms is obtained and visualized in Fig. \ref{fig:sorting_result}. {\color{black}In terms of the predetermined and inferred sorting results for all firms, the accuracy of the proposed approach is calculated as 0.65 by Eq. \eqref{eq:accuracy}.}

\begin{figure}[htbp]
\centering
\includegraphics[scale=0.4]{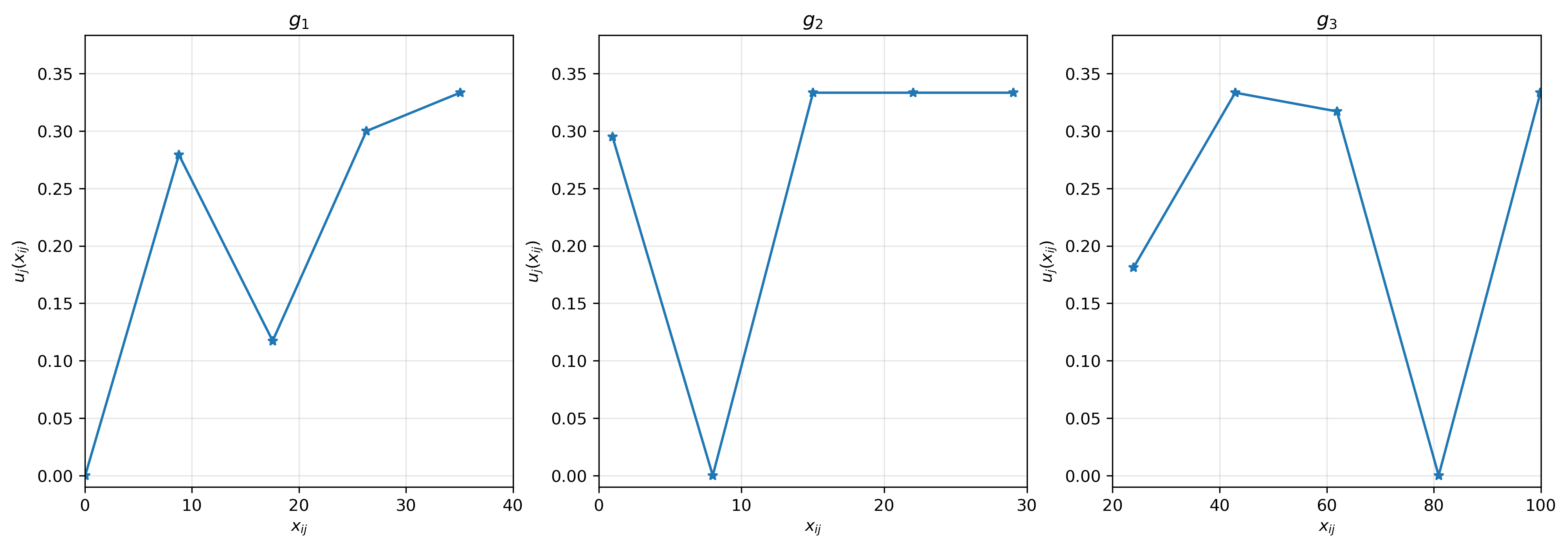}
\caption{The marginal utility functions in the UTA-like functional space}
\label{fig:transformed_U}
\end{figure}

Furthermore, we employ Eq. \eqref{eq:trandormation_function} to map the marginal utility functions and category thresholds into the UTA-like functional space. The transformed marginal utility functions are shown in Fig. \ref{fig:transformed_U}, and the transformed category threshold vector is derived as $b^S=(0,0.5238,0.6456,0.7255,1.0799)^{\rm T}$.
\vspace{-1em}

{\color{black}

\subsection{Comparisons and parameter analysis}\label{sec:5.2}
In this subsection, we present comparisons and parameter analysis using the data from the illustrative example.

\subsubsection{Comparisons with the approach considering monotonic preferences}
We first compare the performance of the proposed approach with the approach that considers monotonic preferences. Without loss of generality, in the illustrative example, we consider monotonically increasing preferences by adding the constraints $u_j(\beta^{l+1}_j)-u_j(\beta^l_j)\ge 0$, $l=1,\ldots,s_j$, $j\in M$ to the models \eqref{m:max_margin} and \eqref{m:min_slope}. On this basis, we implement Algorithm \ref{alg:1}, and the sorting result for the twenty firms is derived as $F_{Mono}=(C_2, C_2, C_2, C_2, C_2,  C_2, C_1, C_4, C_2, C_2, C_2,  C_3, C_2,  C_4, C_2, C_4,  C_4, C_2, C_2, C_2)^{\rm T}$. Furthermore, the marginal utility functions in the UTA-like functional space are displayed in Fig. \ref{fig:monotonic_transformed_U}, and the category thresholds in the UTA-like functional space are obtained as $b^S=(0, 0.1603, 0.3206, 0.4809, 1.1603)^{\rm T}$.
\begin{figure}[htbp]
\centering
\includegraphics[scale=0.4]{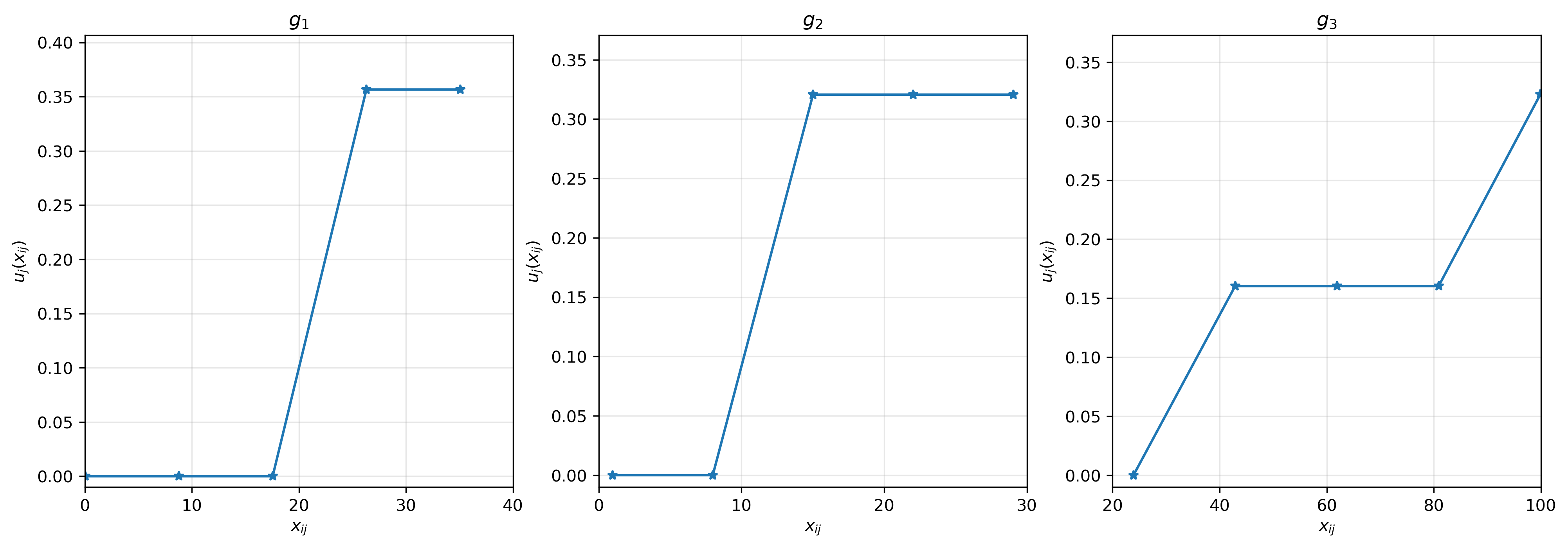}
\caption{The marginal utility functions in the UTA-like functional space of the approach considering monotonic preferences}
\label{fig:monotonic_transformed_U}
\end{figure}

According to Eq. \eqref{eq:accuracy}, the accuracy of the approach considering monotonic preferences is 0.5, which is lower than the accuracy of the proposed approach (0.65). Thus, although considering monotonic preferences results in marginal utility functions that are more interpretable for the decision maker, the performance of the proposed approach, which accommodates potentially non-monotonic preferences, is superior in the illustrative example.

\subsubsection{The impact of the parameters $\alpha$ and $T$}

To investigate the impact of parameters $\alpha$ and $T$ on the performance of the proposed approach, we initially set $\alpha\in \{0.1, 0.3, 0.5, 0.7\}$, maintaining all other parameter settings as described in Section \ref{sec:5.1}. Algorithm \ref{alg:1} is then executed under various values of $\alpha$. Subsequently, we assign $T=\{8,10,12,14,16\}$ while keeping the remaining parameters consistent with those in Section \ref{sec:5.1}, and run Algorithm \ref{alg:1} for different values of $T$. The experimental results are illustrated in Fig. \ref{fig:alpha_T}.
\begin{figure}[htbp]
\centering
\includegraphics[scale=0.4]{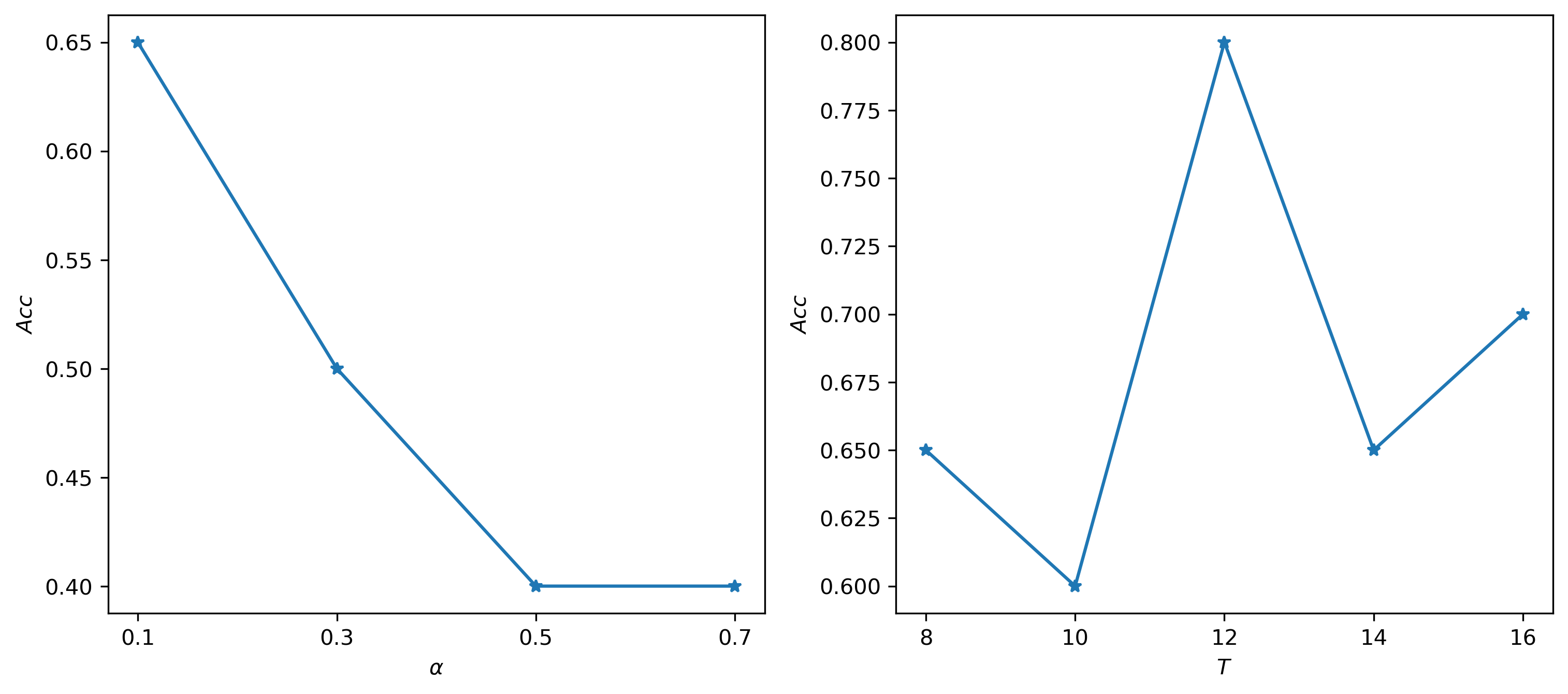}
\caption{The performance of the proposed approach under different $\alpha$ and $T$ values}
\label{fig:alpha_T}
\end{figure}

Fig. \ref{fig:alpha_T} reveals that as $\alpha$ increases, the performance of the proposed approach generally exhibits a downward trend. This decline in performance may be attributed to the decreasing weight of inconsistency in the objective function of the model \eqref{m:max_margin} as $\alpha$ increases, leading to greater inconsistency in the model inferred from the assignment example preference information, which in turn results in poorer performance. Meanwhile, although some non-monotonic fluctuations in accuracy occur as $T$ increases, the overall performance tends to rise. This improvement is likely due to the increased assignment example preference information, which facilitates more adequate training.

}

\section{Experimental analysis}\label{sec:6}
This section undertakes comprehensive experimental analysis to validate the effectiveness of the proposed approach. Specifically, we begin by presenting specific experimental design used in this section. Subsequently, we conduct extensive computational experiments on both artificial and real-world data sets to further justify the effectiveness of the proposed approach.

\subsection{Experimental design}\label{sec:6.1}
In this subsection, we first outline the employed evaluation metrics and illustrate the method for generating artificial data sets. Afterwards, we provide two algorithms that will be used in the subsequent computational experiments.

This subsection begins by introducing some evaluation metrics. To be specific, when using termination criterion \RNum{1}, we employ the accuracy metric to assess the performance, as expressed by Eq. \eqref{eq:accuracy}. When using termination criterion \RNum{2}, we adopt cost saving rate to measure the performance of the proposed approach, as defined below.
\begin{definition}
Let $LA$ represent the cardinality of the set containing all assignment example preference information provided by the decision maker in the incremental preference elicitation process, and $TR$ be the number of initially divided training data, then the cost saving rate $CS$ is calculated by
\begin{equation}\label{eq:CS}
CS = \frac{TR-LA}{TR},
\end{equation}
if $CS>0$,  it signifies that the employed question selection strategy can reduce labeling costs. Moreover, the greater the value of $CS$, the better the performance of the question selection strategy.
\end{definition}

Following this, we present an algorithm designed for the generation of artificial data sets (see Algorithm \ref{alg:3} in the supplemental file). In particular, given the capability of the proposed approach to handle inconsistent assignment example preference information, we introduce some noise during the data generation process.

Additionally, we provide two algorithms that will be used in the following computational experiments (see Algorithms \ref{alg:4} and \ref{alg:5} in the supplemental file). Among them, Algorithm \ref{alg:4} is suitable for the termination criterion \RNum{1}, while Algorithm \ref{alg:5} is suitable for the termination criterion \RNum{2}.

\subsection{Computational experiments on artificial data sets}\label{sec:6.2}
In this subsection, we perform some computational experiments on artificial data sets to validate the effectiveness of the proposed approach. To do so, we first compare the performance of the proposed approach using different probability transformation methods. Following this, we compare the proposed question selection strategies with several benchmark question selection strategies and examine the impact of various parameter settings on their performance.

{\color{black}
\subsubsection{Comparisons of different probability transformation methods}\label{sec:6.2.1}
In this subsection, we compare the performance of different probability transformation methods on the proposed approach.

We set the parameters as follows:  $n=100$, $m=4$, $q=3$, $s_j=4$, $j=1,2,3,4$, $\eta=0.05$, $T=30$, $r=0.6$, $lr=0.2$, $\alpha=0.1$. Algorithm \ref{alg:3} is executed ten times to generate ten different datasets. For each dataset, Algorithms \ref{alg:4} and \ref{alg:5} are executed ten times using different question selection strategies: $ER$, $ES$, $LR$, $LS$, $MR$, and $MS$. We record the accuracy and cost saving rate for each run. Based on these runs, we calculate the average values of $Acc^t$ and $CS$ across the ten datasets. The results are presented in Figs. \ref{fig:simu_acc_proba}~-~\ref{fig:simu_cs_proba}.

\begin{figure}[htbp]
\centering
\includegraphics[scale=0.4]{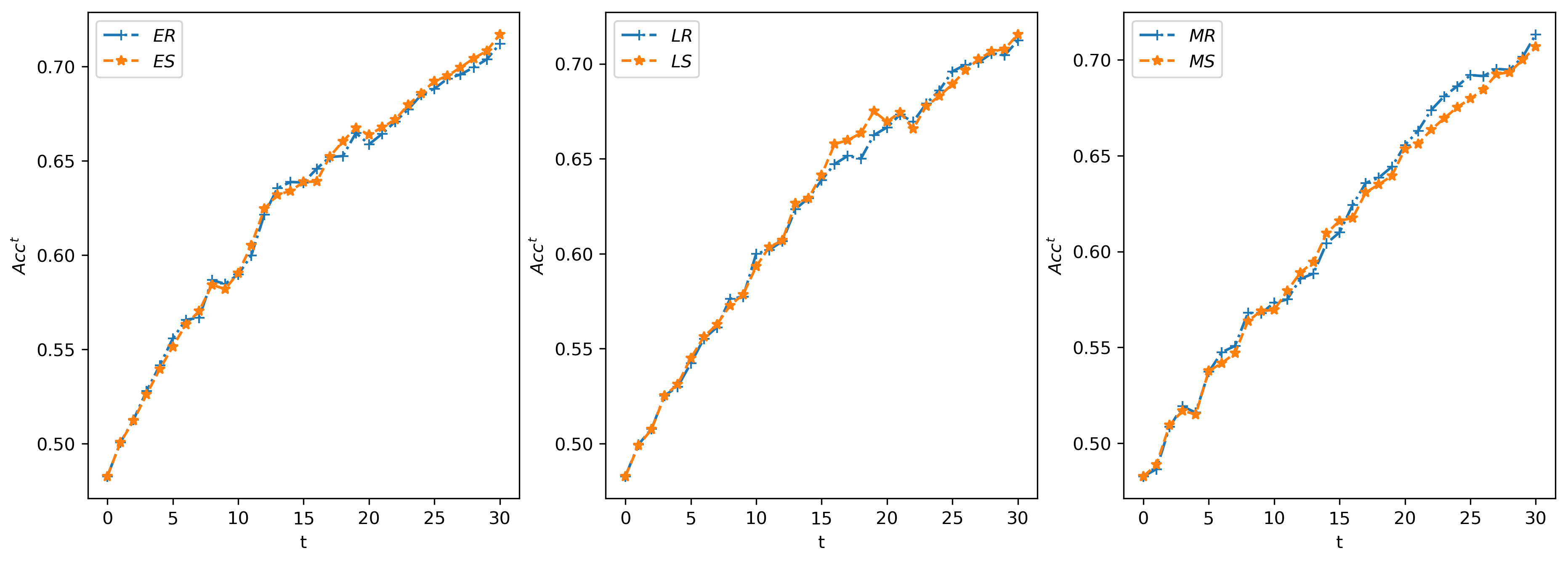}
\caption{The average values of  $Acc^t$ with different probability transformation methods}
\label{fig:simu_acc_proba}
\end{figure}

\begin{figure}[htbp]
\centering
\includegraphics[scale=0.4]{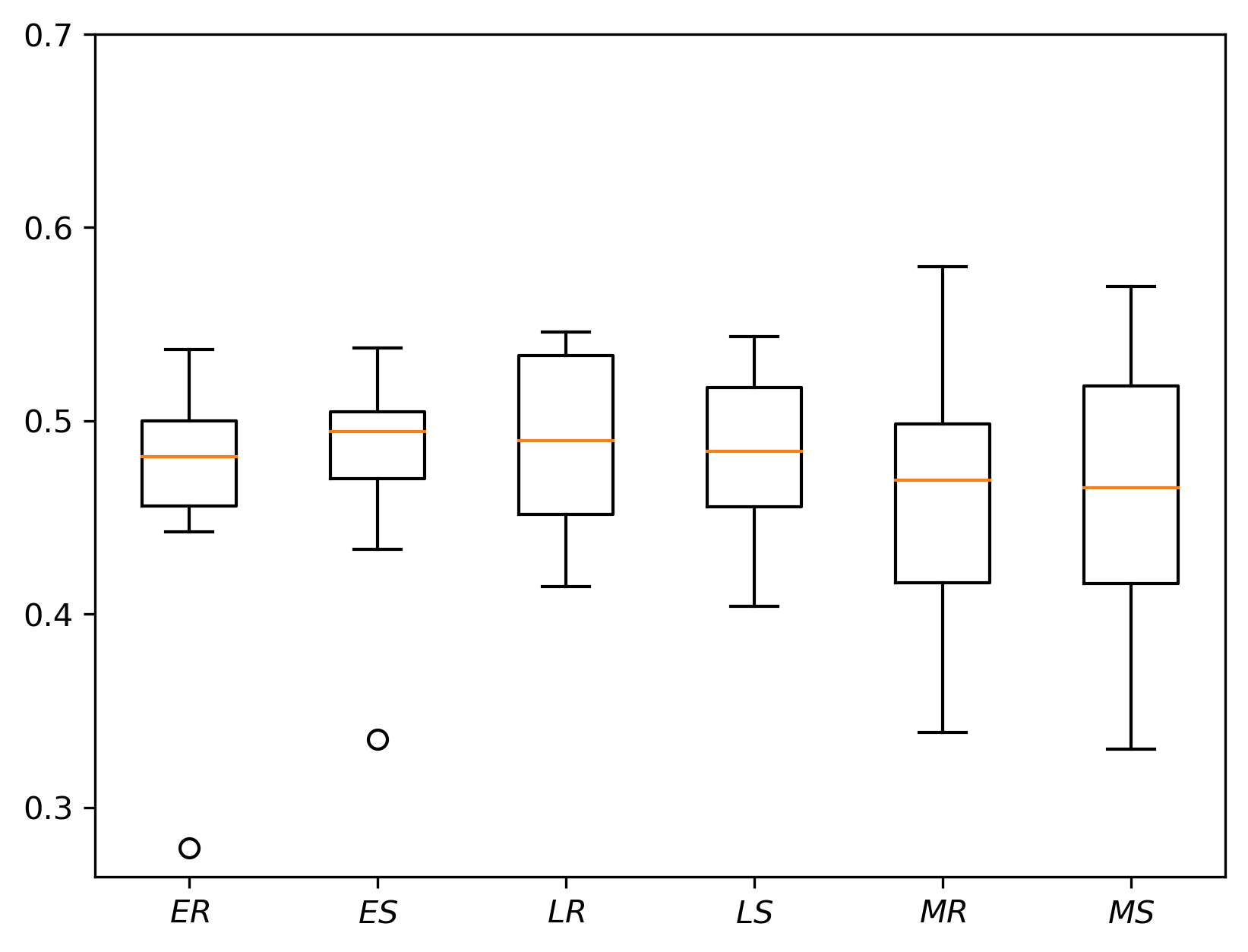}
\caption{The average values of  $CS$ with different probability transformation methods}
\label{fig:simu_cs_proba}
\end{figure}

Figs. \ref{fig:simu_acc_proba}~-~\ref{fig:simu_cs_proba} demonstrate that the differences between $ER$ and $ES$, $LR$ and $LS$, as well as $MR$ and $MS$ are minimal in terms of both accuracy and cost saving metrics. Therefore, the adoption of diverse probability transformation methods does not fundamentally change the nature of the proposed approach.
}

\subsubsection{Comparisons of different question selection strategies and parameter analysis}\label{sec:6.2.2}
In this subsection, we compare the proposed question selection strategies and several benchmark strategies and analyze the impact of different parameter settings on their performance. To achieve this, we first present the considered benchmark strategies.

$\bullet$ \ $RAND$ strategy: this strategy randomly selects an alternative from the set $A \textbackslash A^{R,t}$ in each iteration of the incremental preference elicitation process.

{\color{black}
$\bullet$ \ Probability sorting-based strategy: this strategy is based on a probabilistic variant of the threshold-based value-driven sorting procedure proposed by \cite{Liu23joc}, which allows for calculation of the probability of assigning an alternative to each category. For convenience, let $\varphi_{ih}$ be the consistency degree of assigning an alternative $a_i$ to a category $C_h$, then it can be defined as
\begin{equation}
\begin{aligned}
\varphi_{ih}=\left\{\begin{array}{ll}
b_1-\varepsilon -U(a_i), & h=1 \\
\min\{U(a_i)-b_{h-1},b_h - \varepsilon-U(a_i)\}, & h=2,\ldots,q-1, \\
U(a_i)-b_{q-1}, & h=q
\end{array}\quad i\in N.
\right.
\end{aligned}
\end{equation}

On this basis, the probability $p_{ih}$ of assigning an alternative $a_i$ to a category $C_h$ is calculated by
\begin{equation}\label{eq:ps}
p_{ih}=\frac{e^{\tau \varphi_{ih}}}{\sum\nolimits_{h=1}^q e^{\tau \varphi_{ih}}}, i\in N, h\in Q,
\end{equation}
where $\tau>0$ is the temperature parameter. It is noteworthy that Eq. \eqref{eq:ps} has similar expressions to Eq. \eqref{eq:softmax}. Without loss of generality, we let $\tau=1$ in the subsequent experiments.

We then combine Eq. \eqref{eq:ps} with the three metrics introduced in Section \ref{sec:4.3.1} to measure the information amount associated with each alternative. This results in three question selection strategies: $PES$, $PLS$ and  $PMS$. Specifically, the $PES$ strategy relies on the entropy-based metric, the $PLS$ strategy uses the least confidence-based metric and $PMS$ strategy employs the margin of confidence-based metric.
}

Subsequently, we describe the detailed experimental settings across different parameters to compare the proposed question selection strategies with the benchmark strategies: $RAND$, $PES$, $PLS$  and $PMS$. As analyzed in Section \ref{sec:6.2.1}, different probability transformation methods have minimal impact on the performance of the proposed approach. Therefore, without loss of generality, we present the results using only the Softmax function-based transformation method when considering the probability-based measure in the following experiments, i.e., the strategies $ES$, $LS$ and $MS$.

\begin{figure}[htbp]
\centering
\includegraphics[scale=0.33]{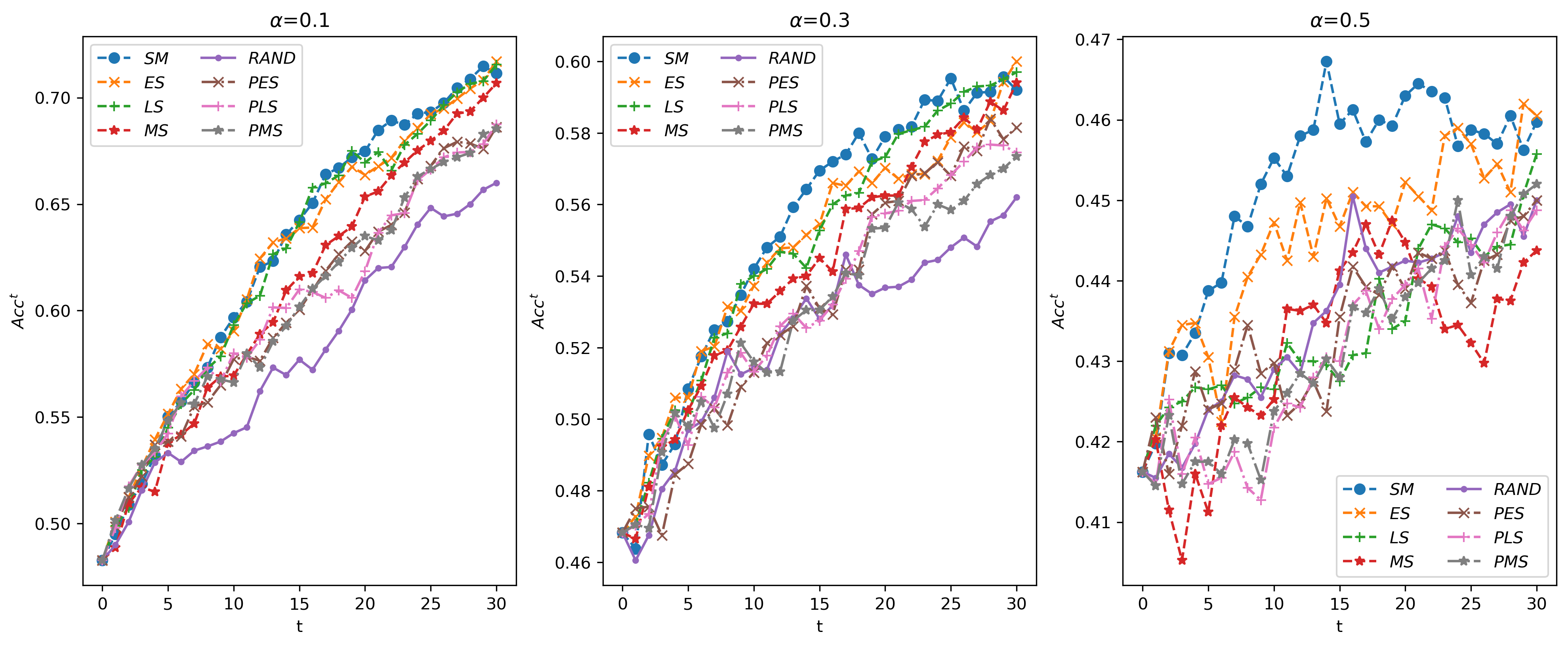}
\caption{The average values of $Acc^t$ for different question selection strategies with different values of $\alpha$}
\label{fig:acc_alpha}
\end{figure}

\begin{figure}[htbp]
\centering
\includegraphics[scale=0.35]{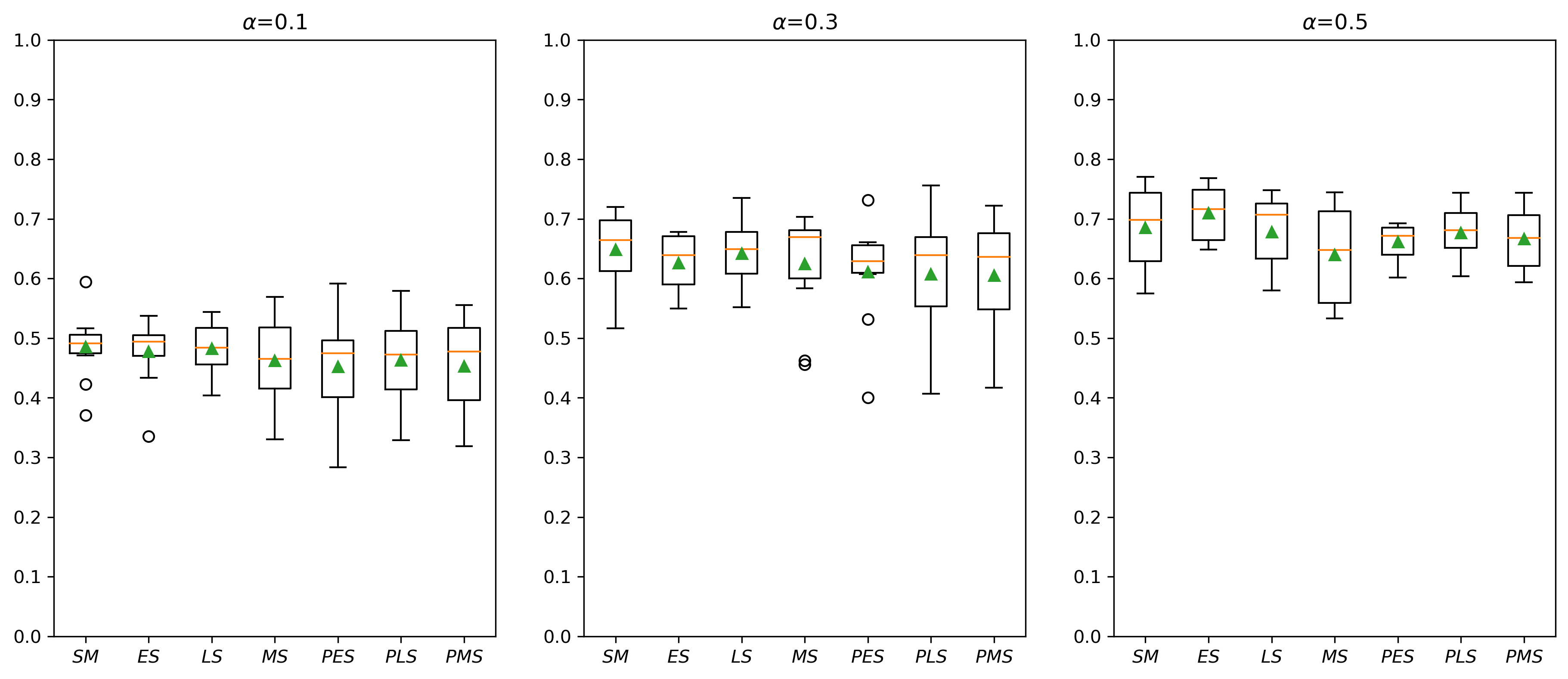}
\caption{The average values of  $CS$ for different question selection strategies with different values of $\alpha$}
\label{fig:cs_alpha}
\end{figure}

{\color{black}
\textbf{Experiment \RNum{1}:} In this experiment, we investigate the influence of the parameter $\alpha$ on the performance of the considered question selection strategies. We set $n=100$, $m=4$, $q=3$, $s_j=4$, $j=1,2,3,4$, $lr=0.2$, $\eta=0.05$, $T=30$, $r=0.6$. For each parameter setting, we execute Algorithm \ref{alg:3} ten times to generate ten data sets. Subsequently, for each question selection strategy, each data set and each value of $\alpha\in \{0.1, 0.3, 0.5\}$, we execute Algorithms \ref{alg:4} and  \ref{alg:5} ten times, and record the accuracy $Acc^t$ and cost saving rate $CS$. The average values of $Acc^t$ and $CS$ are shown in Figs. \ref{fig:acc_alpha}~-~\ref{fig:cs_alpha}.
}

\begin{figure}[htbp]
\centering
\includegraphics[scale=0.33]{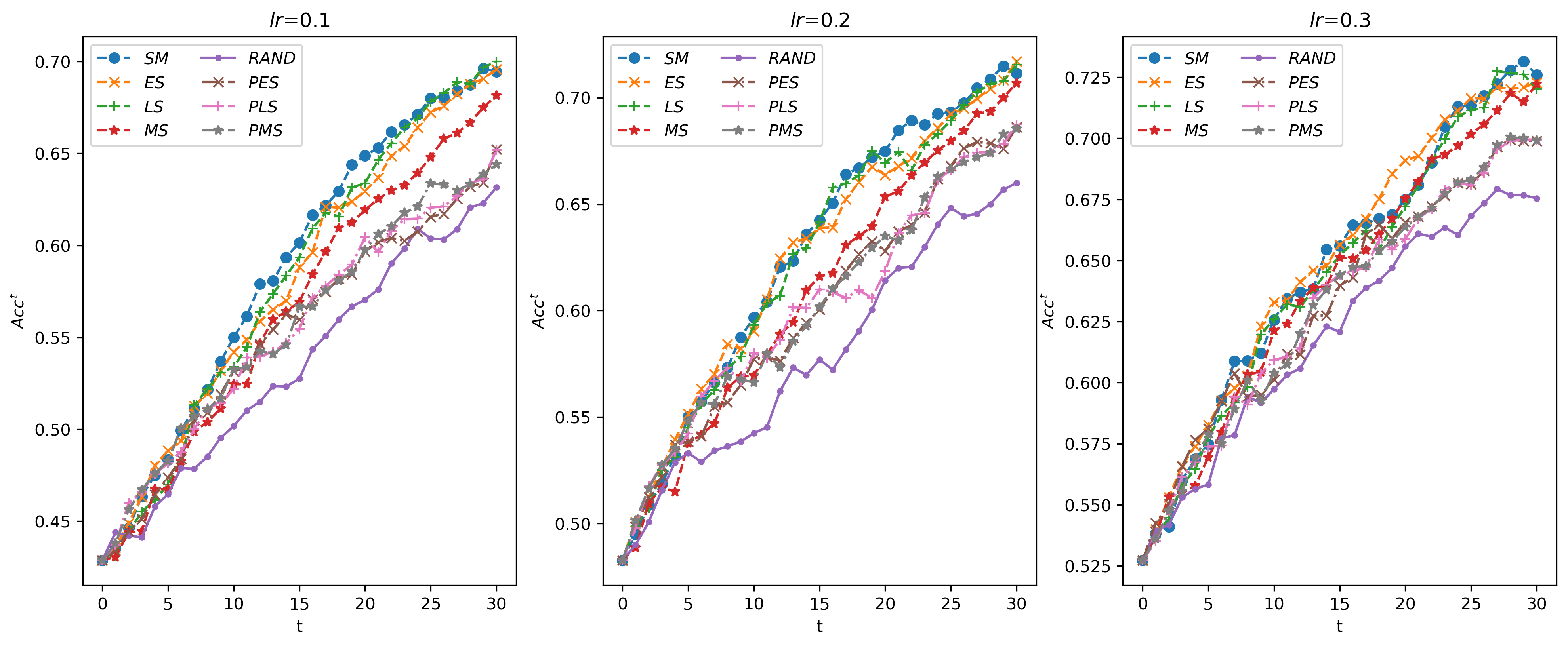}
\caption{The average values of $Acc^t$ for different question selection strategies with different values of $lr$}
\label{fig:acc_lr}
\end{figure}

\begin{figure}[htbp]
\centering
\includegraphics[scale=0.35]{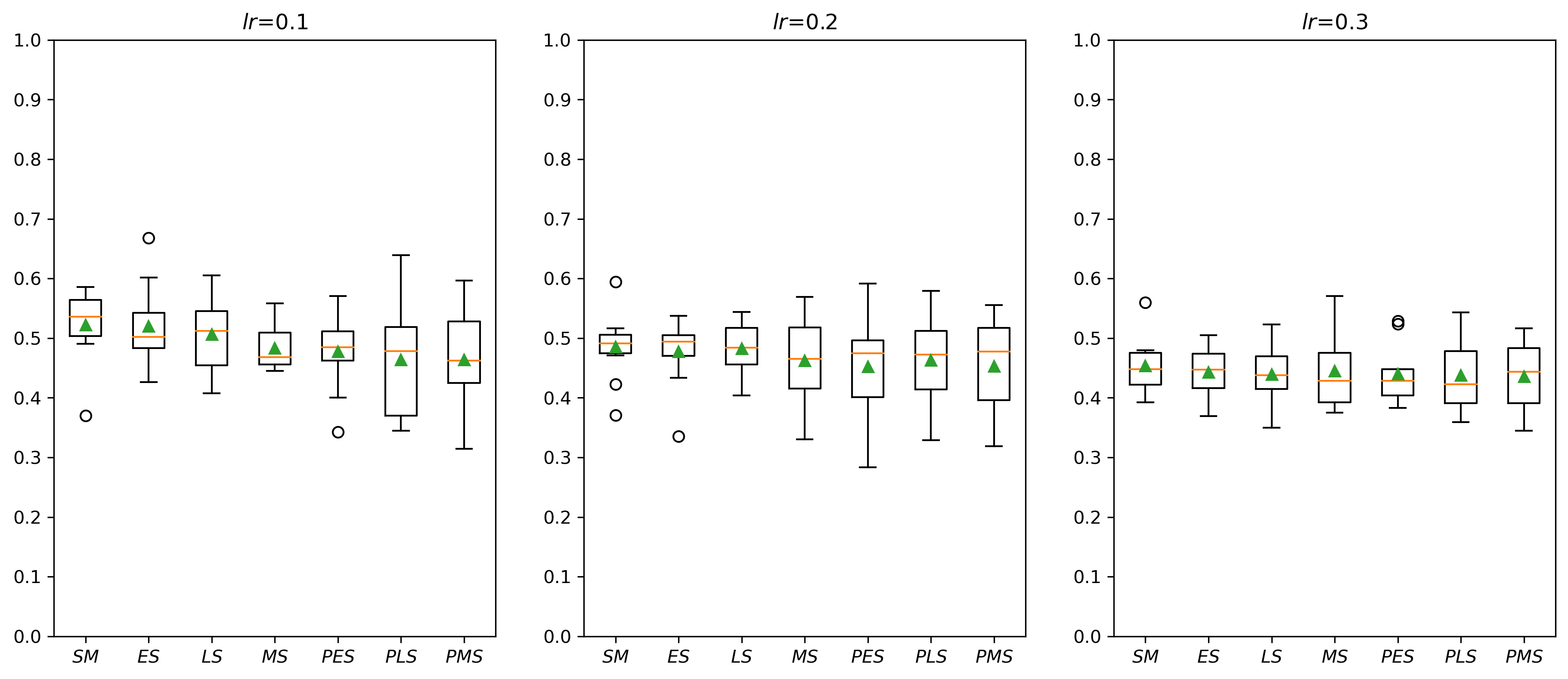}
\caption{The average values of $CS$ for different question selection strategies with different values of $lr$}
\label{fig:cs_lr}
\end{figure}

\textbf{Experiment \RNum{2}:} In this experiment, we investigate the influence of different proportion of initial assignment example preference information to the training data $lr$ on different question selection strategies. To this end, let $n=100$, $m=4$, $q=3$, $s_j=4$, $j=1,2,3,4$, $\eta=0.05$, $T=30$, $r=0.6$, $\alpha=0.1$ and then we execute Algorithm \ref{alg:3} ten times to generate ten data sets. Afterwards, we execute Algorithms \ref{alg:4} and  \ref{alg:5} ten times for each considered question selection strategies, each data set and each value of $lr\in\{0.1,0.2,0.3\}$, and record the values of $Acc^t$ and $CS$. On this basis, the average values of $Acc^t$ and $CS$ on these ten data sets are calculated and shown in Figs. \ref{fig:acc_lr}~-~\ref{fig:cs_lr}.

\begin{figure}[htbp]
\centering
\includegraphics[scale=0.33]{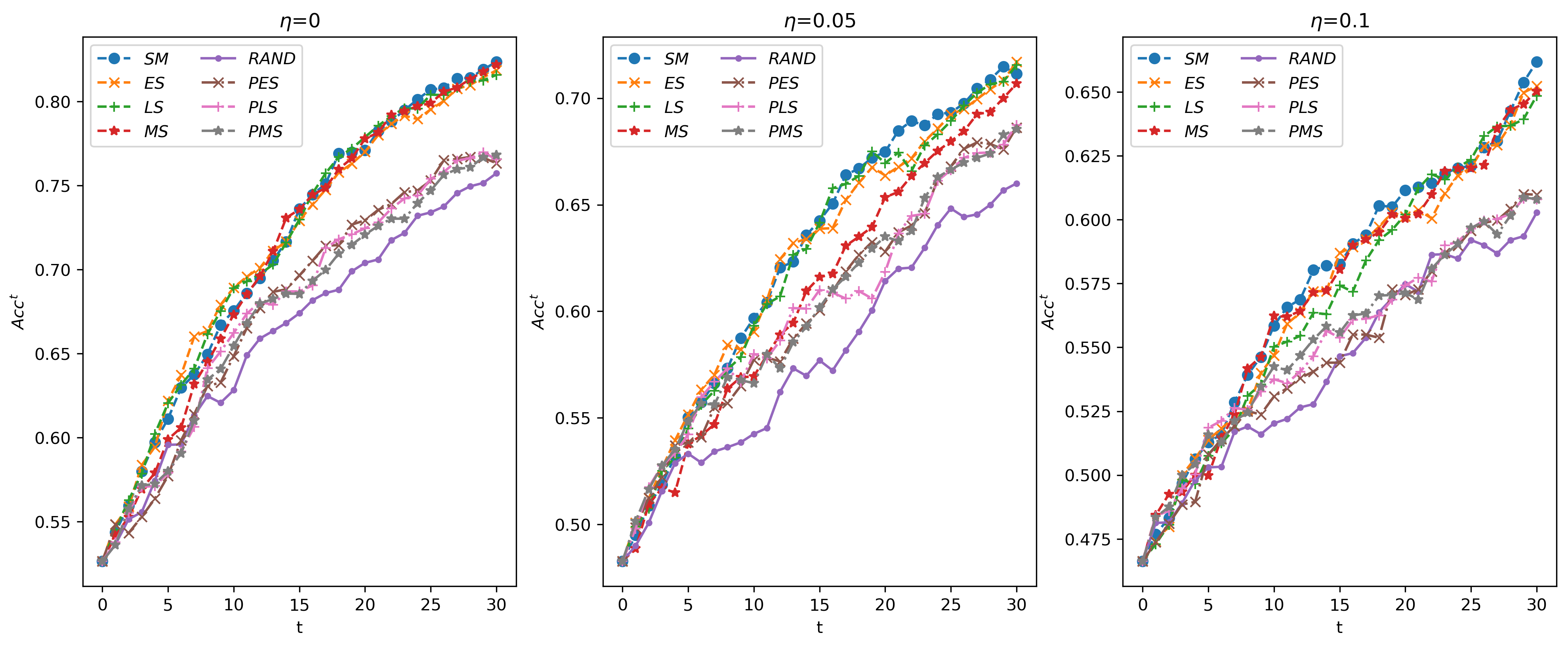}
\caption{The average values of  $Acc^t$ for different question selection strategies with different values of $\eta$}
\label{fig:acc_noise}
\includegraphics[scale=0.35]{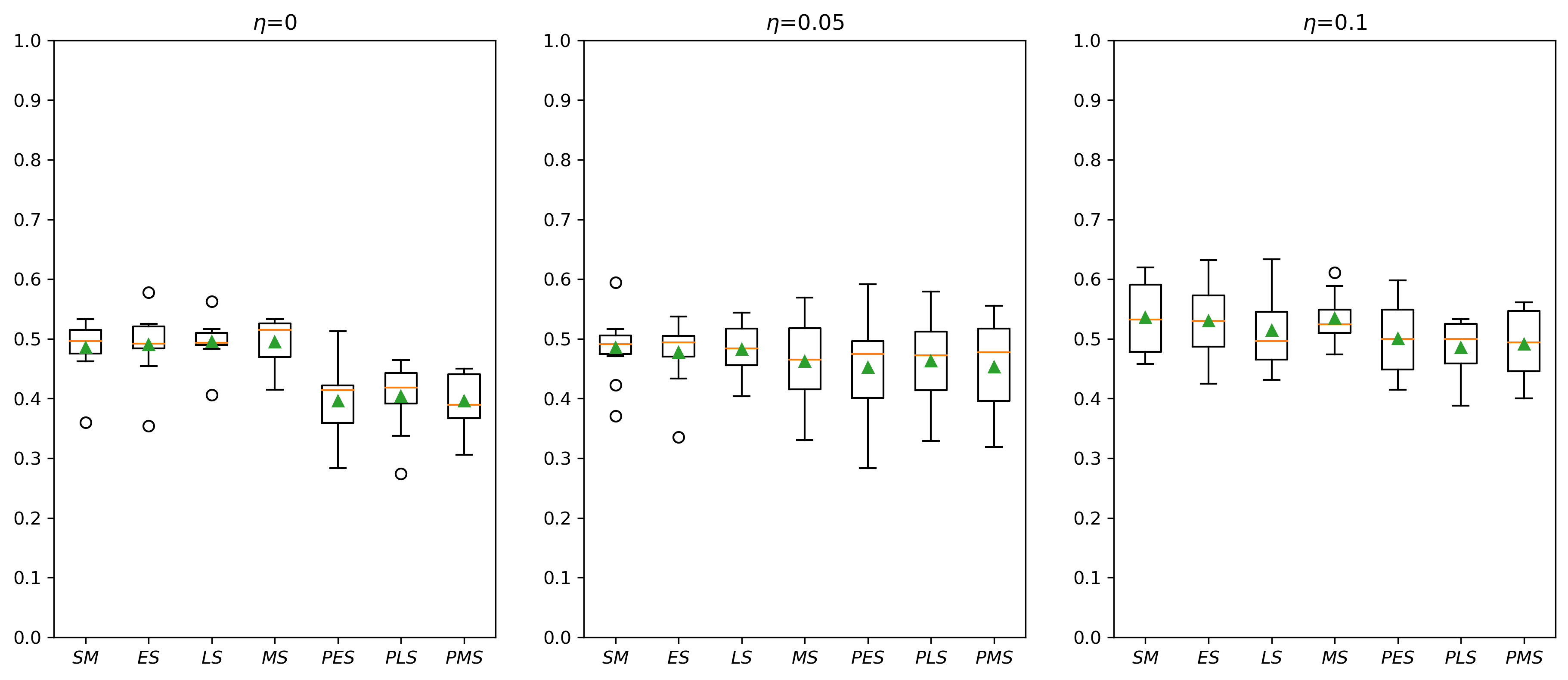}
\caption{The average values of  $CS$ for different question selection strategies with different values of $\eta$}
\label{fig:cs_noise}
\end{figure}

\textbf{Experiment \RNum{3}:} This experiment focuses on exploring how the proportion of noise $\eta$ in the data set affects the considered question selection strategies. To accomplish this, let $n=100$, $m=4$, $q=3$, $s_j=4$, $j=1,2,3,4$, $lr=0.2$, $T=30$, $r=0.6$, $\alpha=0.1$ and vary $\eta$ in $\{0, 0.05, 0.1\}$. For a fixed value of $\eta$, Algorithm \ref{alg:3} is utilized to generate ten different data sets in terms of the values of $n$, $m$, $q$, $s_j$. For each considered question selection strategies and each data set, Algorithms \ref{alg:4} and  \ref{alg:5} are implemented ten times, respectively. The values of $Acc^t$ and $CS$ under different question selection strategies and each data set are recorded. Followed by this, for each value of $\eta$, the average values of $Acc^t$ and $CS$ on these ten different data sets are calculated and visualized in Figs. \ref{fig:acc_noise}~-~\ref{fig:cs_noise}.

\begin{figure}[htbp]
\centering
\includegraphics[scale=0.33]{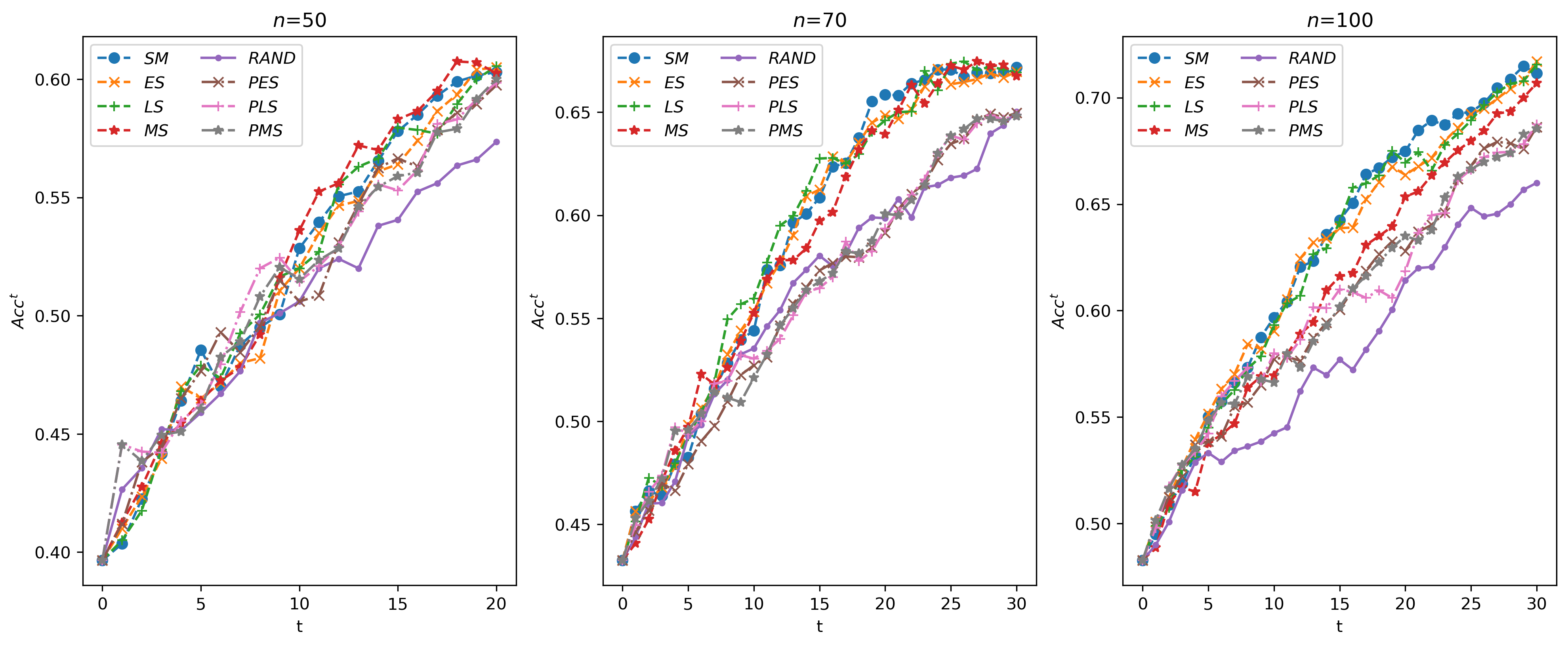}
\caption{The average values of  $Acc^t$ for different question selection strategies with different values of $n$}
\label{fig:acc_n}
\end{figure}

\begin{figure}[htbp]
\centering
\includegraphics[scale=0.35]{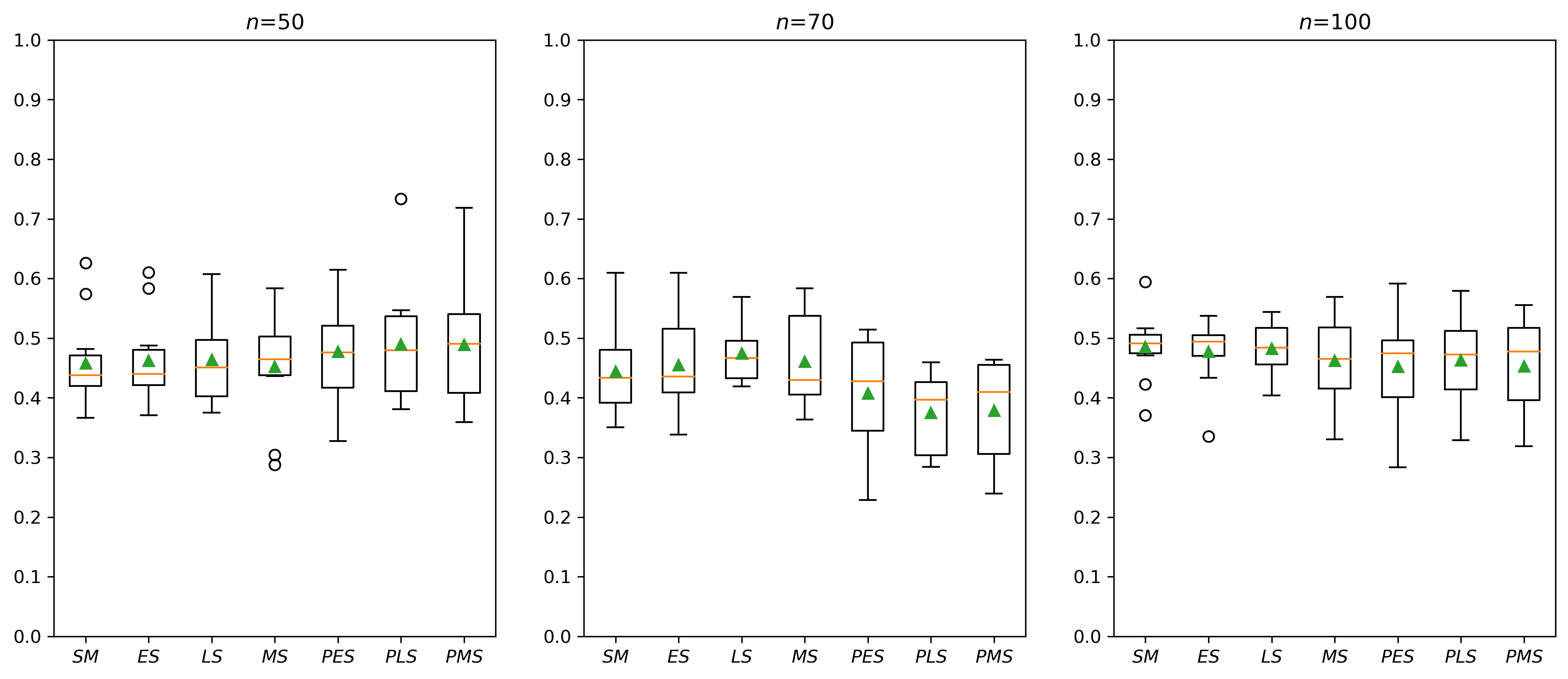}
\caption{The average values of  $CS$ for different question selection strategies with different values of $n$}
\label{fig:cs_n}
\end{figure}

\textbf{Experiment \RNum{4}:} In this experiment, we study the influence of the number of alternatives $n$ on the proposed approach. To do so, let $n\in \{50,70,100\}$, $m=4$, $q=3$, $s_j=4$, $j=1,2,3,4$, $lr=0.2$, $T=30$, $r=0.6$, $\eta=0.05$, $\alpha=0.1$. First, for a fixed value of $n$, we utilize Algorithm \ref{alg:3} to generate ten different data sets and implement Algorithms \ref{alg:4} and  \ref{alg:5} ten times for each data set. On this basis, we calculate the average values of $Acc^t$ and $CS$ for different question selection strategies on each data set. Finally, the average values of $Acc^t$ and $CS$ for different question selection strategies are computed based on the results of different data sets with different values of $n$. The results are displayed in Figs. \ref{fig:acc_n}~-~\ref{fig:cs_n}.

\begin{figure}[htbp]
\centering
\includegraphics[scale=0.33]{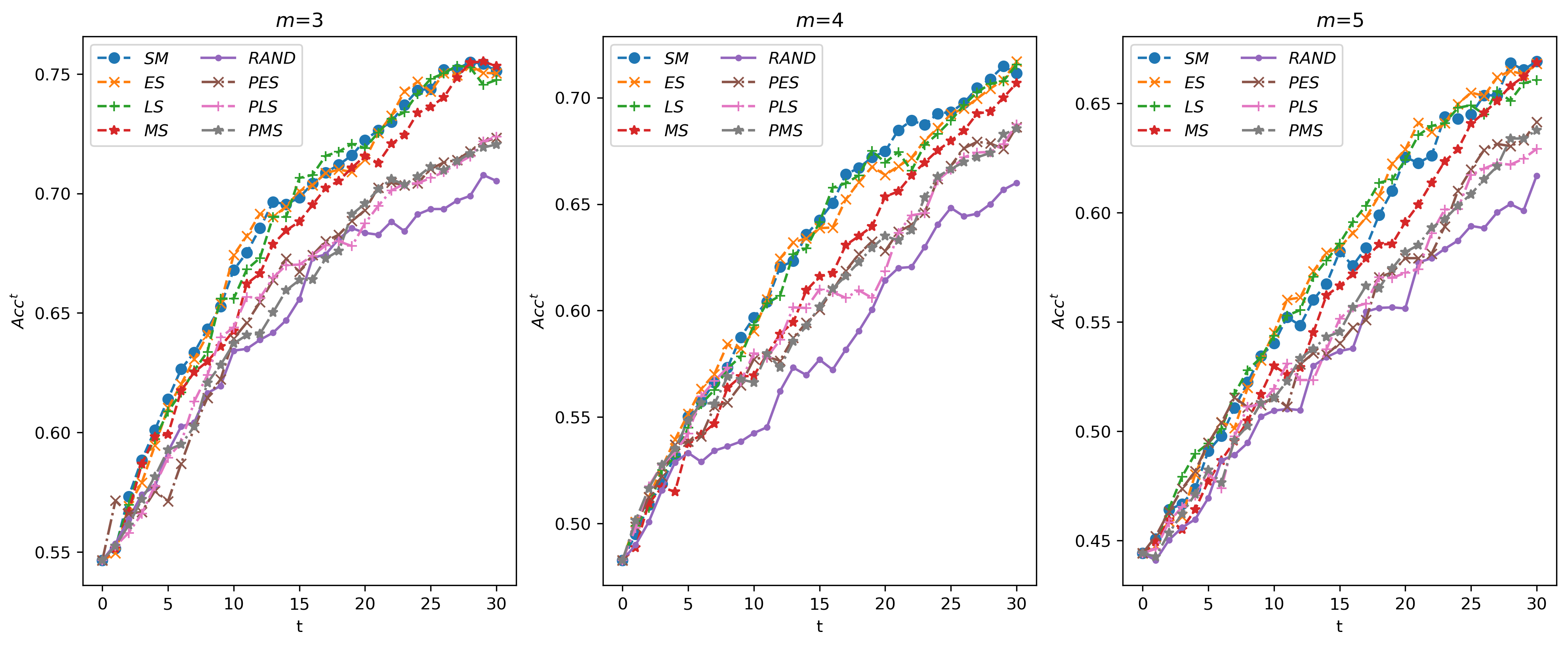}
\caption{The  average values of $Acc^t$ for different question selection strategies with different values of $m$}
\label{fig:acc_m}
\centering
\includegraphics[scale=0.35]{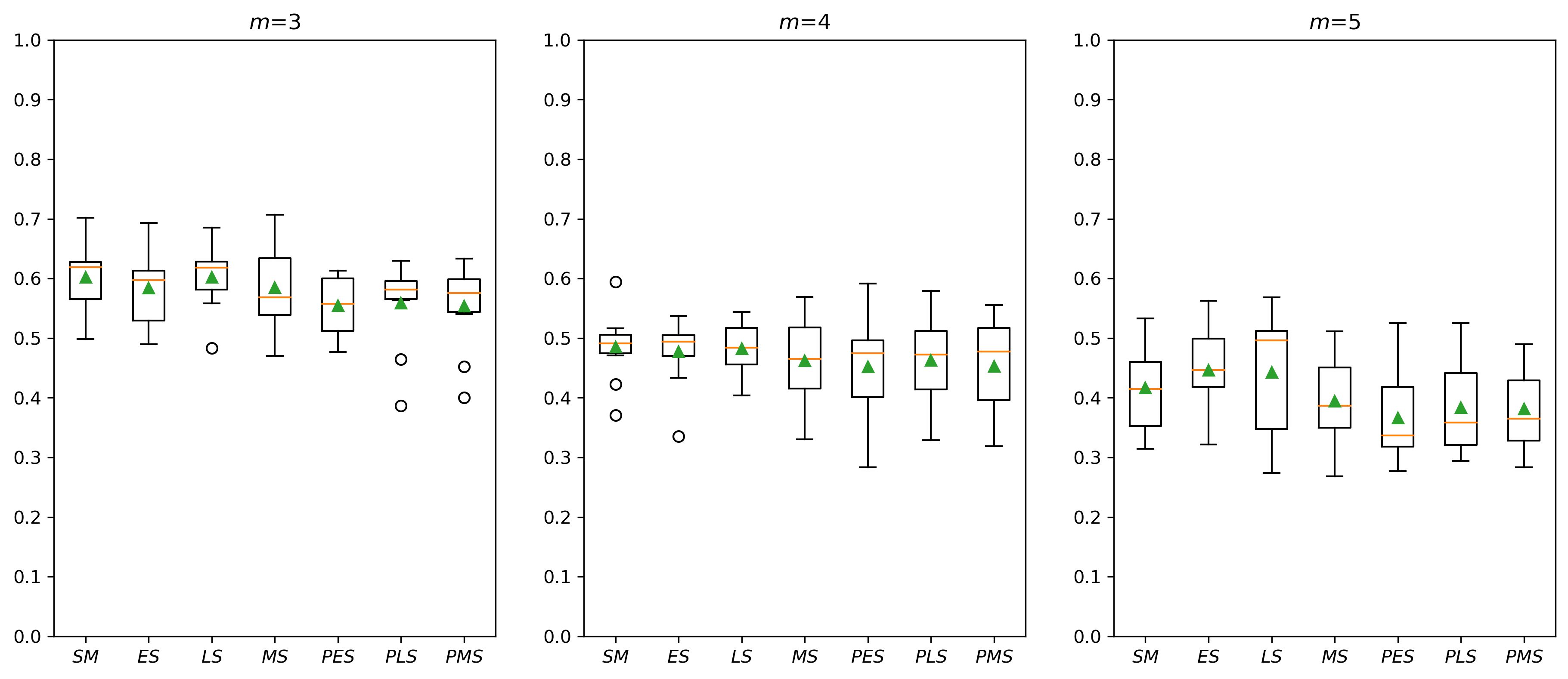}
\caption{The average values of  $CS$ for different question selection strategies with different values of $m$}
\label{fig:cs_m}
\end{figure}

\textbf{Experiment \RNum{5}:} In this experiment, our attention is the impact of the number of criteria $m$ on the proposed approach. To this end, we set $n=100$, $m\in\{3,4,5\}$, $q=3$, $s_j=4$, $j=1,2,3,4$, $lr=0.2$, $T=30$, $r=0.6$, $\eta=0.05$, $\alpha=0.1$. On this basis, Algorithm \ref{alg:3} is employed to generate ten different data sets for each fixed value of $m$. For each data set, Algorithms \ref{alg:4} and  \ref{alg:5} are implemented ten times to derive the average values of $Acc^t$ and $CS$ for different question selection strategies. Afterwards, the average values of $Acc^t$ and $CS$ for different question selection strategies on different data sets are determined, and the results are recorded in Figs. \ref{fig:acc_m}~-~\ref{fig:cs_m}.

\begin{figure}
\centering
\includegraphics[scale=0.33]{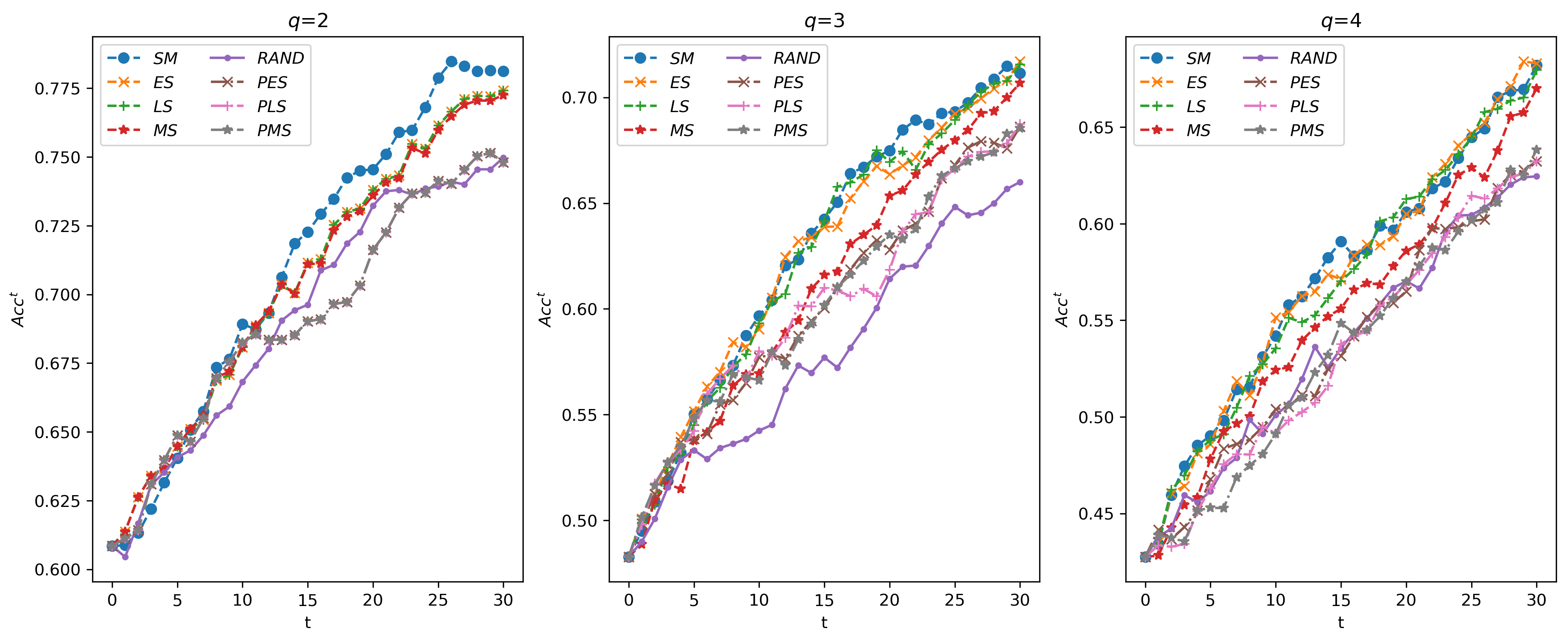}
\caption{The average values of  $Acc^t$ for different question selection strategies with different values of $q$}
\label{fig:acc_q}
\includegraphics[scale=0.35]{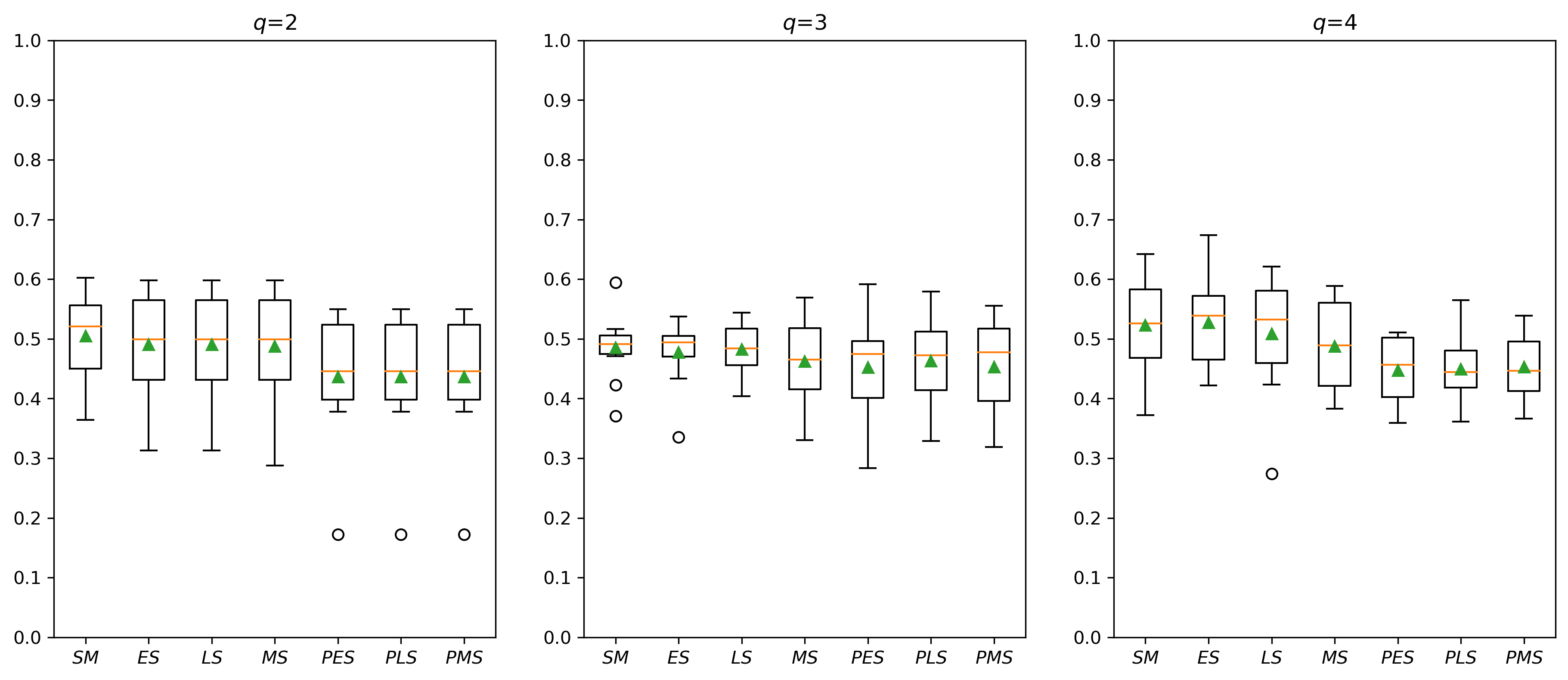}
\caption{The average values of  $CS$ for different question selection strategies with different values of $q$}
\label{fig:cs_q}
\end{figure}

\textbf{Experiment \RNum{6}:} In this experiment, we analyze the impact of the number of categories $q$ on the considered question selection strategies. To do so, we set $n=100$, $m=4$, $s_j=4$, $j=1,2,3,4$, $lr=0.2$, $T=30$, $r=0.6$, $\eta=0.05$, $\alpha=0.1$, and the values of $q$ are set as 2, 3, 4, respectively. As with previous experiments, we first execute Algorithm \ref{alg:3} for each combination of parameters to
generate ten data sets. For each considered question selection strategy and data set, we execute Algorithms \ref{alg:4} and  \ref{alg:5} ten times and calculate the average values of $Acc^t$ and $CS$ for different strategies on each data set. Subsequently, the average results across all data sets are computed, as shown in Figs. \ref{fig:acc_q}~-~\ref{fig:cs_q}.

{\color{black}
By analyzing the results shown in Figs. \ref{fig:acc_alpha}~-~\ref{fig:cs_q}, we can draw the following conclusions:

(1) In Experiments \RNum{1}~-~\RNum{6}, the performance of the proposed strategies $SM$, $ES$, $LS$ and $MS$ outperforms the benchmark strategies $RAND$, $PES$, $PLS$ and $PMS$ in most cases across various parameter settings. This is evident from two perspectives. First, the $Acc^t$ values of the proposed strategies consistently exceed those of the benchmark strategies throughout the incremental preference elicitation process.
Second, the $CS$ values for the strategies $SM$, $ES$, $LS$, $MS$, $PES$, $PMS$ and $PLS$ are all greater than 0, indicating a significant reduction in the number of assignment example preference information from decision makers compared to the $RAND$ strategy. Additionally, the average $CS$ values of the proposed strategies  are higher than those of the benchmark strategies $PES$, $PLS$ and $PMS$ in most parameter settings, reinforcing their superior performance.

(2) Two notable scenarios arise under different parameter settings. The first involves the change in $Acc^t$ when $\alpha=0.5$ (see Fig. \ref{fig:acc_alpha}). In this scenario, the performance trend of the considered strategies is unclear, likely due to high inconsistencies in the inferred sorting model, making it difficult to reflect the actual situation. Therefore, although $\alpha$ ranges from 0 to 1, we recommend setting it between 0 and 0.5 for better performance. The second scenario concerns  the change in $Acc^t$ and $CS$ when $n=50$ (see Figs. \ref{fig:acc_n} and \ref{fig:cs_n}). In this scenario, the distinction between the strategies $SM$, $ES$, $LS$, and $MS$ and the strategies $PES$, $PLS$, and $PMS$ is not obvious, with the average $CS$ values of $PES$, $PLS$, and $PMS$ being higher than those of the strategies $SM$, $ES$, $LS$, and $MS$. This indicates that the proposed strategies performs better with larger values of $n$.

(3) The $Acc^t$ results from Experiments \RNum{1}~-~\RNum{6} reveal that as $\alpha$, $\eta$, $m$, $q$ increase, the initial accuracy ($Acc^0$) of the eight considered strategies decreases. This is because higher $\alpha$ values cause the models \eqref{m:max_margin} and \eqref{m:min_slope} to pay less attention to the inconsistencies, leading to a higher extent of inconsistencies in the inferred sorting model and thus reducing the performance. As $\eta$  increase, the proportion of noise contained in the data sets (increased extent of inconsistencies) further diminishes the initial accuracy. Additionally, higher values of $m$ and $q$  increase the complexity of the data sets, straining the models' ability to handle complex data and resulting in decreased initial accuracy.
Conversely, as $lr$ and $n$ increase, $Acc^0$ values of the eight considered strategies improve, due to more initial assignment example preference information, allowing for more thorough model training.

(4) The $CS$ results from Experiments \RNum{1}~-~\RNum{6} show that  $CS$ values increase with rising $\alpha$ and $\eta$, decrease with increasing $lr$ and $m$, and remain relatively stable as $n$ and $q$ change. The changes in $CS$ values are closely tied to the target accuracy and initial accuracy. Different parameter settings result in varying initial and target accuracy results, leading to different trends in $CS$ values.

}

\subsection{Computational experiments on real-world data sets}\label{sec:6.3}
In this subsection, we further validate the effectiveness of the proposed approach using computational experiments on some real-world data sets, taken from the existing literature related to multi-criteria decision making across various domains, including transport, logistics, environment, education and so on \citep{Kadzinski21ejor,Liu23joc}. The details of these data sets are summarized in Table \ref{table:real_data}.
\begin{table}[htbp]
\centering
\setlength{\abovecaptionskip}{0pt}
\setlength{\belowcaptionskip}{10pt}
\caption{The details of the nine real-world data sets}
\label{table:real_data}
\begin{tabular}{lcc}
\toprule
Data sets & The number of alternatives $n$ & The number of criteria $m$    \\ \midrule
Buses (BU) & 76 & 8 \\
Environmental zones (EZ) & 69 & 5 \\
Students (ST) & 76 & 6 \\
Couple's embryos (CE) & 51 & 7 \\
Suppliers (SU) & 50 & 6 \\
Nanomaterials (NA) & 48 & 8 \\
Storage location (SL) & 50 & 4 \\
Research units HS1EK (RH) & 93 & 4 \\
Research units NZ1M (RN) & 78 & 4\\
\bottomrule
\end{tabular}
\end{table}

\begin{figure}[htbp]
\centering
\includegraphics[scale=0.3]{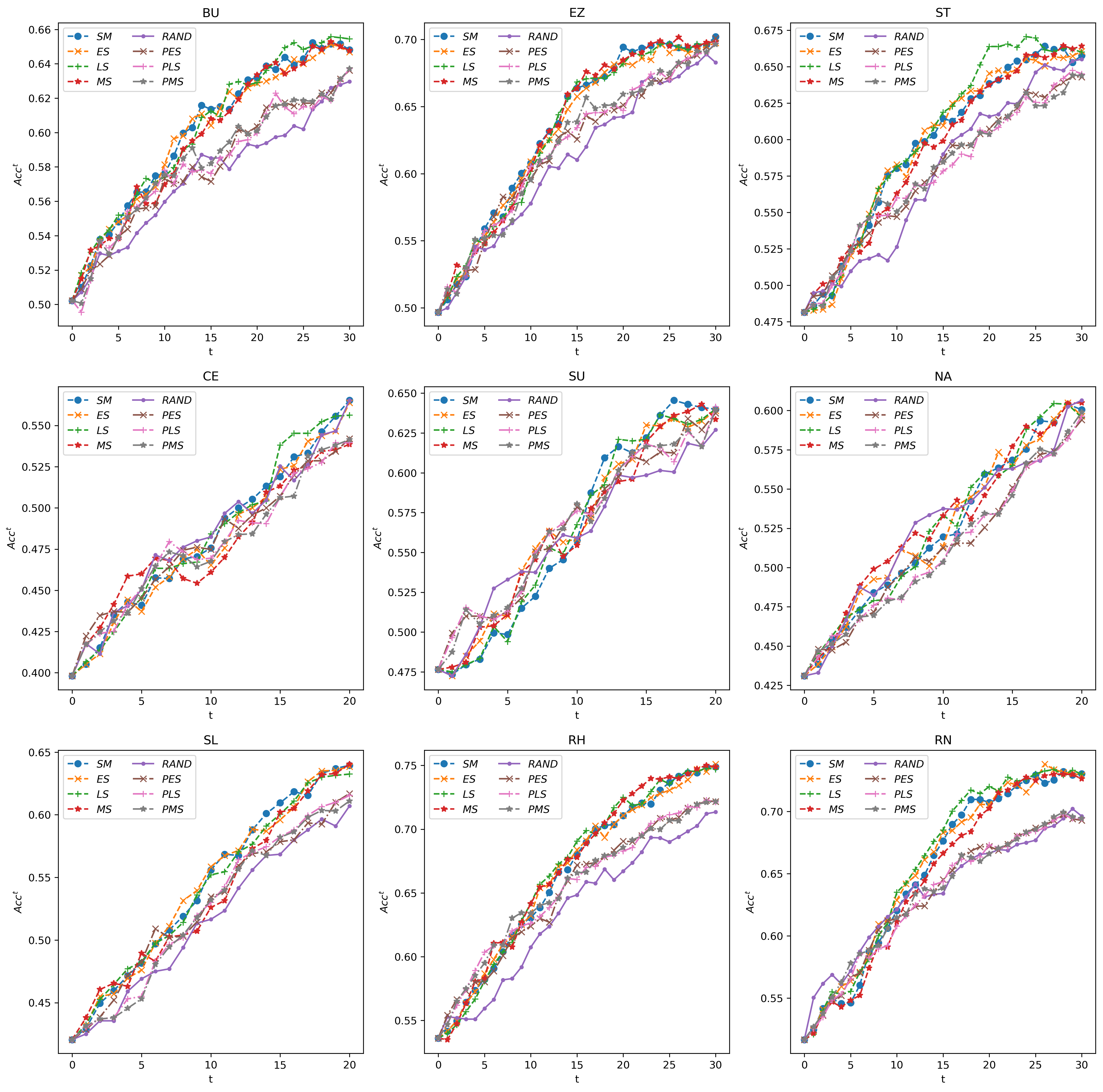}
\caption{The average values of  $Acc^t$ for different question selection strategies over the considered real-world data sets}
\label{fig:real_acc}

\includegraphics[scale=0.38]{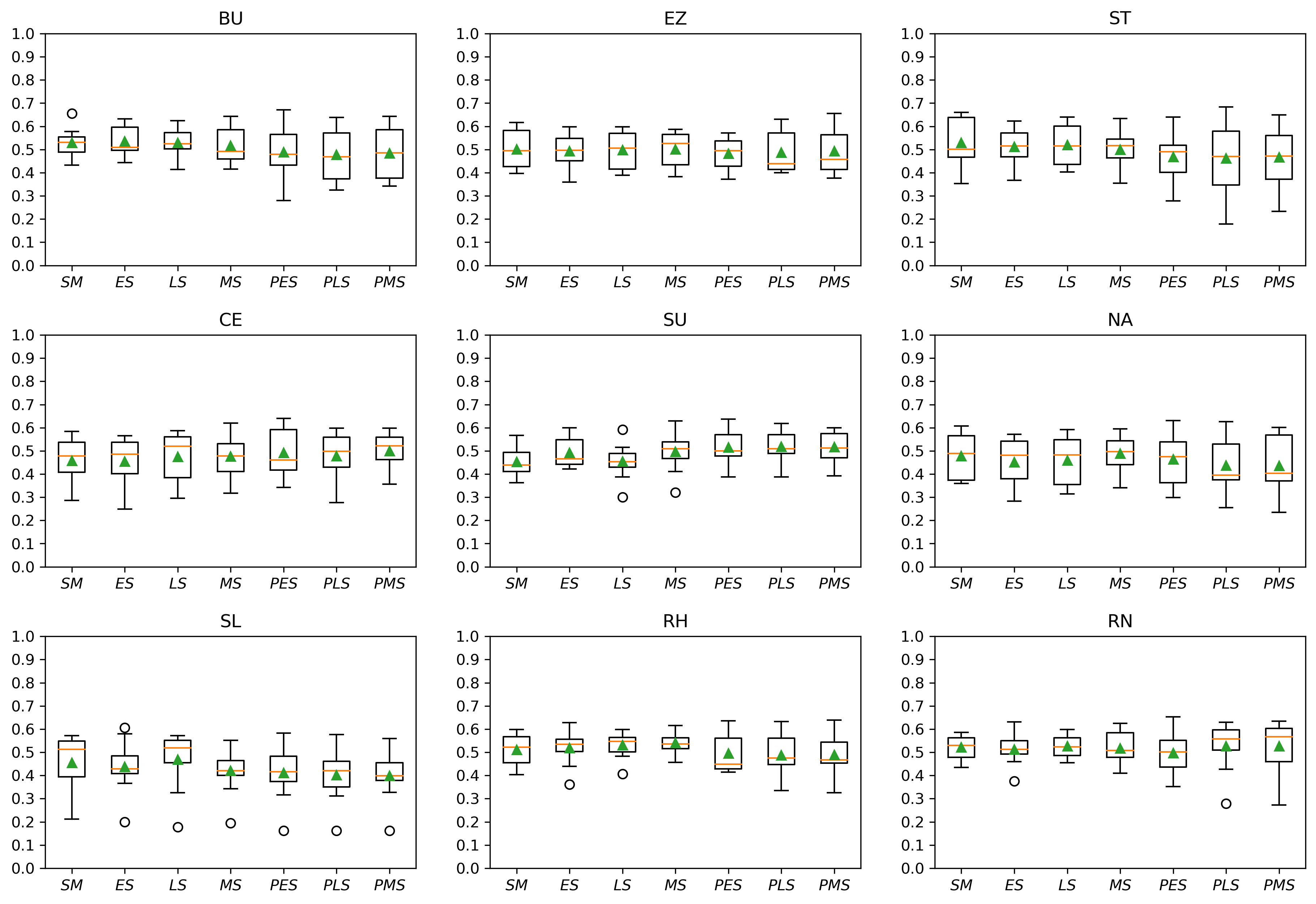}
\caption{The average values of  $CS$ for different question selection strategies over the considered real-world data sets}
\label{fig:real_cs}
\end{figure}

To construct sorting problem instances, for each considered real-world data set, we set $q=3$, $s_j=4$, $j\in M$ and $\eta=0.05$, and implement Steps 2~-~6 of Algorithm \ref{alg:3} ten times to generate ten data sets with category assignments. In the sequel,
we let $r=0.6$, $lr=0.2$, $\alpha=0.1$, $T=20$ if $n\le 50$ and $T=30$ if $n>50$, and then implement Algorithms \ref{alg:4} and \ref{alg:5}
for each generated data set. We repeat the above process ten times for each generated real-world data set and then calculate the average values of $Acc^t$ and $CS$ as the final results, which are illustrated in Figs. \ref{fig:real_acc}~-~\ref{fig:real_cs}.

{\color{black}
The results shown in Figs. \ref{fig:real_acc}~-~\ref{fig:real_cs} validate some conclusions derived from the computational experiments on artificial data sets. First, the $Acc^t$ and $CS$ values of the proposed strategies $SM$, $ES$, $LS$ and $MS$ outperform those of the strategies $RAND$, $PES$, $PLS$ and $PMS$ on most real-world data sets (BU, EZ, ST, SL, RH). This once again verifies the effectiveness of the proposed approach. Second, for the data sets $CE$ and $SU$, the differences in $Acc^t$ values among the eight considered strategies are relatively small, and the average $CS$ values of the strategies $PES$, $PLS$ and $PMS$ are higher than those of the strategies $SM$, $ES$, $LS$ and $MS$. This observation is consistent with the finding (2) presented in Section \ref{sec:6.2}.

Furthermore, several novel findings are observed. Specifically, for the data set NA, the differences in $Acc^t$ values across all considered strategies are relatively small. However, the average $CS$ values of the strategies $SM$ and $MS$ surpass those of the strategies $PLS$ and $PMS$, and the average $CS$ value of the strategy $PES$ exceeds that of the strategies $ES$ and $LS$. For the data set RN, the strategies $SM$, $ES$, $LS$ and $MS$ perform better in $Acc^t$ compared to the strategies $RAND$, $PES$, $PLS$ and $PMS$, but the average $CS$ values of the strategies $PLS$ and $PES$ are higher than those of the remaining strategies. This indicates that for certain data sets, the considered strategies may show inconsistent results in terms of $Acc^t$ and $CS$ metrics, presenting some challenges to the question selection strategies in the incremental preference elicitation process.

Additionally, the proposed question selection strategies exhibit distinct performance on different data sets, and the disparities among them are not notably conspicuous. This phenomenon may primarily stem from the fact that all question selection strategies proposed in this paper are grounded in the uncertainty sampling framework of active learning.

}

\section{Conclusion}\label{sec:6}
In this paper, we propose an incremental preference elicitation-based approach to learning potentially non-monotonic preferences for MCS problems. To be specific, we first develop a max-margin optimization-based model for addressing both inconsistent assignment example preference information and potentially non-monotonic preferences. Subsequently, based on the max-margin optimization-based model, we introduce two types of measures for assessing the information amount associated with each alternative, i.e., sum margin-based measure and probability-based measure.  Afterwards, we introduce three specific metrics for measuring the information amount of alternatives. Building upon this foundation, seven question selection strategies are proposed to identify the most informative alternative during each iteration of the incremental preference elicitation process. The proposed question selection strategies operate within the uncertainty sampling framework in active learning. Following this, in light of different termination criteria, we present two distinct incremental preference elicitation-based algorithms designed to assist the decision analyst in eliciting the preferences of the decision maker. Furthermore, to elaborate the detailed implementation process of the proposed approach, we apply it to address a credit rating problem drawn from the existing literature. Ultimately, to validate the effectiveness of the proposed approach, we undertake extensive computational experiments on both artificial and real-world data sets. The experimental results reveal that the proposed question selection strategies demonstrate advantages when compared to several benchmark strategies.

In addition, we outline several potential directions for future research. First, the proposed approach can be extended to other MCS methods, such as outranking-based methods \citep{Dias18omega,Almeida10ejor}. Second, it is a compelling avenue to explore incremental preference elicitation method for MCS problems with interacting criteria \citep{Liu20joc,Aggarwal19joc}. Third, we intend to study how to deal with large-scale assignment example preference information in MCS problems by integrating the incremental preference elicitation process.
{\color{black}Finally, enhancing our proposed approach by incorporating techniques from the polyhedral method in adaptive choice-based conjoint analysis \citep{Toubia04jmr} presents an intriguing research opportunity.
}

\ack{This work was partly supported by the National Natural Science Foundation of China under Grant Nos.
72371049 and 71971039, the Funds for Humanities and Social Sciences of Ministry of Education of China
under Grant No. 23YJC630219, the Natural Science Foundation of Liaoning Province under Grant No. 2024-MSBA-26, and the Fundamental Research Funds for the Central Universities of China
under Grant No. DUT23RW406.}

\bibliographystyle{model5-names}
\bibliography{reference}

\begin{thebibliography}{51}
\expandafter\ifx\csname natexlab\endcsname\relax\def\natexlab#1{#1}\fi
\providecommand{\bibinfo}[2]{#2}
\ifx\xfnm\relax \def\xfnm[#1]{\unskip,\space#1}\fi
\bibitem[{Aggarwal et~al.(2014)Aggarwal, Kong, Gu, Han \&
  Philip}]{Aggarwal14dc}
\bibinfo{author}{Aggarwal, C.~C.}, \bibinfo{author}{Kong, X.},
  \bibinfo{author}{Gu, Q.}, \bibinfo{author}{Han, J.}, \&
  \bibinfo{author}{Philip, S.~Y.} (\bibinfo{year}{2014}).
\newblock \bibinfo{title}{Active learning: A survey}.
\newblock In {\it \bibinfo{booktitle}{Data Classification}\/} (pp.
  \bibinfo{pages}{599--634}).
\newblock \bibinfo{publisher}{Chapman and Hall/CRC}.
\bibitem[{Aggarwal \& Fallah~Tehrani(2019)}]{Aggarwal19joc}
\bibinfo{author}{Aggarwal, M.}, \& \bibinfo{author}{Fallah~Tehrani, A.}
  (\bibinfo{year}{2019}).
\newblock \bibinfo{title}{Modelling human decision behaviour with preference
  learning}.
\newblock {\it \bibinfo{journal}{INFORMS Journal on Computing}\/},  {\it
  \bibinfo{volume}{31}\/}, \bibinfo{pages}{318--334}.
\bibitem[{Almeida-Dias et~al.(2010)Almeida-Dias, Figueira \&
  Roy}]{Almeida10ejor}
\bibinfo{author}{Almeida-Dias, J.}, \bibinfo{author}{Figueira, J.~R.}, \&
  \bibinfo{author}{Roy, B.} (\bibinfo{year}{2010}).
\newblock \bibinfo{title}{{Electre Tri-C}: A multiple criteria sorting method
  based on characteristic reference actions}.
\newblock {\it \bibinfo{journal}{European Journal of Operational Research}\/},
  {\it \bibinfo{volume}{204}\/}, \bibinfo{pages}{565--580}.
\bibitem[{Belahc\'{e}ne et~al.(2023)Belahc\'{e}ne, Mousseau, Ouerdane, Pirlot
  \& Sobrie}]{Belahcene234or}
\bibinfo{author}{Belahc\'{e}ne, K.}, \bibinfo{author}{Mousseau, V.},
  \bibinfo{author}{Ouerdane, W.}, \bibinfo{author}{Pirlot, M.}, \&
  \bibinfo{author}{Sobrie, O.} (\bibinfo{year}{2023}).
\newblock \bibinfo{title}{Multiple criteria sorting models and methods-part
  {I}: survey of the literature}.
\newblock {\it \bibinfo{journal}{4OR}\/},  {\it \bibinfo{volume}{21}\/},
  \bibinfo{pages}{1--46}.
\bibitem[{Benabbou et~al.(2017)Benabbou, Perny \& Viappiani}]{Benabbou17ai}
\bibinfo{author}{Benabbou, N.}, \bibinfo{author}{Perny, P.}, \&
  \bibinfo{author}{Viappiani, P.} (\bibinfo{year}{2017}).
\newblock \bibinfo{title}{Incremental elicitation of {Choquet} capacities for
  multicriteria choice, ranking and sorting problems}.
\newblock {\it \bibinfo{journal}{Artificial Intelligence}\/},  {\it
  \bibinfo{volume}{246}\/}, \bibinfo{pages}{152--180}.
\bibitem[{Despotis \& Zopounidis(1995)}]{Despotis95ama}
\bibinfo{author}{Despotis, D.~K.}, \& \bibinfo{author}{Zopounidis, C.}
  (\bibinfo{year}{1995}).
\newblock \bibinfo{title}{Building additive utilities in the presence of
  non-monotonic preferences}.
\newblock {\it \bibinfo{journal}{Advances in Multicriteria Analysis}\/},  (pp.
  \bibinfo{pages}{101--114}).
\bibitem[{Devaud et~al.(1980)Devaud, Groussaud \&
  Jacquet-Lagreze}]{Devaud80ewg}
\bibinfo{author}{Devaud, J.~M.}, \bibinfo{author}{Groussaud, G.}, \&
  \bibinfo{author}{Jacquet-Lagreze, E.} (\bibinfo{year}{1980}).
\newblock \bibinfo{title}{{UTADIS}: Une m\'{e}thode de construction de
  fonctions d$^\prime$utilit\'{e} additives rendant compte de jugements
  globaux}.
\newblock In {\it \bibinfo{booktitle}{European Working Group on Multicriteria
  Decision Aid, Bochum}\/}.
\newblock volume~\bibinfo{volume}{94}.
\bibitem[{Dias et~al.(2018)Dias, Antunes, Dantas, de~Castro \&
  Zamboni}]{Dias18omega}
\bibinfo{author}{Dias, L.~C.}, \bibinfo{author}{Antunes, C.~H.},
  \bibinfo{author}{Dantas, G.}, \bibinfo{author}{de~Castro, N.}, \&
  \bibinfo{author}{Zamboni, L.} (\bibinfo{year}{2018}).
\newblock \bibinfo{title}{A multi-criteria approach to sort and rank policies
  based on {Delphi} qualitative assessments and {ELECTRE TRI}: The case of
  smart grids in {Brazil}}.
\newblock {\it \bibinfo{journal}{Omega}\/},  {\it \bibinfo{volume}{76}\/},
  \bibinfo{pages}{100--111}.
\bibitem[{Doumpos \& Figueira(2019)}]{Doumpos19omega}
\bibinfo{author}{Doumpos, M.}, \& \bibinfo{author}{Figueira, J.~R.}
  (\bibinfo{year}{2019}).
\newblock \bibinfo{title}{A multicriteria outranking approach for modeling
  corporate credit ratings: An application of the {Electre Tri-nC} method}.
\newblock {\it \bibinfo{journal}{Omega}\/},  {\it \bibinfo{volume}{82}\/},
  \bibinfo{pages}{166--180}.
\bibitem[{Doumpos et~al.(2009)Doumpos, Marinakis, Marinaki \&
  Zopounidis}]{Doumpos09ejor}
\bibinfo{author}{Doumpos, M.}, \bibinfo{author}{Marinakis, Y.},
  \bibinfo{author}{Marinaki, M.}, \& \bibinfo{author}{Zopounidis, C.}
  (\bibinfo{year}{2009}).
\newblock \bibinfo{title}{An evolutionary approach to construction of
  outranking models for multicriteria classification: The case of the {ELECTRE
  TRI} method}.
\newblock {\it \bibinfo{journal}{European Journal of Operational Research}\/},
  {\it \bibinfo{volume}{199}\/}, \bibinfo{pages}{496--505}.
\bibitem[{Doumpos \& Zopounidis(2011)}]{Doumpos11ejor}
\bibinfo{author}{Doumpos, M.}, \& \bibinfo{author}{Zopounidis, C.}
  (\bibinfo{year}{2011}).
\newblock \bibinfo{title}{Preference disaggregation and statistical learning
  for multicriteria decision support: A review}.
\newblock {\it \bibinfo{journal}{European Journal of Operational Research}\/},
  {\it \bibinfo{volume}{209}\/}, \bibinfo{pages}{203--214}.
\bibitem[{Doumpos \& Zopounidis(2019)}]{Doumpos19springer}
\bibinfo{author}{Doumpos, M.}, \& \bibinfo{author}{Zopounidis, C.}
  (\bibinfo{year}{2019}).
\newblock \bibinfo{title}{Preference disaggregation for multicriteria decision
  aiding: An overview and perspectives}.
\newblock In {\it \bibinfo{booktitle}{New Perspectives in Multiple Criteria
  Decision Making}\/} (pp. \bibinfo{pages}{115--130}).
\newblock \bibinfo{address}{Cham}: \bibinfo{publisher}{Springer International
  Publishing}.
\bibitem[{Doumpos et~al.(2014)Doumpos, Zopounidis \&
  Galariotis}]{Doumpos14ejor}
\bibinfo{author}{Doumpos, M.}, \bibinfo{author}{Zopounidis, C.}, \&
  \bibinfo{author}{Galariotis, E.} (\bibinfo{year}{2014}).
\newblock \bibinfo{title}{Inferring robust decision models in multicriteria
  classification problems: An experimental analysis}.
\newblock {\it \bibinfo{journal}{European Journal of Operational Research}\/},
  {\it \bibinfo{volume}{236}\/}, \bibinfo{pages}{601--611}.
\bibitem[{Esmaelian et~al.(2016)Esmaelian, Shahmoradi \&
  Vali}]{Esmaelian16asoc}
\bibinfo{author}{Esmaelian, M.}, \bibinfo{author}{Shahmoradi, H.}, \&
  \bibinfo{author}{Vali, M.} (\bibinfo{year}{2016}).
\newblock \bibinfo{title}{A novel classification method: A hybrid approach
  based on extension of the {UTADIS} with polynomial and {PSO-GA} algorithm}.
\newblock {\it \bibinfo{journal}{Applied Soft Computing}\/},  {\it
  \bibinfo{volume}{49}\/}, \bibinfo{pages}{56--70}.
\bibitem[{Fern\'{a}ndez et~al.(2019)Fern\'{a}ndez, Figueira \&
  Navarro}]{Fernandez19asoc}
\bibinfo{author}{Fern\'{a}ndez, E.}, \bibinfo{author}{Figueira, J.~R.}, \&
  \bibinfo{author}{Navarro, J.} (\bibinfo{year}{2019}).
\newblock \bibinfo{title}{An indirect elicitation method for the parameters of
  the {ELECTRE TRI-nB} model using genetic algorithms}.
\newblock {\it \bibinfo{journal}{Applied Soft Computing}\/},  {\it
  \bibinfo{volume}{77}\/}, \bibinfo{pages}{723--733}.
\bibitem[{Gehrlein et~al.(2023)Gehrlein, Miebs, Brunelli \&
  Kadzi\'{n}ski}]{Gehrlein23omega}
\bibinfo{author}{Gehrlein, J.}, \bibinfo{author}{Miebs, G.},
  \bibinfo{author}{Brunelli, M.}, \& \bibinfo{author}{Kadzi\'{n}ski, M.}
  (\bibinfo{year}{2023}).
\newblock \bibinfo{title}{An active preference learning approach to aid the
  selection of validators in blockchain environments}.
\newblock {\it \bibinfo{journal}{Omega}\/},  {\it \bibinfo{volume}{118}\/},
  \bibinfo{pages}{102869}.
\bibitem[{Ghaderi \& Kadzi\'{n}ski(2021)}]{Ghaderi21omega}
\bibinfo{author}{Ghaderi, M.}, \& \bibinfo{author}{Kadzi\'{n}ski, M.}
  (\bibinfo{year}{2021}).
\newblock \bibinfo{title}{Incorporating uncovered structural patterns in value
  functions construction}.
\newblock {\it \bibinfo{journal}{Omega}\/},  {\it \bibinfo{volume}{99}\/},
  \bibinfo{pages}{102203}.
\bibitem[{Ghaderi et~al.(2017)Ghaderi, Ruiz \& Agell}]{Ghaderi17ejor}
\bibinfo{author}{Ghaderi, M.}, \bibinfo{author}{Ruiz, F.}, \&
  \bibinfo{author}{Agell, N.} (\bibinfo{year}{2017}).
\newblock \bibinfo{title}{A linear programming approach for learning
  non-monotonic additive value functions in multiple criteria decision aiding}.
\newblock {\it \bibinfo{journal}{European Journal of Operational Research}\/},
  {\it \bibinfo{volume}{259}\/}, \bibinfo{pages}{1073--1084}.
\bibitem[{Greco et~al.(2010)Greco, Mousseau \& S{\l}owi\'{n}ski}]{Greco10ejor}
\bibinfo{author}{Greco, S.}, \bibinfo{author}{Mousseau, V.}, \&
  \bibinfo{author}{S{\l}owi\'{n}ski, R.} (\bibinfo{year}{2010}).
\newblock \bibinfo{title}{Multiple criteria sorting with a set of additive
  value functions}.
\newblock {\it \bibinfo{journal}{European Journal of Operational Research}\/},
  {\it \bibinfo{volume}{207}\/}, \bibinfo{pages}{1455--1470}.
\bibitem[{Guo et~al.(2019)Guo, Liao \& Liu}]{Guo19eswa}
\bibinfo{author}{Guo, M.}, \bibinfo{author}{Liao, X.}, \& \bibinfo{author}{Liu,
  J.} (\bibinfo{year}{2019}).
\newblock \bibinfo{title}{A progressive sorting approach for multiple criteria
  decision aiding in the presence of non-monotonic preferences}.
\newblock {\it \bibinfo{journal}{Expert Systems with Applications}\/},  {\it
  \bibinfo{volume}{123}\/}, \bibinfo{pages}{1--17}.
\bibitem[{H\"{u}llermeier \&
  S{\l}owi\'{n}ski(2024{\natexlab{a}})}]{Hullermeier24-4or1}
\bibinfo{author}{H\"{u}llermeier, E.}, \& \bibinfo{author}{S{\l}owi\'{n}ski,
  R.} (\bibinfo{year}{2024}{\natexlab{a}}).
\newblock \bibinfo{title}{Preference learning and multiple criteria decision
  aiding: differences, commonalities, and synergies-part
  \uppercase\expandafter{\romannumeral 1}}.
\newblock {\it \bibinfo{journal}{4OR}\/}, .
\newblock \bibinfo{note}{In press, doi: 10.1007/s10288-023-00560-6}.
\bibitem[{H\"{u}llermeier \&
  S{\l}owi\'{n}ski(2024{\natexlab{b}})}]{Hullermeier24-4or2}
\bibinfo{author}{H\"{u}llermeier, E.}, \& \bibinfo{author}{S{\l}owi\'{n}ski,
  R.} (\bibinfo{year}{2024}{\natexlab{b}}).
\newblock \bibinfo{title}{Preference learning and multiple criteria decision
  aiding: differences, commonalities, and synergies-part
  \uppercase\expandafter{\romannumeral 2}}.
\newblock {\it \bibinfo{journal}{4OR}\/}, .
\newblock \bibinfo{note}{In press, doi: 10.1007/s10288-023-00561-5}.
\bibitem[{Jacquet-Lagr\'{e}ze \& Siskos(2001)}]{Jacquet01ejor}
\bibinfo{author}{Jacquet-Lagr\'{e}ze, E.}, \& \bibinfo{author}{Siskos, Y.}
  (\bibinfo{year}{2001}).
\newblock \bibinfo{title}{Preference disaggregation: 20 years of mcda
  experience}.
\newblock {\it \bibinfo{journal}{European Journal of Operational Research}\/},
  {\it \bibinfo{volume}{130}\/}, \bibinfo{pages}{233--245}.
\bibitem[{Kadzi\'{n}ski \& Ciomek(2021)}]{Kadzinski21ejor}
\bibinfo{author}{Kadzi\'{n}ski, M.}, \& \bibinfo{author}{Ciomek, K.}
  (\bibinfo{year}{2021}).
\newblock \bibinfo{title}{Active learning strategies for interactive
  elicitation of assignment examples for threshold-based multiple criteria
  sorting}.
\newblock {\it \bibinfo{journal}{European Journal of Operational Research}\/},
  {\it \bibinfo{volume}{293}\/}, \bibinfo{pages}{658--680}.
\bibitem[{Kadzi\'{n}ski et~al.(2020)Kadzi\'{n}ski, Ghaderi \&
  D\k{a}browski}]{Kadzinski20ejor}
\bibinfo{author}{Kadzi\'{n}ski, M.}, \bibinfo{author}{Ghaderi, M.}, \&
  \bibinfo{author}{D\k{a}browski, M.} (\bibinfo{year}{2020}).
\newblock \bibinfo{title}{Contingent preference disaggregation model for
  multiple criteria sorting problem}.
\newblock {\it \bibinfo{journal}{European Journal of Operational Research}\/},
  {\it \bibinfo{volume}{281}\/}, \bibinfo{pages}{369--387}.
\bibitem[{Kadzi\'{n}ski et~al.(2017)Kadzi\'{n}ski, Ghaderi, W\k{a}sikowski \&
  Agell}]{Kadzinski17caor}
\bibinfo{author}{Kadzi\'{n}ski, M.}, \bibinfo{author}{Ghaderi, M.},
  \bibinfo{author}{W\k{a}sikowski, J.}, \& \bibinfo{author}{Agell, N.}
  (\bibinfo{year}{2017}).
\newblock \bibinfo{title}{Expressiveness and robustness measures for the
  evaluation of an additive value function in multiple criteria preference
  disaggregation methods: An experimental analysis}.
\newblock {\it \bibinfo{journal}{Computers \& Operations Research}\/},  {\it
  \bibinfo{volume}{87}\/}, \bibinfo{pages}{146--164}.
\bibitem[{Kadzi\'{n}ski et~al.(2021)Kadzi\'{n}ski, Martyn, Cinelli,
  S{\l}owi\'{n}ski, Corrente \& Greco}]{Kadzinski21kbs}
\bibinfo{author}{Kadzi\'{n}ski, M.}, \bibinfo{author}{Martyn, K.},
  \bibinfo{author}{Cinelli, M.}, \bibinfo{author}{S{\l}owi\'{n}ski, R.},
  \bibinfo{author}{Corrente, S.}, \& \bibinfo{author}{Greco, S.}
  (\bibinfo{year}{2021}).
\newblock \bibinfo{title}{Preference disaggregation method for value-based
  multi-decision sorting problems with a real-world application in
  nanotechnology}.
\newblock {\it \bibinfo{journal}{Knowledge-Based Systems}\/},  {\it
  \bibinfo{volume}{218}\/}, \bibinfo{pages}{106879}.
\bibitem[{Kadzi\'{n}ski et~al.(2016)Kadzi\'{n}ski, S{\l}owi\'{n}ski \&
  Greco}]{Kadzinski16ins}
\bibinfo{author}{Kadzi\'{n}ski, M.}, \bibinfo{author}{S{\l}owi\'{n}ski, R.}, \&
  \bibinfo{author}{Greco, S.} (\bibinfo{year}{2016}).
\newblock \bibinfo{title}{Robustness analysis for decision under uncertainty
  with rule-based preference model}.
\newblock {\it \bibinfo{journal}{Information Sciences}\/},  {\it
  \bibinfo{volume}{328}\/}, \bibinfo{pages}{321--339}.
\bibitem[{Kadzi\'{n}ski \& Tervonen(2013)}]{Kadzinski13dss}
\bibinfo{author}{Kadzi\'{n}ski, M.}, \& \bibinfo{author}{Tervonen, T.}
  (\bibinfo{year}{2013}).
\newblock \bibinfo{title}{Stochastic ordinal regression for multiple criteria
  sorting problems}.
\newblock {\it \bibinfo{journal}{Decision Support Systems}\/},  {\it
  \bibinfo{volume}{55}\/}, \bibinfo{pages}{55--66}.
\bibitem[{Kadzi\'{n}ski et~al.(2015)Kadzi\'{n}ski, Tervonen \&
  Rui~Figueira}]{Kadzinski15omega}
\bibinfo{author}{Kadzi\'{n}ski, M.}, \bibinfo{author}{Tervonen, T.}, \&
  \bibinfo{author}{Rui~Figueira, J.} (\bibinfo{year}{2015}).
\newblock \bibinfo{title}{Robust multi-criteria sorting with the outranking
  preference model and characteristic profiles}.
\newblock {\it \bibinfo{journal}{Omega}\/},  {\it \bibinfo{volume}{55}\/},
  \bibinfo{pages}{126--140}.
\bibitem[{Khannoussi et~al.(2022)Khannoussi, Olteanu, Labreuche \&
  Meyer}]{Khannoussi224or}
\bibinfo{author}{Khannoussi, A.}, \bibinfo{author}{Olteanu, A.-L.},
  \bibinfo{author}{Labreuche, C.}, \& \bibinfo{author}{Meyer, P.}
  (\bibinfo{year}{2022}).
\newblock \bibinfo{title}{Simple ranking method using reference profiles:
  incremental elicitation of the preference parameters}.
\newblock {\it \bibinfo{journal}{4OR}\/},  {\it \bibinfo{volume}{20}\/},
  \bibinfo{pages}{499--530}.
\bibitem[{Khannoussi et~al.(2024)Khannoussi, Olteanu, Meyer \&
  Benabbou}]{Khannoussi24or}
\bibinfo{author}{Khannoussi, A.}, \bibinfo{author}{Olteanu, A.-L.},
  \bibinfo{author}{Meyer, P.}, \& \bibinfo{author}{Benabbou, N.}
  (\bibinfo{year}{2024}).
\newblock \bibinfo{title}{A regret-based query selection strategy for the
  incremental elicitation of the criteria weights in an {SRMP} model}.
\newblock {\it \bibinfo{journal}{Operational research}\/},  {\it
  \bibinfo{volume}{24}\/}, \bibinfo{pages}{12}.
\bibitem[{Li \& Zhang(2024)}]{Li24tcss}
\bibinfo{author}{Li, Z.}, \& \bibinfo{author}{Zhang, Z.}
  (\bibinfo{year}{2024}).
\newblock \bibinfo{title}{Threshold-based value-driven method to support
  consensus reaching in multicriteria group sorting problems: A minimum
  adjustment perspective}.
\newblock {\it \bibinfo{journal}{IEEE Transactions on Computational Social
  Systems}\/},  {\it \bibinfo{volume}{11}\/}, \bibinfo{pages}{1230--1243}.
\bibitem[{Li et~al.(2024)Li, Zhang \& Yu}]{Li24jors}
\bibinfo{author}{Li, Z.}, \bibinfo{author}{Zhang, Z.}, \& \bibinfo{author}{Yu,
  W.} (\bibinfo{year}{2024}).
\newblock \bibinfo{title}{Consensus reaching for ordinal classification-based
  group decision making with heterogeneous preference information}.
\newblock {\it \bibinfo{journal}{Journal of the Operational Research
  Society}\/},  {\it \bibinfo{volume}{75}\/}, \bibinfo{pages}{224--245}.
\bibitem[{de~Lima~Silva \& de~Almeida~Filho(2020)}]{de20caie}
\bibinfo{author}{de~Lima~Silva, D.~F.}, \& \bibinfo{author}{de~Almeida~Filho,
  A.~T.} (\bibinfo{year}{2020}).
\newblock \bibinfo{title}{Sorting with {TOPSIS} through boundary and
  characteristic profiles}.
\newblock {\it \bibinfo{journal}{Computers \& Industrial Engineering}\/},  {\it
  \bibinfo{volume}{141}\/}, \bibinfo{pages}{106328}.
\bibitem[{de~Lima~Silva et~al.(2020)de~Lima~Silva, Ferreira \&
  de~Almeida-Filho}]{de20eswa}
\bibinfo{author}{de~Lima~Silva, D.~F.}, \bibinfo{author}{Ferreira, L.}, \&
  \bibinfo{author}{de~Almeida-Filho, A.~T.} (\bibinfo{year}{2020}).
\newblock \bibinfo{title}{A new preference disaggregation {TOPSIS} approach
  applied to sort corporate bonds based on financial statements and expert's
  assessment}.
\newblock {\it \bibinfo{journal}{Expert systems with applications}\/},  {\it
  \bibinfo{volume}{152}\/}, \bibinfo{pages}{113369}.
\bibitem[{Liu et~al.(2023)Liu, Kadzi\'{n}ski \& Liao}]{Liu23joc}
\bibinfo{author}{Liu, J.}, \bibinfo{author}{Kadzi\'{n}ski, M.}, \&
  \bibinfo{author}{Liao, X.} (\bibinfo{year}{2023}).
\newblock \bibinfo{title}{Modeling contingent decision behavior: A bayesian
  nonparametric preference-learning approach}.
\newblock {\it \bibinfo{journal}{INFORMS Journal on Computing}\/},  {\it
  \bibinfo{volume}{35}\/}, \bibinfo{pages}{764--785}.
\bibitem[{Liu et~al.(2020{\natexlab{a}})Liu, Kadzi\'{n}ski, Liao \&
  Mao}]{Liu20joc}
\bibinfo{author}{Liu, J.}, \bibinfo{author}{Kadzi\'{n}ski, M.},
  \bibinfo{author}{Liao, X.}, \& \bibinfo{author}{Mao, X.}
  (\bibinfo{year}{2020}{\natexlab{a}}).
\newblock \bibinfo{title}{Data-driven preference learning methods for
  value-driven multiple criteria sorting with interacting criteria}.
\newblock {\it \bibinfo{journal}{INFORMS Journal on Computing}\/},  {\it
  \bibinfo{volume}{33}\/}, \bibinfo{pages}{586--606}.
\bibitem[{Liu et~al.(2020{\natexlab{b}})Liu, Kadzi\'{n}ski, Liao, Mao \&
  Wang}]{Liu20ejor}
\bibinfo{author}{Liu, J.}, \bibinfo{author}{Kadzi\'{n}ski, M.},
  \bibinfo{author}{Liao, X.}, \bibinfo{author}{Mao, X.}, \&
  \bibinfo{author}{Wang, Y.} (\bibinfo{year}{2020}{\natexlab{b}}).
\newblock \bibinfo{title}{A preference learning framework for multiple criteria
  sorting with diverse additive value models and valued assignment examples}.
\newblock {\it \bibinfo{journal}{European Journal of Operational Research}\/},
  {\it \bibinfo{volume}{286}\/}, \bibinfo{pages}{963--985}.
\bibitem[{Liu et~al.(2019)Liu, Liao, Kadzi\'{n}ski \&
  S{\l}owi\'{n}ski}]{Liu19ejor}
\bibinfo{author}{Liu, J.}, \bibinfo{author}{Liao, X.},
  \bibinfo{author}{Kadzi\'{n}ski, M.}, \& \bibinfo{author}{S{\l}owi\'{n}ski,
  R.} (\bibinfo{year}{2019}).
\newblock \bibinfo{title}{Preference disaggregation within the regularization
  framework for sorting problems with multiple potentially non-monotonic
  criteria}.
\newblock {\it \bibinfo{journal}{European Journal of Operational Research}\/},
  {\it \bibinfo{volume}{276}\/}, \bibinfo{pages}{1071--1089}.
\bibitem[{Liu et~al.(2016)Liu, Liao, Zhao \& Yang}]{Liu16omega}
\bibinfo{author}{Liu, J.}, \bibinfo{author}{Liao, X.}, \bibinfo{author}{Zhao,
  W.}, \& \bibinfo{author}{Yang, N.} (\bibinfo{year}{2016}).
\newblock \bibinfo{title}{A classification approach based on the outranking
  model for multiple criteria {ABC} analysis}.
\newblock {\it \bibinfo{journal}{Omega}\/},  {\it \bibinfo{volume}{61}\/},
  \bibinfo{pages}{19--34}.
\bibitem[{Nefla et~al.(2019)Nefla, \"{O}zt\"{u}rk, Viappiani \&
  Brigui-Chtioui}]{Nefla19adt}
\bibinfo{author}{Nefla, O.}, \bibinfo{author}{\"{O}zt\"{u}rk, M.},
  \bibinfo{author}{Viappiani, P.}, \& \bibinfo{author}{Brigui-Chtioui, I.}
  (\bibinfo{year}{2019}).
\newblock \bibinfo{title}{Interactive elicitation of a majority rule sorting
  model with maximum margin optimization}.
\newblock In {\it \bibinfo{booktitle}{Proceedings of the 6th International
  Conference on Algorithmic Decision Theory}\/} (pp.
  \bibinfo{pages}{141--157}).
\newblock \bibinfo{address}{Durham, NC, USA}.
\bibitem[{\"{O}zpeynirci et~al.(2018)\"{O}zpeynirci, \"{O}zpeynirci \&
  Mousseau}]{Ozpeynirci18AOR}
\bibinfo{author}{\"{O}zpeynirci, S.}, \bibinfo{author}{\"{O}zpeynirci, O.}, \&
  \bibinfo{author}{Mousseau, V.} (\bibinfo{year}{2018}).
\newblock \bibinfo{title}{An interactive algorithm for multiple criteria
  constrained sorting problem}.
\newblock {\it \bibinfo{journal}{Annals of Operations Research}\/},  {\it
  \bibinfo{volume}{267}\/}, \bibinfo{pages}{447--466}.
\bibitem[{Pelissari \& Duarte(2022)}]{Pelissari22eswa}
\bibinfo{author}{Pelissari, R.}, \& \bibinfo{author}{Duarte, L.~T.}
  (\bibinfo{year}{2022}).
\newblock \bibinfo{title}{{SMAA-Choquet-FlowSort}: A novel
  user-preference-driven {Choquet classifier} applied to supplier evaluation}.
\newblock {\it \bibinfo{journal}{Expert Systems with Applications}\/},  {\it
  \bibinfo{volume}{207}\/}, \bibinfo{pages}{117898}.
\bibitem[{Ru et~al.(2023)Ru, Liu, Kadzi\'{n}ski \& Liao}]{Ru23ejor}
\bibinfo{author}{Ru, Z.}, \bibinfo{author}{Liu, J.},
  \bibinfo{author}{Kadzi\'{n}ski, M.}, \& \bibinfo{author}{Liao, X.}
  (\bibinfo{year}{2023}).
\newblock \bibinfo{title}{Probabilistic ordinal regression methods for multiple
  criteria sorting admitting certain and uncertain preferences}.
\newblock {\it \bibinfo{journal}{European Journal of Operational Research}\/},
  {\it \bibinfo{volume}{311}\/}, \bibinfo{pages}{596--616}.
\bibitem[{Teso et~al.(2016)Teso, Passerini \& Viappiani}]{Teso16ijcai}
\bibinfo{author}{Teso, S.}, \bibinfo{author}{Passerini, A.}, \&
  \bibinfo{author}{Viappiani, P.} (\bibinfo{year}{2016}).
\newblock \bibinfo{title}{Constructive preference elicitation by setwise
  max-margin learning}.
\newblock In {\it \bibinfo{booktitle}{Proceedings of the Twenty-Fifth
  International Joint Conference on Artificial Intelligence}\/} (pp.
  \bibinfo{pages}{2067--2073}).
\newblock \bibinfo{address}{New York, NY, USA}.
\bibitem[{Toubia et~al.(2004)Toubia, Hauser \& Simester}]{Toubia04jmr}
\bibinfo{author}{Toubia, O.}, \bibinfo{author}{Hauser, J.~R.}, \&
  \bibinfo{author}{Simester, D.~I.} (\bibinfo{year}{2004}).
\newblock \bibinfo{title}{Polyhedral methods for adaptive choice-based conjoint
  analysis}.
\newblock {\it \bibinfo{journal}{Journal of Marketing Research}\/},  {\it
  \bibinfo{volume}{41}\/}, \bibinfo{pages}{116--131}.
\bibitem[{W\'{o}jcik et~al.(2023)W\'{o}jcik, Kadzi\'{n}ski \&
  Ciomek}]{Wojcik23kbs}
\bibinfo{author}{W\'{o}jcik, M.}, \bibinfo{author}{Kadzi\'{n}ski, M.}, \&
  \bibinfo{author}{Ciomek, K.} (\bibinfo{year}{2023}).
\newblock \bibinfo{title}{Selection of a representative sorting model in a
  preference disaggregation setting: A review of existing procedures, new
  proposals, and experimental comparison}.
\newblock {\it \bibinfo{journal}{Knowledge-Based Systems}\/},  {\it
  \bibinfo{volume}{278}\/}, \bibinfo{pages}{110871}.
\bibitem[{Wu \& Liao(2023)}]{Wu23omega}
\bibinfo{author}{Wu, X.}, \& \bibinfo{author}{Liao, H.} (\bibinfo{year}{2023}).
\newblock \bibinfo{title}{A compensatory value function for modeling risk
  tolerance and criteria interactions in preference disaggregation}.
\newblock {\it \bibinfo{journal}{Omega}\/},  {\it \bibinfo{volume}{117}\/},
  \bibinfo{pages}{102836}.
\bibitem[{Zhang \& Li(2023)}]{Zhang23aor}
\bibinfo{author}{Zhang, Z.}, \& \bibinfo{author}{Li, Z.}
  (\bibinfo{year}{2023}).
\newblock \bibinfo{title}{Consensus-based {TOPSIS-Sort-B} for multi-criteria
  sorting in the context of group decision-making}.
\newblock {\it \bibinfo{journal}{Annals of Operations Research}\/},  {\it
  \bibinfo{volume}{325}\/}, \bibinfo{pages}{911--938}.
\bibitem[{Zopounidis \& Doumpos(2002)}]{Zopounidis02ejor}
\bibinfo{author}{Zopounidis, C.}, \& \bibinfo{author}{Doumpos, M.}
  (\bibinfo{year}{2002}).
\newblock \bibinfo{title}{Multicriteria classification and sorting methods: A
  literature review}.
\newblock {\it \bibinfo{journal}{European Journal of Operational Research}\/},
  {\it \bibinfo{volume}{138}\/}, \bibinfo{pages}{229--246}.

\end{thebibliography}

\newpage

\appendix

\pagenumbering{arabic}
\setcounter{page}{1}

\renewcommand{\thealgorithm}{\Alph{section}\arabic{algorithm}}
\setcounter{algorithm}{0}

\renewcommand{\thefigure}{\Alph{section}\arabic{figure}}
\setcounter{figure}{0}

\renewcommand{\thefigure}{\Alph{section}\arabic{figure}}
\setcounter{figure}{0}

\section{The algorithms for simulation experiments}\label{appendix:0}
\begin{algorithm}
\begin{small}
\caption{Generation of artificial data sets}\label{alg:3}
\begin{algorithmic}[1]
     \Require  The number of alternatives $n$, the number of criteria $m$, the number of categories $q$, the number of subintervals for each criterion $s_j$, $j\in M$, and the proportion of noise $\eta$.
     \Ensure The decision matrix $X=(x_{ij})_{n\times m}$ and the sorting result for alternatives $F=(f_1,f_2,\ldots,f_n)^{\rm T}$.
        \end{algorithmic}
\begin{enumerate}[\bf Step 1:]
  \item Randomly generate $x_{ij}$ in the interval $[0,100]$ following a uniform distribution, then form the decision matrix $X=(x_{ij})_{n\times m}$.
  \item Ascertain the minimum and maximum performance values for each criterion $g_j$. In light of the number of subintervals for each criterion $s_j$, $j\in M$, determine the characteristic points $\beta_j^l$, $l=1,\ldots,s_j+1$, $j\in M$ as described in Section \ref{sec:2}.
  \item Randomly generate the marginal utility for each characteristic point $u_j(\beta_j^l)$ within the interval $[0,1]$ from a uniform distribution, $l=1,\ldots,s_j+1$, $j\in M$.
  \item Calculate the comprehensive utility for each alternative $U(a_i)$, $i\in N$ by Eq. \eqref{eq:global_v}. On this basis, determine the category thresholds with the aim of achieving a roughly balanced distribution of alternatives across all categories.
  \item Employ the threshold-based MCS method to derive the sorting result for alternatives as $F^0=(f_1^0,f_2^0,\ldots,f_n^0)^{\rm T}$.
  \item Randomly select $[n\cdot \eta]$ alternatives and modify their sorting results. For other alternatives, keep their sorting results unchanged. Donote the final sorting result for all alternatives as $F=(f_1,f_2,\ldots,f_n)^{\rm T}$.
  \item Output $X$ and $F$.
\end{enumerate}
\end{small}
    \end{algorithm}

\begin{algorithm}[!h]
\begin{small}
\caption{The comparison algorithm when considering termination criterion \RNum{1}}\label{alg:4}
\begin{algorithmic}[1]
     \Require  The data set $D$, the proportion of training data set $r$, the proportion of initial assignment example preference information to the training data $lr$, and the maximum number of questions the decision maker can answer $T$.
     \Ensure The accuracy of the considered question selection strategy on the test data set.
        \end{algorithmic}
\begin{enumerate}[\bf Step 1:]
  \item In light of the proportion of training data set $r$, randomly partition the data set $D$ into a training data set $D_1$ and a test data set $D_2$, while maintaining a distribution that closely aligns with the original data set $D$. In particular, let $n_1=[n\cdot r]$ and $n_2=n - [n\cdot r]$ be the number of alternatives included in the training data set $D_1$ and the test data set $D_2$, respectively.
  \item Let $A$ denote all alternatives included in the training data set $D_1$. Randomly select $[lr \cdot n_1]$ alternatives from $A$ as the initial set of reference alternatives $A^{R}$, and then simulate an artificial decision maker whose answers during the incremental preference elicitation process align consistently with the sorting result for alternatives in the training data set $D_1$. By doing so, the initial set of assignment example preference information $S$ are determined.
  \item Let $t=0$, $S^t=S$ and $A^{R,t}=A^R$.
  \item For the considered question selection strategy, in terms of $S^t$, solve the models \eqref{m:max_margin} and \eqref{m:min_slope} to obtain the shape of marginal utility functions and category thresholds.
  \item For the considered question selection strategy, calculate the sorting result for all alternatives in the test data set $D_2$ based on the threshold-based MCS model. Based on the inferred and real sorting result for all alternatives in the test data set $D_2$, compute the accuracy metric by Eq. \eqref{eq:accuracy}, denoted as $Acc^t$.
  \item If $t<T$, proceed to the next step. Otherwise, directly go to Step 8.
  \item For the considered question selection strategy, implement Steps 3, 5~-~7 of Algorithm \ref{alg:1}, and then update $S^{t}$ and
  $A^{R,{t+1}}$ as $S^{t+1}=S^t\cup \{a^{*,t}\rightarrow  C_{h^*}\}$ and $A^{R,{t+1}}=A^{R,t}\cup \{a^{*,t}\}$. Let $t=t+1$, return to Step 4.
  \item Output the accuracy $Acc^t$, $t=0,1,\ldots, T$.
\end{enumerate}
\end{small}
    \end{algorithm}

\begin{algorithm}[!h]
\begin{small}
\caption{The comparison algorithm when considering termination criterion \RNum{2}}\label{alg:5}
\begin{algorithmic}[1]
     \Require  The data set $D$, the proportion of training data set $r$, the proportion of initial assignment example preference information to the training data $lr$.
     \Ensure The cost saving rate of the considered question selection strategy on the test data set.
        \end{algorithmic}
\begin{enumerate}[\bf Step 1:]
    \item Randomly partition the data set $D$ into a training data set $D_1$ and a test data set $D_2$ based on the proportion of training data set $r$, while maintaining a distribution that closely aligns with the original data set $D$. Let $A$ denote all alternatives included in the training data set $D_1$. Let $A^R=A$, and obtain the set of assignment example preference information $S$ in terms of the real sorting result for alternatives in the training data set $D_1$.
    \item Based on $S$, implement Steps 2~-~4 in Algorithm \ref{alg:2} to compute the sorting result for alternatives in the test data set $D_2$ and the accuracy metric.
    \item Repeat Steps 1 and 2 ten times, recording the accuracy metric for each time, and take the average accuracy as the target accuracy $Acc_{target}$.
    \item The same as Steps 1~-~6 in Algorithm \ref{alg:4}.
    \item If $Acc^t\ge Acc_{target}$, directly go to Step 7. Otherwise, go to the next step.
    \item Implement Steps 8~-~9 in Algorithm \ref{alg:4}.
    \item Let $aq=t$ be the number of questions the decision maker answered during the incremental preference elicitation process. By Eq. \eqref{eq:CS}, calculate the cost saving rate $CS$ as $\frac{n_1-aq-[lr\cdot n_1]}{n_1}$.
    \item Output the cost saving rate of the considered question selection strategy on the test data set.
\end{enumerate}
\end{small}
    \end{algorithm}

\newpage    
\section{Simulation results to analyze the relationship between potentially non-monotonic preferences and inconsistencies}\label{appendix:1}
\setcounter{algorithm}{0}

To assess whether the flexibility of potentially non-monotonic preferences can entirely address inconsistencies in the assignment example preference information, we conduct some simulation experiments. To do so, we propose a new optimization model to minimize the extent of inconsistencies in the assignment example preference information, based on the model \eqref{m:max_margin}. This new model is formulated by replacing the objective function of the model \eqref{m:max_margin} with $\min \ \sum\nolimits_{a_i\in A^{R,t}}\delta^+_i + \delta^-_i$ and removing the final constraint from model \eqref{m:max_margin}. This modified model is referred to as the model (M-4).
On this basis, we provide an algorithm to perform simulation analysis as follows (see Algorithm \ref{alg:A_1}).

\begin{algorithm}
\caption{Simulation algorithm to analyze the relationship between potentially non-monotonic preferences and inconsistencies}\label{alg:A_1}
\begin{algorithmic}[1]
     \Require  The number of alternatives $n$, the number of criteria $m$, the number of categories $q$, the number of subintervals for each criterion $s_j$, $j\in M$, and the proportion of noise $\eta$.
     \Ensure The minimum extent of inconsistencies $ICI$ in the assignment example preference information.
        \end{algorithmic}
\begin{enumerate}[\bf Step 1:]
  \item Use Algorithm \ref{alg:3} to generate a data set in terms of $n$, $m$, $q$, $s_j$, $j\in M$ and $\eta$.
  \item Treat all alternatives and their sorting results in the simulated data set as the assignment example preference information. Solve the model (M-4) to derive the minimum extent of inconsistencies $ICI$.
  \item Output $ICI$.
\end{enumerate}
    \end{algorithm}

\begin{figure}[!t]
\centering
\includegraphics[scale=0.35]{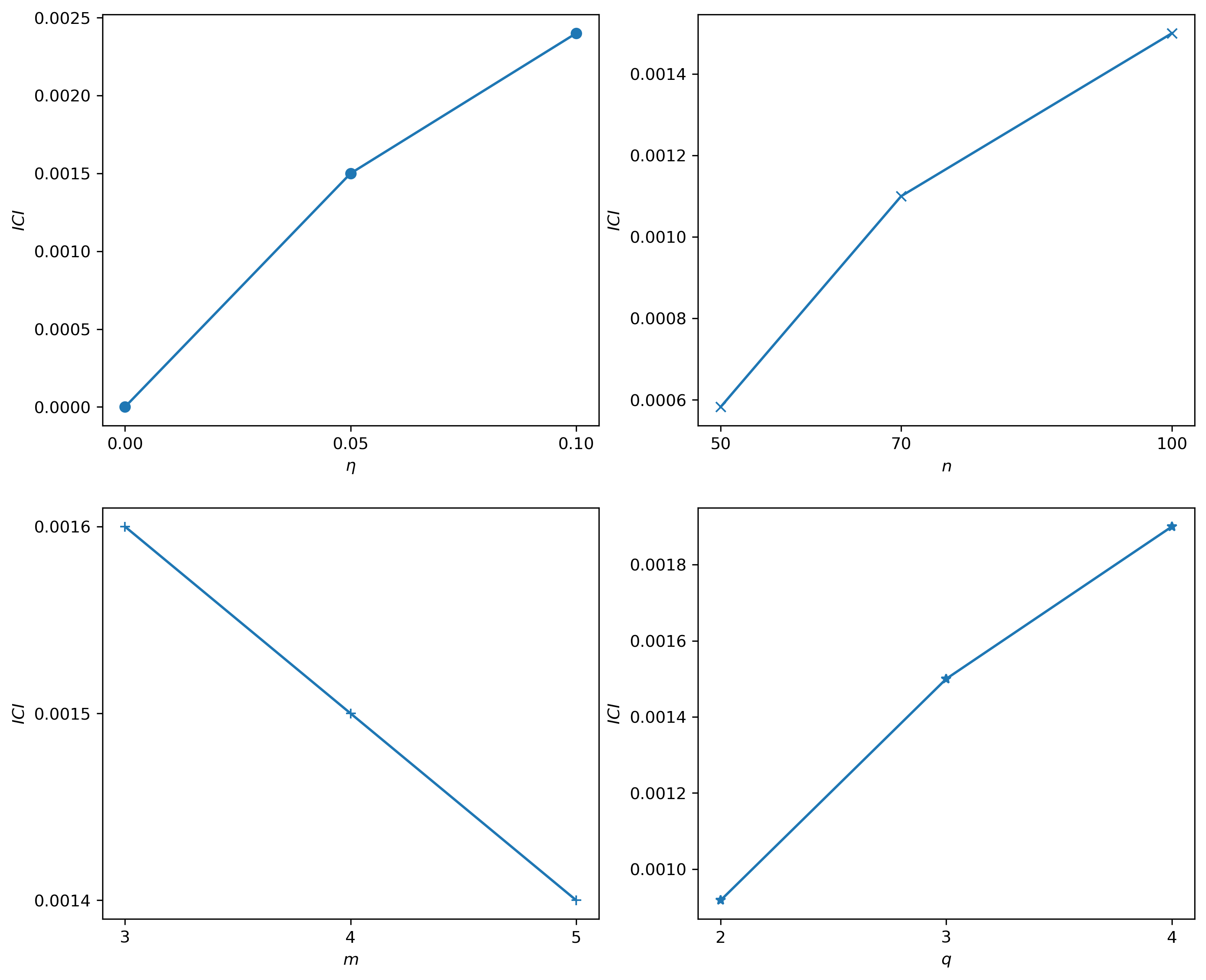}
\caption{The average minimum extent of inconsistencies $ICI$ for different parameter settings}
\label{fig:ICI}
\end{figure}

Let us consider the following parameter settings: (1) $n=100$, $m=4$, $q=3$, $s_j=4$, $j=1,2,3,4$, $\eta\in\{0,0.05,0.1\}$. (2) $n\in\{50,70,100\}$, $m=4$, $q=3$, $s_j=4$, $j=1,2,3,4$, $\eta=0.05$. (3) $n=100$, $m\in\{3,4,5\}$, $q=3$, $s_j=4$, $j=1,2,3,4$, $\eta=0.05$. (4)  $n=100$, $m=4$, $q\in\{2,3,4\}$, $s_j=4$, $j=1,2,3,4$, $\eta=0.05$. For each parameter setting, we implement Algorithm \ref{alg:A_1} 1000 times, and record the average value of the minimum extent of inconsistencies $ICI$. The results are displayed in Fig. \ref{fig:ICI}.

From Fig. \ref{fig:ICI}, it is observed that the average minimum extent of inconsistencies is zero only when the proportion of noise, $\eta$, is equal to zero. For all non-zero values of $\eta$, the minimum extent of inconsistencies exceeds zero. This suggests that while our model accommodates potentially non-monotonic preferences, the extent of inconsistencies in the assignment example preference information is not sheltered by the flexibility of potentially non-monotonic preferences.

Furthermore, as the proportion of noise in the data sets increases and the number of alternatives and categories rises, the extent of inconsistencies also increases. This implies that the complexity of the MCS problem contributes to a higher degree of inconsistencies. Conversely, when the number of criteria increases, the extent of inconsistencies tends to decrease. This reduction may be attributed to the increased flexibility for utility functions when a higher number of criteria are provided. This flexibility enhances the model's ability to tolerate inconsistencies, thereby decreasing the extent of inconsistencies in the assignment example preference information.

\section{Discussions about the marginal utility functions' shape derived in the illustrative example}\label{appendix:2}

The marginal utility functions derived in the illustrative example, i.e., Fig. \ref{fig:transformed_U}, exhibit non-monotonic changes for the three considered criteria in the credit rating problem. In what follows, we provide some discussions about the derived marginal utility functions' shape.

For the criterion $g_1$, there is a significant increase in marginal utility as $x_{ij}$ rises from 0 to around 10, followed by a sharp decrease and another increase. This suggests that the increasing of $g_1$ initially has a positive effect, but after a certain point, it has a negative effect, followed by another positive effect. In practice, a high cash to total assets indicates good liquidity and strong debt repayment ability, which positively influences the financial state. However, an excessively high ratio suggest that the firm is not effectively utilizing its cash for investment or growth, negatively affecting the financial state. Further increases might ensure a robust cash reserve to handle market fluctuations, thus positively impacting the financial state.

For the criterion $g_2$, at low values of $x_{ij}$, the marginal utility decreases. As $x_{ij}$ increases, the marginal utility then rises sharply and eventually stabilizes at high values. A low long term debt and stockholders equity to fixed assets might indicate that the firm does not have enough long-term capital to support its fixed assets, which can lead to a worse financial state, as the ratio increases, the firm starts to have more long-term capital to support its fixed assets. When the ratio reaches a certain level, further increases have diminishing marginal effects on the financial state because the firm already has sufficient long-term capital and fixed assets allocation to support its operations.

For the criterion $g_3$, as $x_{ij}$ increases from a low value, the marginal utility rises to a peak, then drops significantly, and finally rises again. An initial increase in the total liabilities to total assets can indicate that the firm is leveraging its assets to finance growth, which might positively influence the financial state up to a certain point. However, as the ratio continues to increase, the financial risk associated with higher liabilities becomes more pronounced, leading to a decrease in the marginal utility. At very high values, another increase in the ratio might indicate the firm is utilizing its debt efficiently and managing it well, leading to an improvement in the financial state.

Although the shape of marginal utility functions may appear unusual, it can still partially explain the impact of each criterion on the firms’ financial states in the context of credit rating.

\end{document}